\documentclass{ecai} 
\usepackage{latexsym}
\usepackage{amssymb}
\usepackage{amsmath}
\usepackage{amsthm}
\usepackage{booktabs}
\usepackage{enumitem}
\usepackage{graphicx}
\usepackage{color}

\usepackage{array}
\usepackage{amsfonts}
\usepackage{subcaption}
\usepackage{url}

\usepackage{fancyhdr}

\fancypagestyle{firstpage}{%
  \fancyhead{}
  \fancyfoot[L]{Preprint}
}

\newcommand{\BibTeX}{B\kern-.05em{\sc i\kern-.025em b}\kern-.08em\TeX}

\begin{document}

%%%%%%%%%%%%%%%%%%%%%%%%%%%%%%%%%%%%%%%%%%%%%%%%%%%%%%%%%%%%%%%%%%%%%%%%

\begin{frontmatter}

%%% Use this command to specify your submission number.
%%% In doubleblind mode, it will be printed on the first page.

\paperid{462} 

%%% Use this command to specify the title of your paper.

\title{Can we Constrain Concept Bottleneck Models to\\Learn Semantically Meaningful Input Features?}

%%% Use this combinations of commands to specify all authors of your 
%%% paper. Use \fnms{} and \snm{} to indicate everyone's first names 
%%% and surname. This will help the publisher with indexing the 
%%% proceedings. Please use a reasonable approximation in case your 
%%% name does not neatly split into "first names" and "surname".
%%% Specifying your ORCID digital identifier is optional. 
%%% Use the \thanks{} command to indicate one or more corresponding 
%%% authors and their email address(es). If so desired, you can specify
%%% author contributions using the \footnote{} command.

\author[A]{\fnms{Jack}~\snm{Furby}\thanks{Corresponding Author. Email: furbyjl@cardiff.ac.uk}}
\author[B]{\fnms{Daniel}~\snm{Cunnington}}
\author[C]{\fnms{Dave}~\snm{Braines}} 
\author[A]{\fnms{Alun}~\snm{Preece}} 

\address[A]{Cardiff University, UK}
\address[B]{Imperial College London}
\address[C]{IBM Research Europe}

%%% Use this environment to include an abstract of your paper.

\begin{abstract}
    Concept Bottleneck Models (CBMs) are regarded as inherently interpretable because they first predict a set of human-defined concepts which are used to predict a task label. For inherent interpretability to be fully realised, and ensure trust in a model’s output, it's desirable for concept predictions to use semantically meaningful input features. For instance, in an image, pixels representing a broken bone should contribute to predicting a fracture. However, current literature suggests that concept predictions often rely on irrelevant input features. We hypothesise that this occurs when dataset labels include inaccurate concept annotations, or the relationship between input features and concepts is unclear. In general, the effect of dataset labelling on concept representations remains an understudied area. In this paper, we demonstrate that CBMs can learn to map concepts to semantically meaningful input features, by utilising datasets with a clear link between the input features and the desired concept predictions. This is achieved, for instance, by ensuring multiple concepts do not always co-occur and, therefore provide a clear training signal for the CBM to distinguish the relevant input features for each concept. We validate our hypothesis on both synthetic and real-world image datasets, and demonstrate under the correct conditions, CBMs can learn to attribute semantically meaningful input features to the correct concept predictions.
\end{abstract}

\end{frontmatter}

\thispagestyle{firstpage}

\section{Introduction} \label{Introduction_section}

Concept Bottleneck Models (CBMs) \citep{pmlr-v119-koh20a} have been positioned as improving human-machine collaboration as they are inherently interpretable \citep{pmlr-v119-koh20a}. This capability is enabled by the model predicting a vector of human-defined concepts and a task label. Concept predictions can be inspected to reveal the reasoning of a task prediction more easily, since the user can probe the CBM with various combinations of concepts. However, concept predictions may be misleading to humans interpreting the machine's outputs if the model does not predict concepts based on their expected input features, but the human assumes it does. Let's say we have a model diagnosing patients from X-ray images using concepts such as ``fracture'' or ``edema'', a radiologist may assume the model is making use of the same features they used, potentially resulting in the radiologist missing overlooked features or including irrelevant ones. For humans to make full use of a model they will need to trust the model's output, but a lack of understanding of the causes for a decision may result in a loss of trust \citep{MILLER20191}. To fully realise the interpretability benefits CBMs provide, ideally the input data is mapped to concepts that are aligned with human intuition.

During training, both the concepts and the task labels are supervised with the model split into two parts, a \textit{concept encoder} to map inputs to concepts, and the \textit{task predictor} to map concepts to task labels. Splitting the model enables a human to \textit{intervene} in the concept predictions by modifying them and passing them back through the task predictor \citep{pmlr-v119-koh20a}.

We define \textit{semantically meaningful} as the prediction of concepts based on input features that share the same meaning. Such as if the concept ``fracture'' is predicted as present then the pixels representing the bone and break should be the primary input features used. Current literature indicates CBMs do not map input features to concepts semantically \citep{furby2023towards,margeloiu2021concept}. It is thought that CBMs are taking shortcuts by using the same pixels to predict multiple concepts \citep{furby2023towards} or encoding more information for each concept, allowing concepts to be predicted from one another \citep{glanceNets,zarlenga2023towards}. Existing research is limited as it only focuses on the dataset Caltech-UCSD Birds-200-2011 (CUB) \citep{WahCUB_200_2011}, which leaves open the question of how CBMs learn to represent concepts using other datasets. To the best of our knowledge, the dataset has had little consideration for how it affects concept representations. We hypothesise if a CBM is unable to learn to map input features to concepts semantically, it is caused by inaccurate or weakly defined concept annotations, and by instead using a dataset with accurate concept annotations with a clear link to input features, a CBM can learn semantically meaningful concepts. As CBMs are only trained on task and concept labels they have no ground truth as to what features they should use from input samples and instead are left to discover this. Considering this reliance on the training dataset, we should avoid concepts that always appear together in case a CBM cannot distinguish between them and concepts that do not have a visual representation in the input.

This paper seeks to answer the following research questions:

\begin{itemize}
    \item Can CBMs learn to predict concepts based on semantically meaningful input features?
    \item How does the co-occurrence of concepts in a dataset impact the input features CBMs use to predict concepts?
    \item What dataset attributes contribute toward training CBMs to predict concepts using semantically meaningful input features?
\end{itemize}

We approach these questions by evaluating CBMs trained on two datasets: a synthetic playing card image dataset, which enables us to explore a best-case scenario where concept annotations are accurate and all concepts are accompanied by a visual representation, and a chest X-ray image dataset, an example of real-world images. We configured these datasets with multiple concept configurations to explore how the co-occurrence of concepts affected the representations CBMs learn. We have not modified the underlying CBM architecture or training methods.

This paper presents a comprehensive analysis of the concept representations CBMs learn with various dataset configurations. We focus on training CBMs on images containing visual features that depict concepts. Our contributions are threefold: (1) In contrast to previous studies, we demonstrate the ability of CBMs to learn semantically meaningful concept representations. We show this by (2) introducing a new synthetic image dataset with fine-grained concept annotations. (3) We perform an in-depth analysis of CBMs and conclude that two factors are critical for CBMs to learn semantically meaningful input features: (i) accuracy of concept annotations and (ii) high variability in the combinations of concepts co-occurring.

\section{Related Work} \label{related_work_section}

As discussed in the introduction, \citep{furby2023towards,margeloiu2021concept} explored how input features map to concepts for a dataset with \textit{class-level concepts} \citep{pmlr-v119-koh20a}, where each sample in a class has the same set of concepts. They found concepts were not predicted using semantically meaningful input features. They used CUB, a popular dataset for CBM research. Most CUB concepts represent bird parts, however, if the image is cropped, or the bird changes appearance with gender or age, concepts may no longer match the visual appearance in the image. To support this, \citep{furby2023towards,margeloiu2021concept} use saliency maps produced with model-specific techniques, such as Integrated Gradients (IG) \citep{10.5555/3305890.3306024} or Layer-wise Relevance Propagation (LRP) \citep{10.1371/journal.pone.0130140}, to work out the relevance of pixels used for concept predictions and saliency maps to visualise the relevance. Saliency maps are a popular way to visualise input feature relevance and have been used in situations such as to communicate bias a model has learned \citep{ribeiro-etal-2016-trust}. Saliency maps are local explanations which means any explanation only relates to a single sample and model. We believe there is no prior work which looks at the attribution of relevance to the input features that contribute to a model's output(s) with CBMs trained on other datasets such as ones with \textit{instance-level concepts}, where each sample in a dataset has its own concept vector \citep{pmlr-v119-koh20a}. Instance-level concepts avoid the inaccurate concept annotations seen with CUB as concepts can be fine-grained, only annotating concepts as present when their visual representations can be identified in the input. Without additional evaluation on datasets with alternative dataset configurations, such as ones with instance-level concepts, it may be assumed CBMs are incapable of predicting concepts from semantically meaningful input features.

Other metrics to analyse the concepts learned by CBMs fall under the term \textit{concept leakage} \citep{Mahinpei2021PromisesAP}, and originate from \textit{disentanglement} metrics \citep{disentangle}, where it's generally desired for concepts to only encode information that is required to predict themselves. \citep{Mahinpei2021PromisesAP} evaluates whether concepts excluded from training can be predicted better than a random guess, while \citep{glanceNets} trains models on fixed data ranges and tests whether concepts can be predicted outside of these constraints. \citep{zarlenga2023towards} introduces Oracle and Niche scores to measure \textit{concept purity}. These scores measure inter-concept predictability w.r.t. the expected predictive performance of the dataset. Concept purity is argued to be better suited to CBMs than alternative disentanglement metrics. Finally, \citep{raman2023do} measures whether correlated but unrelated concepts in a dataset affect concept predictions, finding CBMs tend to learn combinations of concepts in the training data.

Alternative to CBMs, other approaches to achieving semantically meaningful concept predictions have been shown with alternative model architectures such as \citep{protopnet,glanceNets,Huang_Song_Hu_Zhang_Wang_Song_2024}. As our work is positioned on exploring the previously mentioned gap in the literature looking at the affect the dataset has on CBMs learned concept representations. For this reason other model architectures are out of scope.

We position our work for specialised tasks where annotated data is available or easily created \citep{oikarinen2023labelfree} due to the complexity of creating large-scale datasets with concept annotations \citep{kazhdan2021disentanglement}. Moving beyond this, recent developments with Large Language Models \citep{oikarinen2023labelfree,Yang_2023_CVPR} and language-vision models, Contrastive Language-Image Pretraining (CLIP) \citep{Radford2021LearningTV}, have the potential to automate the annotation process. Large Language Models are used to generate potential concepts and CLIP discards concepts not relevant to the input image. For now, however, we cannot guarantee all generated concepts relate to semantically meaningful input features in the input image. This can be observed in \citep{oikarinen2023labelfree} with the concept ``long tail with white stripes'' for an image of a bird where the tail is obscured in the image.

\section{Concept Bottleneck Models}

A CBM takes an input, in our case an image, to the concept encoder to predict a vector of concepts. These concept predictions are then used as the input for the task predictor to predict a task label. Concept predictions are values in the range of 0 to 1 where 0 means the model is confident the concept is not present and a prediction of 1 means the model is confident the concept is present. Predictions of 0.5 and above are counted as present.

CBMs are trained by supervising both the concepts and the task given the training set $\{x^{(i)}, y^{(i)}, c^{(i)}\}_{i=1}^n$ where we are provided with a set of inputs $x \in \mathbb{R}^d$, corresponding targets $y \in \mathcal{Y}$ and vectors of $k$ concepts $c \in \mathbb{R}^k$. A CBM in the form $f(g(x))$ uses two functions: $g : \mathbb{R}^d \rightarrow \mathbb{R}^k$ to map an input to the concept space and $f : \mathbb{R}^k \rightarrow \mathcal{Y}$ to map concepts to the task output. This is such that the task prediction is made using only the predicted concepts.

There are three methods we can use to train a CBM; the \textit{independent} method where each of $f$ and $g$ are trained separately and then joined together after training, the \textit{sequential} method where $f$ is trained and then $g$ trained using the output of $f$, and the \textit{joint} method where $f$ and $g$ are trained together.

In the related work section, we introduced two primary dataset configurations: class-level and instance-level concepts. Class-level concepts apply to entire classes, meaning all instances within a class share the same concepts. For example, if we represented individual playing cards with concepts for their suit and rank, we would apply the concepts of ``Two'' and ``Hearts'' to every sample in the ``Two of Hearts'' class. Instance-level concepts, however, are assigned to individual samples, allowing two samples from the same class to have different concepts. This approach is more versatile and can accommodate a wider range of situations. For instance, in a dataset of poker hands, concepts might represent specific cards like ``Four of Clubs'' and ``Ten of Diamonds'', while task classes could be ``Straight Flus'' and ``Three of a Kind''. If we have a full set of playing cards one task class can be formed from multiple combinations of playing cards, and thus can only be represented using instance-level concepts.

\section{Experiment Set-up} \label{setup_section}

\subsection{Datasets}

\begin{figure}[ht!]
     \centering
     \begin{subfigure}[b]{0.13\textwidth}
         \centering
         \includegraphics[width=\textwidth]{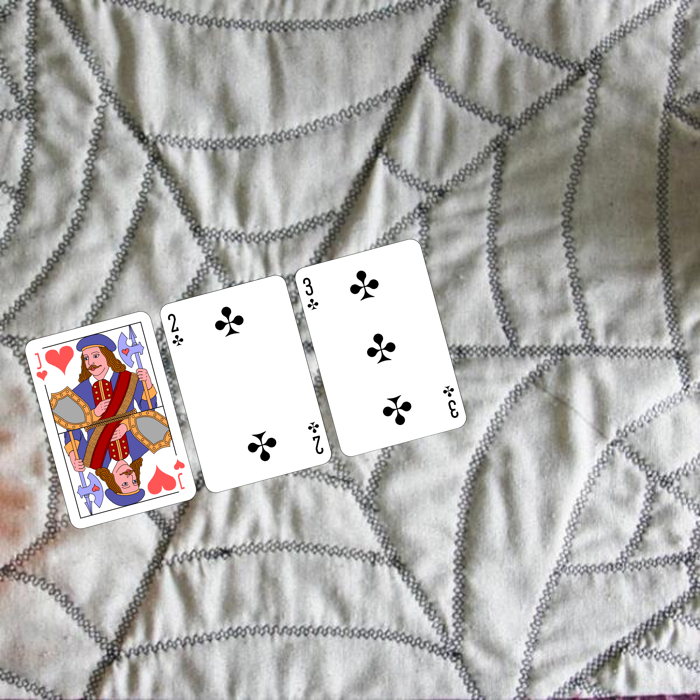}
         \caption{Random cards\newline}
     \end{subfigure}
     \hfill 
     \begin{subfigure}[b]{0.13\textwidth}
         \centering
         \includegraphics[width=\textwidth]{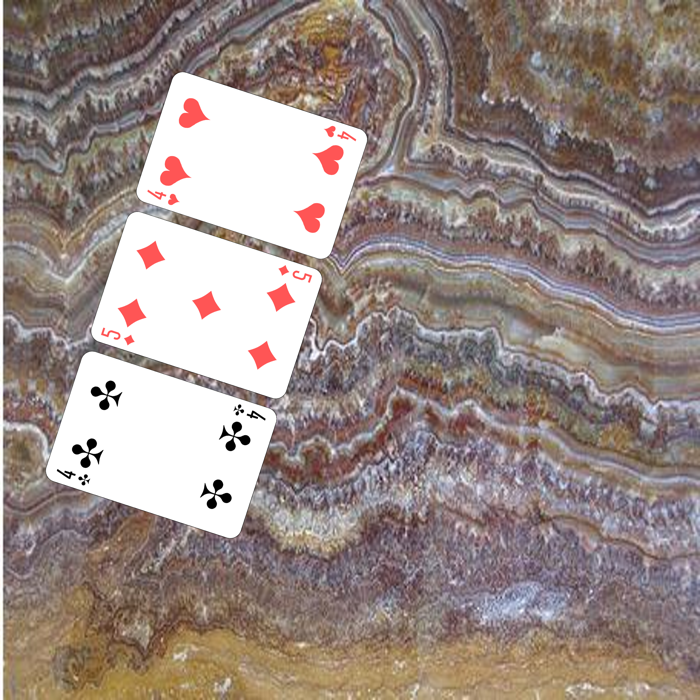}
         \caption{Poker cards\newline}
     \end{subfigure}
     \hfill
     \begin{subfigure}[b]{0.13\textwidth}
         \centering
         \includegraphics[width=\textwidth]{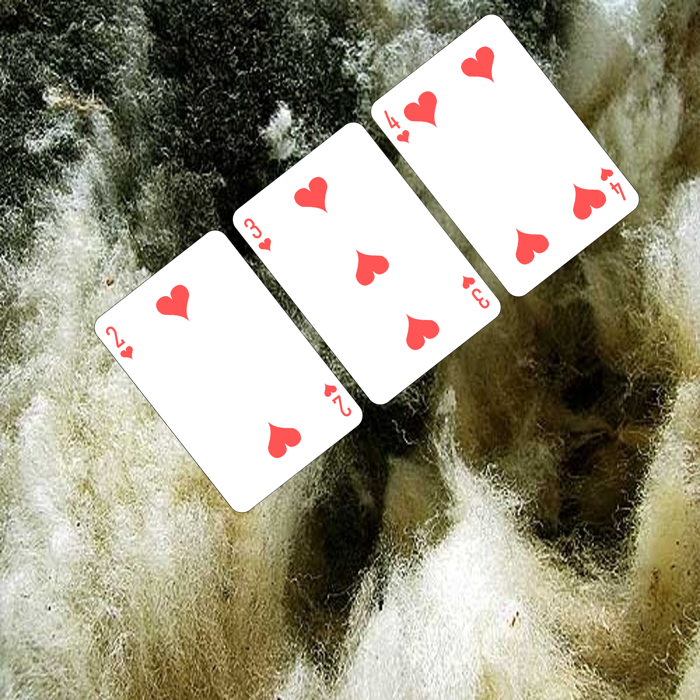}
         \caption{Class-level poker cards}
     \end{subfigure}
     \hfill
     \vspace{0.3cm}
     
        \caption{Samples from the playing cards dataset}
        \label{fig:main_example_playing_cards}
\vspace{0.5cm}
\end{figure}

\textbf{Playing cards} is a synthetic image dataset we introduce to analyse CBMs free of inaccurate concept annotations. The dataset consists of multiple image variations, of which we use three in this paper: \textit{Random cards}, \textit{poker cards} and \textit{class-level poker cards}. Example samples can be seen in Figure~\ref{fig:main_example_playing_cards}. Each variation consists of 10,000 images where concepts represent playing cards, and with the task of classifying hand ranks in the game Three Card Poker. Concepts for random cards are selected at random from the full 52 possible playing cards for each image to ensure there is no correlation between which concepts occur together, but this results in an uneven occurrences of task classes. For instance, the class ``high card'' has 5191 training samples while the class ``straight flush'' has 20 training samples. For poker cards task classes are balanced with concepts selected from the sets of triplets available based on the class used in each sample, resulting in some concepts appearing together more often than others. For the most extreme, there are 48 unique triplets of concepts for the class ``straight flush'' while this class has 1166 training samples. Class-level poker cards has the same task classes as random cards and poker cards, but only using 11 concepts instead of 52. This means some concepts are only used for one task class while others are used for many. We have listed the full concepts and relations to task classes in Table~\ref{tab:class-level poker task concepts}. Each image variation has a 70\%-30\% split between training and validation images. Random cards and poker cards use instance-level concepts while class-level poker cards use class-level concepts. In all cases, if a concept is annotated as present, then it is visible in the corresponding image. The dataset is available for download.\footnote{Playing cards dataset: Available for camera-ready submission}

% \url{https://huggingface.co/datasets/JackFurby/playing-cards}

\begin{table}[ht!]
\caption{Concepts for class-level poker cards are arranged such that some are used for one task class while others are used for many task classes.}
\centering
\hspace{8mm}
\begin{tabular}{ll}
\toprule
Task label & Concepts \\
\midrule
Straight Flush & 2 of $\heartsuit$, 3 of $\heartsuit$, and 4 of $\heartsuit$ \\
Three of a Kind & 4 of $\clubsuit$, 4 of $\diamondsuit$, and 4 of $\spadesuit$ \\
Straight & 3 of $\heartsuit$, 4 of $\clubsuit$, and 5 of $\diamondsuit$ \\
Flush & 4 of $\diamondsuit$, 6 of $\diamondsuit$, and 9 of $\diamondsuit$ \\
Pair & 5 of $\clubsuit$, 5 of $\diamondsuit$, and 10 of $\heartsuit$ \\
High Card & 4 of $\spadesuit$, 5 of $\diamondsuit$, and 10 of $\heartsuit$ \\
\bottomrule
\end{tabular}
\label{tab:class-level poker task concepts}
\vspace{0.5cm}
\end{table}

\begin{figure}[ht!]
\centering
    \begin{tabular}{m{0.15\textwidth} m{0.28\textwidth} }
        Input &
        Concepts
     \\
     \midrule
        \frame{\includegraphics[width=0.15\textwidth]{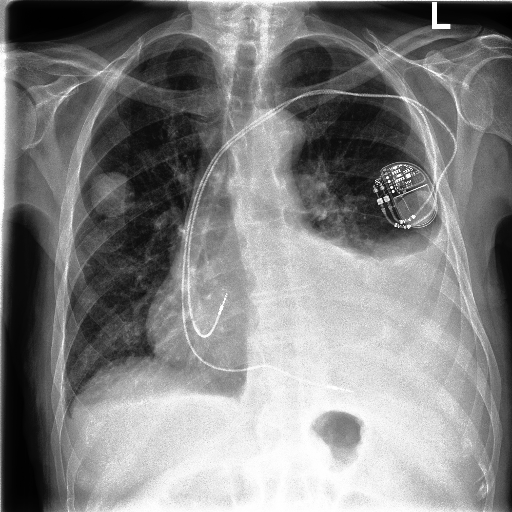}} &
        Enlarged cardiomediastinum: present \newline
        Cardiomegaly: present \newline
        Lung opacity: present \newline
        Lung lesion: present \newline
        Edema: not present \newline
        Consolidation: not present \newline
        Pneumonia: not present \newline
        Atelectasis: present \newline
        Pneumothorax: not present \newline
        Pleural effusion: present \newline
        Plaural other: not present \newline
        Fracture: not present \newline
        Support devices: present
    \end{tabular}

    \caption{Example CheXpert sample with concept annotations}
    \label{fig:example_chexpert_sample}
    \vspace{0.2cm}
\end{figure}

\textbf{CheXpert \citep{chexpert}} is a real-world image dataset with visually represented observations. Each sample has 14 observations such as ``fracture'' and ``edema'', of which 12 are pathologies. The other two observations are ``support devices'', e.g. a pacemaker, and ``no\_findings''. We use 13 of these observations as concepts while the 14th, ``no\_findings'', as the task label. ``no\_findings'' is positive if all pathologies are not annotated as present. We have provided an example sample with concept annotation in Figure~\ref{fig:example_chexpert_sample}. CheXpert has instance-level concepts and contains 224,316 chest X-ray images that were scaled to 512 x 512 pixels. We use the official dataset splits from \citep{chexpert} which has 223,414 training images, 234 validation images and 668 test images. Training annotations were automatically generated from radiology reports and were labelled as 1 when an observation was confidently present, 0 when an observation was confidently not present, and -1 when uncertain. To translate this to binary labels we used U-ones annotations which set any missing values to 0 and any uncertain annotations to 1. Validation images were labelled by three board-certified radiologists while test images were labeled by eight board-certified radiologists. Validation and test images only include binary annotations. We also created a modified version of the dataset to use class-level concepts with the most common concept vector for three, four and five concepts present. Class-level CheXpert has 44,974 samples, 21,760 samples, and 636 samples for three, four and five concepts present respectively. We refer to the original version of CheXpert as \textit{instance-level CheXpert} and the modified version as \textit{class-level CheXpert}.

\subsection{Models}

All of our models use a similar structure where the concept encoder receives an input and outputs a concept vector with one value for each concept. This is followed by a sigmoid function which receives the concept vector as an input and outputs concept predictions. This increases the independence of concept representations \citep{zarlenga2023towards}. Finally, this is followed by the task predictor which receives concept predictions as an input and outputs the task prediction. We use Binary Cross Entropy loss to train the concept encoder, and Cross Entropy loss for the task predictor. Concept accuracy is the average binary accuracy of concept predictions with a 0.5 threshold. We repeated training 5 times for each dataset.

The playing cards models use a VGG-11 architecture with batch normalisation \citep{Simonyan15} for the concept encoder and two linear layers with a ReLU activation function for the task predictor. We trained these models to minimise the concept and task loss. Our random cards models achieve an average concept accuracy of 99.932\%, poker card models have an average concept accuracy of 99.914\% and class-level poker cards have an average concept accuracy of 99.99\%. To compare CBM with a standard Neural Network (NN) we also trained models on poker cards using the same model architecture, but without concept loss, achieving an average concept accuracy of 50.25\%.

Our CheXpert models use a Densenet121 architecture \citep{densenet} for the concept encoder which is initialised with pre-trained weights from ImageNet and two linear layers with a ReLU activation function for the task predictor which is not pre-trained. We trained these models to maximise the Area Under the receiver operating characteristic Curve (AUC) of concept predictions, following previous work \citep{ye2020weakly,51875}, and minimise the task loss. Our instance-level CheXpert models achieve an average concept accuracy of 75.23\% while class-level CheXpert models archived an average concept accuracy of 57.826\%, 62.143\% and 63.615\% for the dataset version with three, four and five concepts present respectively.

Full details on model training and performance for playing card models and CheXpert models can be found in the Appendix~\ref{model_training_appendix}.

\begin{figure}[ht!]
%sample: 2504
\centering
\begin{tabular}{b{0.1\textwidth} m{0.09\textwidth} m{0.09\textwidth} m{0.09\textwidth} }
    &
    Input &
    &
 \\
    &
    \frame{\includegraphics[width=0.1\textwidth]{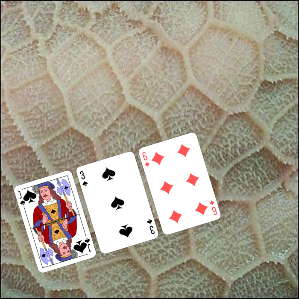}} & 
    &
 \\
    &
    Jack of Spades &  
    Three of Spades &  
    Six of\newline Diamonds
 \\
    \rotatebox[origin=m]{90}{\parbox{0.1\textwidth}{Independent and\newline Sequential - Random cards}} &
    \frame{\includegraphics[width=0.1\textwidth]{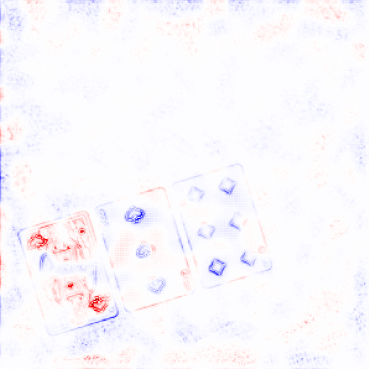}} &  
    \frame{\includegraphics[width=0.1\textwidth]{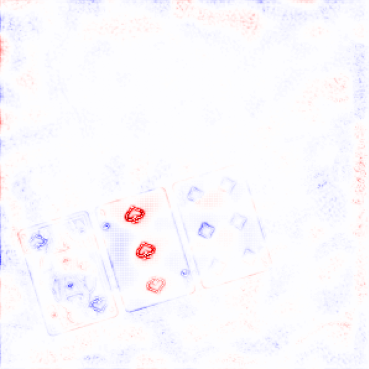}} &  
    \frame{\includegraphics[width=0.1\textwidth]{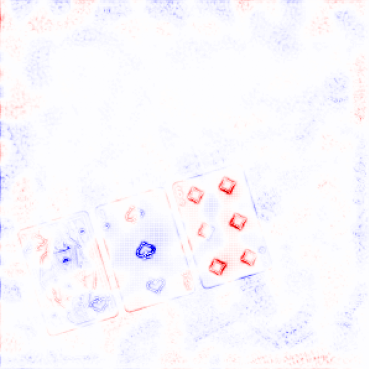}} 
 \\
    \midrule
    \rotatebox[origin=m]{90}{\parbox{0.1\textwidth}{Independent and\newline Sequential - Poker cards}} &
    \frame{\includegraphics[width=0.1\textwidth]{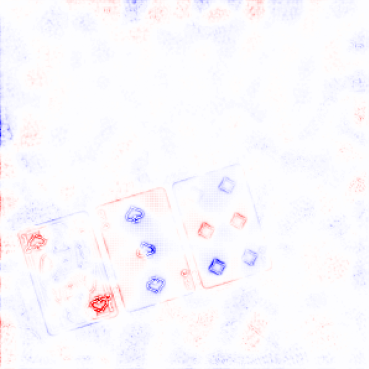}} &  
    \frame{\includegraphics[width=0.1\textwidth]{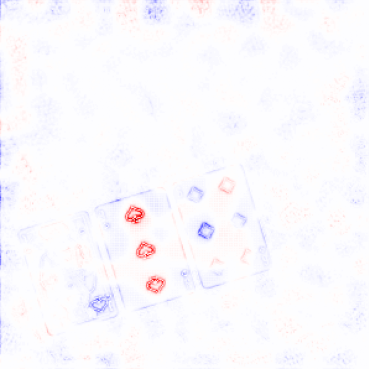}} &  
    \frame{\includegraphics[width=0.1\textwidth]{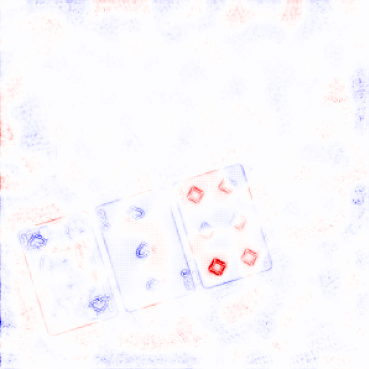}}
 \\
    \midrule
    \rotatebox[origin=m]{90}{\parbox{0.1\textwidth}{Joint - Poker cards}} &  
    \frame{\includegraphics[width=0.1\textwidth]{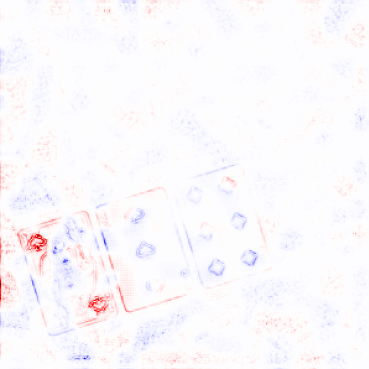}} &  
    \frame{\includegraphics[width=0.1\textwidth]{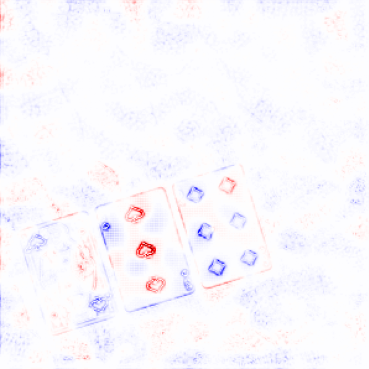}} &  
    \frame{\includegraphics[width=0.1\textwidth]{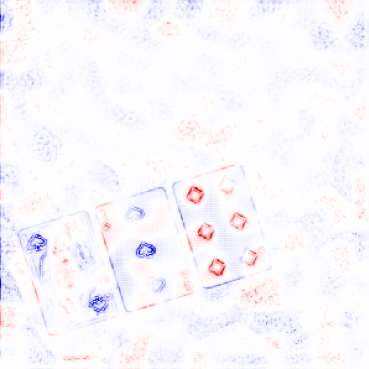}} 
 \\
    \midrule
    \rotatebox[origin=m]{90}{\parbox{0.1\textwidth}{Standard NN}} & 
    \frame{\includegraphics[width=0.1\textwidth]{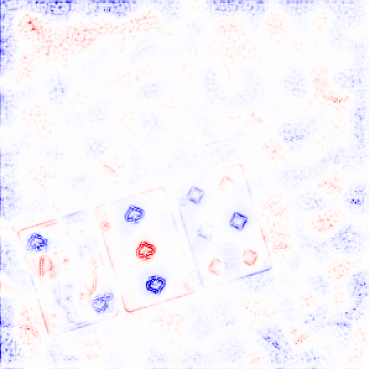}} &  
    \frame{\includegraphics[width=0.1\textwidth]{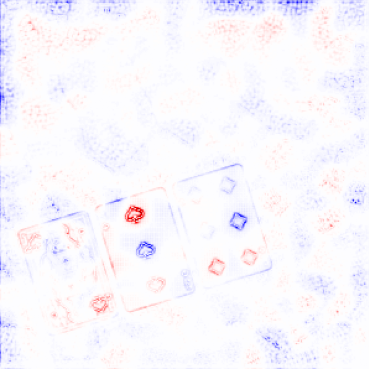}} &  
    \frame{\includegraphics[width=0.1\textwidth]{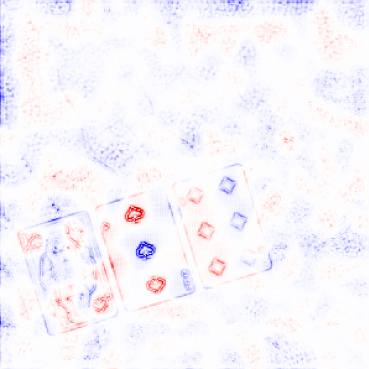}}
\end{tabular}

\caption{Positive relevance (red) is always applied to the expected input features (playing cards) when concept supervision is used in training.}
\label{fig:CBM models concept saliency}
\vspace{0.5cm}
\end{figure}

\section{Experiments and Results} \label{results_section}

\subsection{Input Feature Importance}

We first evaluate feature importance for concept predictions and our playing card models. We use model-specific relevance techniques that utilise the model architecture to produce explanations by propagating a gradient backwards through the model \citep{10.1371/journal.pone.0130140,10.5555/3305890.3306024,gradCAM}. Each of the explanation techniques produces saliency maps where we display positive relevance in red and negative relevance in blue.

We use the technique LRP to explain concept prediction from our playing card models as relevance is attributed to groups of input features instead of on a pixel-by-pixel basis \citep{9369420} when using the rules from \citep{Montavon2019}. LRP rules are ways to control how relevancy is propagated from neuron to neuron in the NN \citep{10.1371/journal.pone.0130140}.

\begin{figure}[ht!]
%sample: 501
\begin{subfigure}[b]{0.5\textwidth}

    \centering
    \begin{tabular}{m{0.2\textwidth} m{0.2\textwidth} m{0.2\textwidth} m{0.2\textwidth} }
        Input &
        Four of Clubs &
        Four of\newline Diamonds &
        Four of Spades
     \\
        \frame{\includegraphics[width=0.18\textwidth]{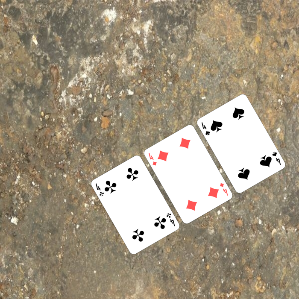}} &
        \frame{\includegraphics[width=0.18\textwidth]{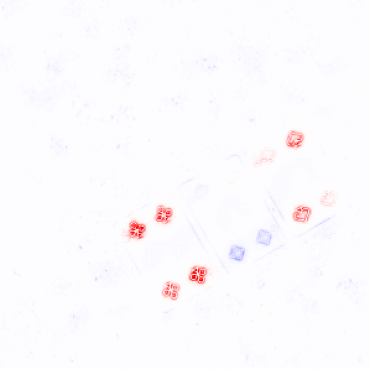}} & 
        \frame{\includegraphics[width=0.18\textwidth]{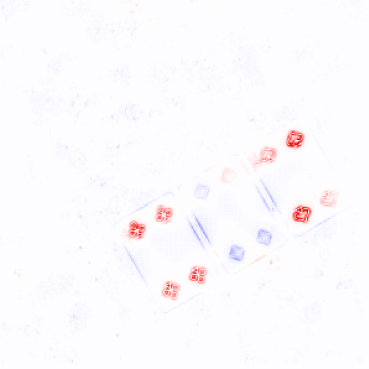}} &
        \frame{\includegraphics[width=0.18\textwidth]{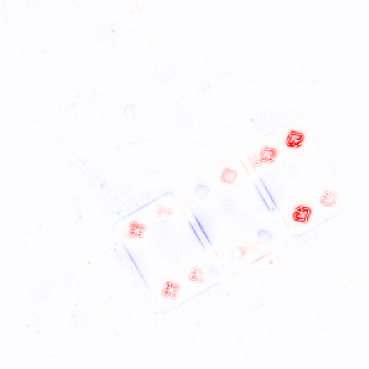}}
    \end{tabular}

    \caption{task class: Three of a kind}
    \label{fig:CBM class-level concept saliency three-of-a-kind}

    \vspace{0.5cm}
    
\end{subfigure}

\begin{subfigure}[b]{0.5\textwidth}

    \centering
    \begin{tabular}{m{0.2\textwidth} m{0.18\textwidth} m{0.2\textwidth} m{0.2\textwidth} }
        Input &
        Five of\newline Diamonds &
        Five of Clubs &
        Ten of Hearts
     \\
        \frame{\includegraphics[width=0.18\textwidth]{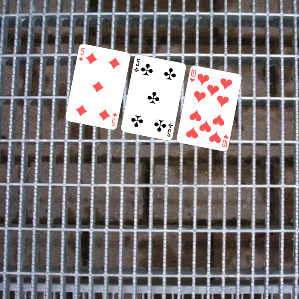}} &
        \frame{\includegraphics[width=0.18\textwidth]{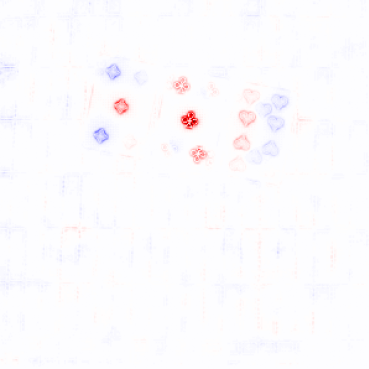}} & 
        \frame{\includegraphics[width=0.18\textwidth]{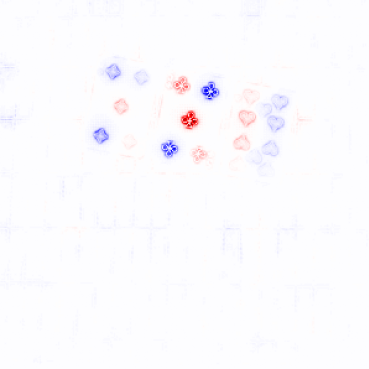}} &
        \frame{\includegraphics[width=0.18\textwidth]{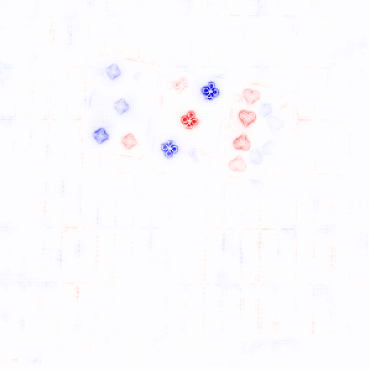}}
    \end{tabular}

    \caption{task class: Pair}
    \label{fig:CBM class-level concept saliency pair}

    \vspace{0.5cm}
    
\end{subfigure}

\caption{Positive saliency (red) does not always map to the semantically meaningful input features for each concept for class-level poker cards.}
\label{fig:CBM class-level concept saliency}
\vspace{0.5cm}
\end{figure}

Figure~\ref{fig:CBM models concept saliency} shows for random and poker card CBMs the most positively relevant input features are symbols on the playing cards corresponding to each concept with negative relevance distributed over the other playing cards. As we do not specify which input features the model should use for each concept, and we can see the models have selected features within the boundaries of each specified playing cards, we consider these reasonable input features for the model to use. Our standard NN was unable to localise relevance to individual playing cards and instead distributed relevance over all three cards present. In Figure~\ref{fig:CBM class-level concept saliency} we show saliency maps for models trained with class-level poker cards. Similar to \citep{furby2023towards,margeloiu2021concept} most concepts using our dataset did not have relevance applied to semantically meaningful input features. A few concepts, however, were an exception. Namely the concepts ``Four of Clubs'' and ``Four of Spades'' as seen in Figure~\ref{fig:CBM class-level concept saliency three-of-a-kind}. As class-level poker cards assigns some concepts to a single task, while others appear for many, the input features the model should use for each concept prediction may be ambiguous. For instance, the concepts Two, Three and Four of Hearts always appear together and therefore the model has no way of separating the pixels for one concept from another, while the concept ``Four of Clubs'' may appear with the ``Four of Diamond'' and ``Four of spades'', or with the ``Three of Hearts'' or ``Five of Diamonds''. In this case, the model has a far better chance of learning the semantically meaningful input features.

To quantitatively analyse if our playing card models are applying relevance to semantically meaningful input features, we can measure the proportion of relevance applied to concepts' visual representations in comparison to the total relevance applied to all concept input features. For playing cards this is the relevance applied to one playing cards compared to all three playing cards in a given sample. If the proportion of positive relevance is high, and the proportion of negative relevance is low, then the model has learned a semantically meaningful concept mapping. We compare our independent/sequential models trained on random cards, poker cards, class-level poker cards, and our standard NN models with averaged proportions from the 5 training repeats for each dataset variant. Each point represents a single concept. 

\begin{figure}[ht!]
    \centering
    \includegraphics[width=0.45\textwidth]{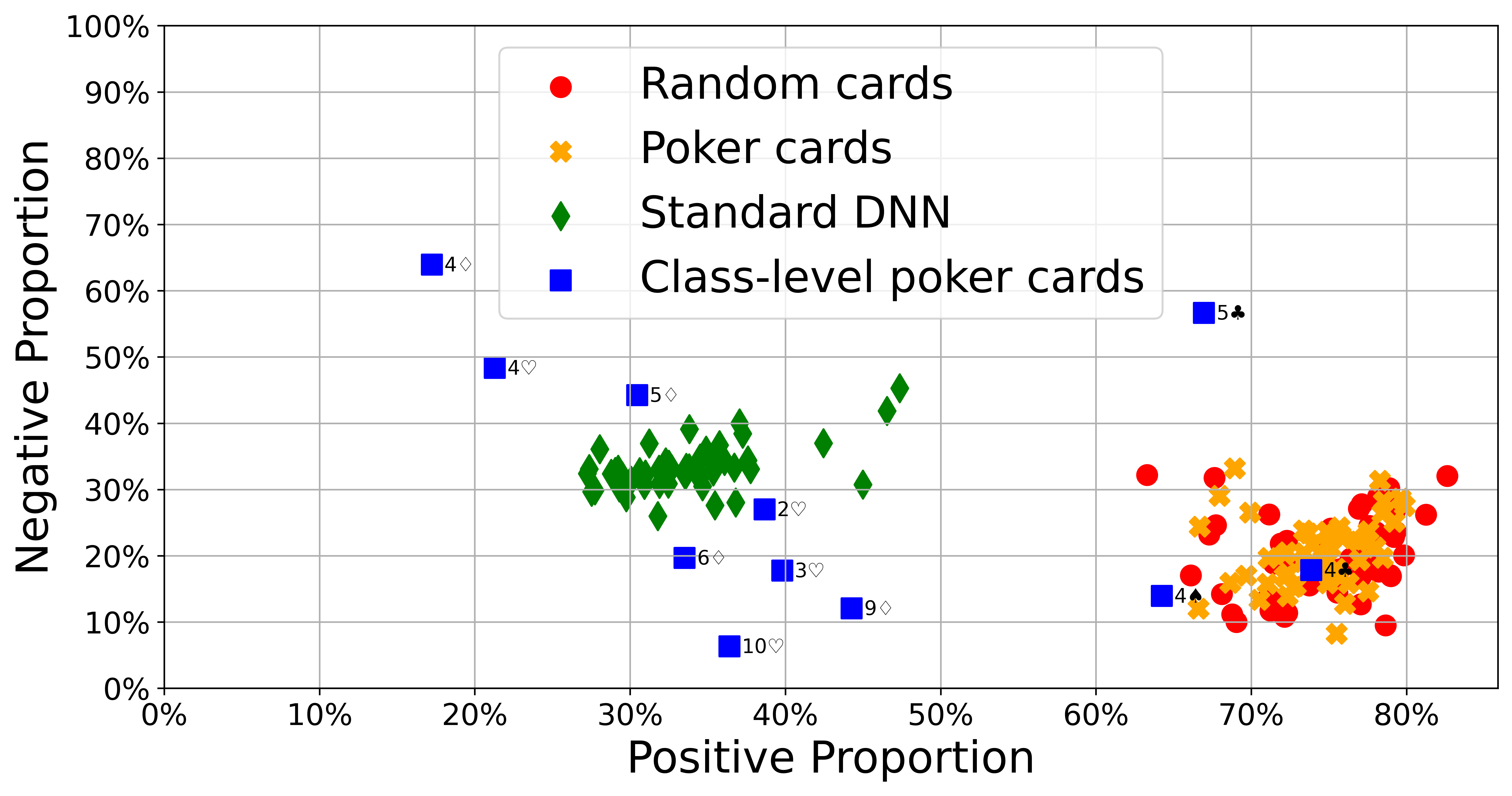}
    \caption{Random cards and poker cards both have a high positive proportion of relevance, indicating the models have learned to use semantically meaningful input features to predict concepts, unlike standard NN and most concepts for class-level poker cards.}
    \label{fig:relevancy proportion}
\vspace{0.5cm}
\end{figure}

The plot in Figure~\ref{fig:relevancy proportion} shows two distinct clusters, one for the standard NN and one for both random cards and poker cards. Random cards and poker cards cluster has the highest proportion of positive relevance with most points being between 70\% and 80\% and the lowest proportion of negative relevance where most concepts fall between 10\% and 30\%. The standard NN has far less positive relevance and slightly more negative relevance, both around 30\% to 40\%. Combining this plot with the saliency maps we saw in Figure~\ref{fig:CBM models concept saliency} confirms the CBMs have learned to apply relevance to semantically meaningful input features for both random cards and poker cards.

The points for class-level poker cards are not clustered together. Most concepts have a low proportion of relevance, meaning relevance is not applied to semantically meaningful input features. A few concepts however, ``Four of Spades'', ``Four of Clubs'' and ``Five of Clubs'' have a high proportion of positive relevance. For the concepts ``Four of Spades'' and ``Four of Clubs'', this confirms what we saw in Figure~\ref{fig:CBM class-level concept saliency three-of-a-kind}, that a semantically meaningful concept mapping has been learned. The same cannot be said for the concept ``Five of Clubs'' as other concepts apply positive relevancy to the pixels representing the concept ``Five of Clubs'' such as the concepts ``Five of Diamonds'' and ``Ten of Hearts'' as seen in Figure~\ref{fig:CBM class-level concept saliency pair}, inflating the positive proportion of relevance seen. Class-level poker cards reveal the challenge of creating a dataset with enough constraints for the model to learn semantically meaningful concept mappings. Even though the concepts ``Four of Spades'' and ``Four of Clubs'' show it possible for a CBM to learn semantically meaningful concept mappings with class-level concepts, the consistency in relevance proportions with random cards and poker cards shows the advantage instance-level concepts can provide.

\begin{figure*}[ht!]

%sample: 220

    \centering
    \begin{subfigure}[b]{\textwidth}
        \centering
        \begin{tabular}{m{0.14\textwidth} m{0.14\textwidth} m{0.14\textwidth} m{0.14\textwidth} m{0.14\textwidth} }
            Pleural effusion &
            Atelectasis &
            Cardiomegaly &  
            Support devices &  
            Lung opacity
         \\
            \frame{\includegraphics[width=0.11\textwidth]{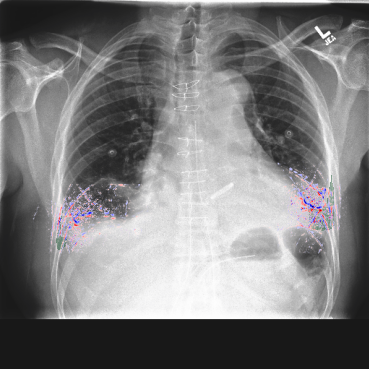}} &  
            \frame{\includegraphics[width=0.11\textwidth]{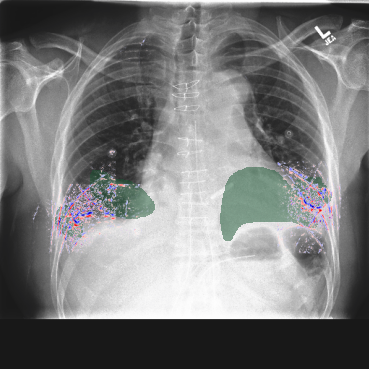}} &  
            \frame{\includegraphics[width=0.11\textwidth]{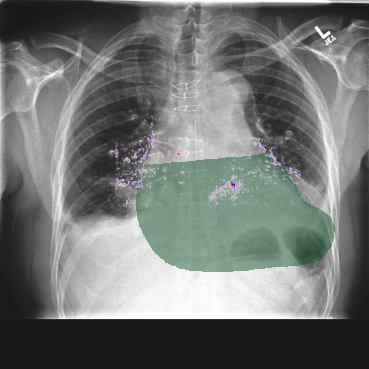}} &  
            \frame{\includegraphics[width=0.11\textwidth]{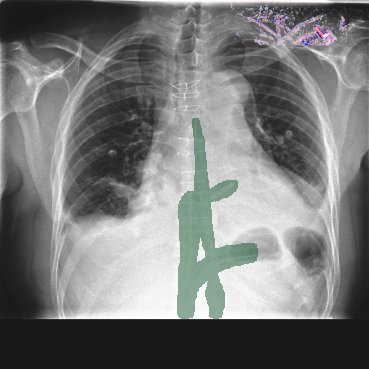}} &  
            \frame{\includegraphics[width=0.11\textwidth]{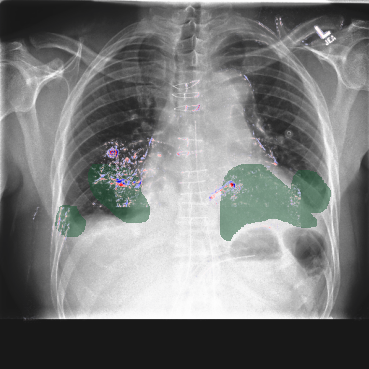}}
         \\
            0.99666 &
            0.72508 &
            0.75666 &  
            0.21335 &  
            0.57415
         \end{tabular}
        \caption{Sequentially trained model}
        \label{fig:chexpert_saliency_sequential_12}
        \vspace{0.5cm}
     \end{subfigure}
     \begin{subfigure}[b]{\textwidth}
        \centering
        \begin{tabular}{m{0.14\textwidth} m{0.14\textwidth} m{0.14\textwidth} m{0.14\textwidth} m{0.14\textwidth} }
            Pleural effusion &
            Atelectasis &
            Cardiomegaly &  
            Support devices &  
            Lung opacity
         \\
            \frame{\includegraphics[width=0.11\textwidth]{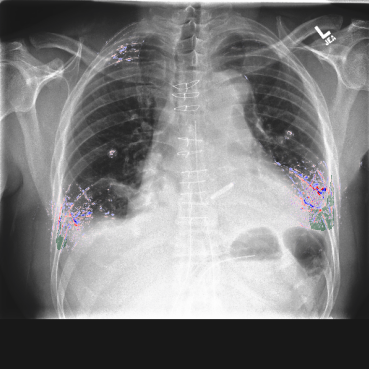}} &  
            \frame{\includegraphics[width=0.11\textwidth]{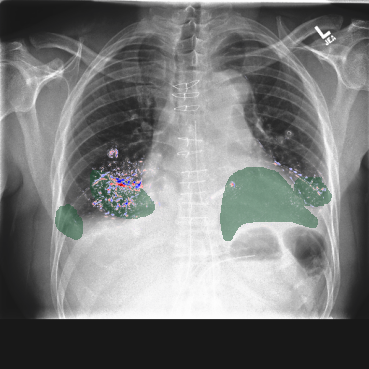}} &  
            \frame{\includegraphics[width=0.11\textwidth]{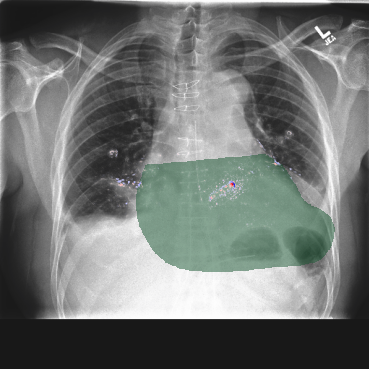}} &  
            \frame{\includegraphics[width=0.11\textwidth]{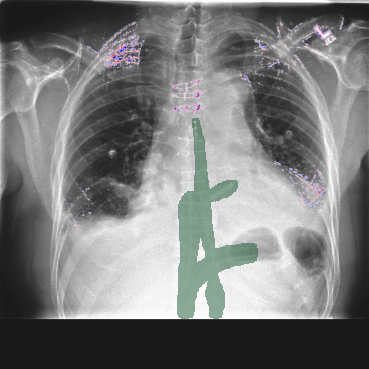}} &  
            \frame{\includegraphics[width=0.11\textwidth]{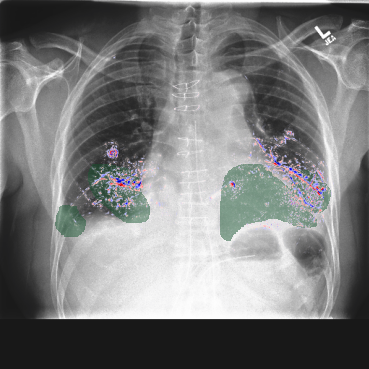}}
         \\
            0.98684 &
            0.78311 &
            0.64232 &  
            0.21329 &  
            0.55349
        \end{tabular}
        \caption{Joint trained model}
        \label{fig:chexpert_saliency_joint_12}
        \vspace{0.5cm}
     \end{subfigure}
        \caption{Concept saliency maps for chest X-rays with instance-level CheXpert shows reasonable localisation of concepts to ground truth areas of the input image. The number beneath each saliency map is the concept prediction made by the model.}
        \label{fig:CBM chexpert concept saliency}
\vspace{0.5cm}
\end{figure*}

\begin{figure}[ht!]

\begin{subfigure}[b]{0.5\textwidth}

    \centering
    \begin{tabular}{m{0.2\textwidth} m{0.2\textwidth} m{0.2\textwidth} }
        Lung\newline Opacity &
        Pleural\newline Effusion &
        Support\newline Devices
     \\
        \frame{\includegraphics[width=0.21\textwidth]{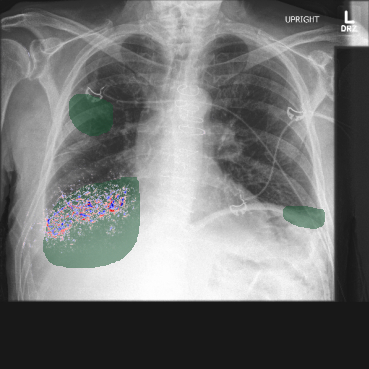}} &
        \frame{\includegraphics[width=0.21\textwidth]{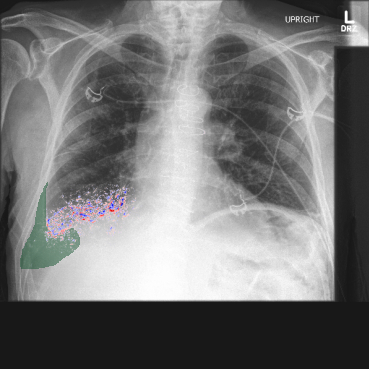}} & 
        \frame{\includegraphics[width=0.21\textwidth]{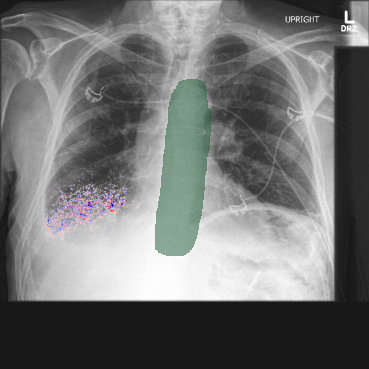}}
     \\
     0.86693 &  
     0.8637 &  
     0.86781  
    \end{tabular}

    \caption{Class-level chexpert trained with three concepts annotated as present}
    \label{fig:CBM class-level chexpert saliency group-0}
    \vspace{0.5cm}
    
\end{subfigure}

\begin{subfigure}[b]{0.5\textwidth}

    \centering
    \begin{tabular}{m{0.2\textwidth} m{0.2\textwidth} m{0.2\textwidth} }
        Edema &
        Pleural\newline Effusion &
        Support\newline Devices
     \\
        \frame{\includegraphics[width=0.21\textwidth]{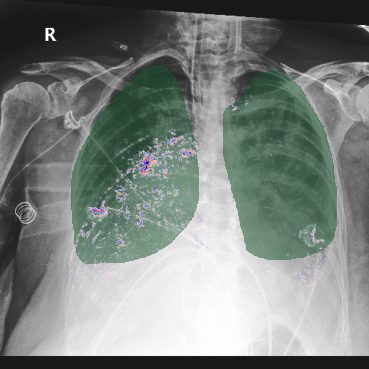}} &
        \frame{\includegraphics[width=0.21\textwidth]{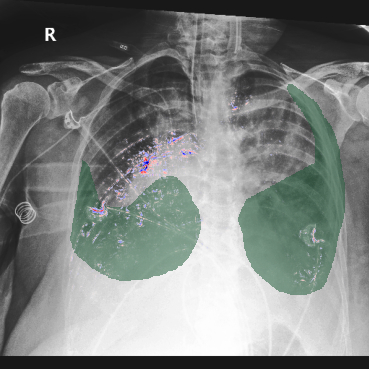}} & 
        \frame{\includegraphics[width=0.21\textwidth]{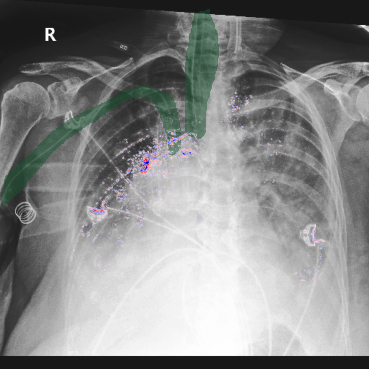}}
     \\
     0.98301 &  
     0.98212 &  
     0.9829  
    \end{tabular}

    \caption{Class-level chexpert trained with four concepts annotated as present}
    \label{fig:CBM class-level chexpert saliency group-1}
    \vspace{0.5cm}
    
\end{subfigure}

\begin{subfigure}[b]{0.5\textwidth}

    \centering
    \begin{tabular}{m{0.2\textwidth} m{0.2\textwidth} m{0.2\textwidth} }
        Consolidation &
        Pleural\newline Effusion &
        Support\newline Devices
     \\
        \frame{\includegraphics[width=0.21\textwidth]{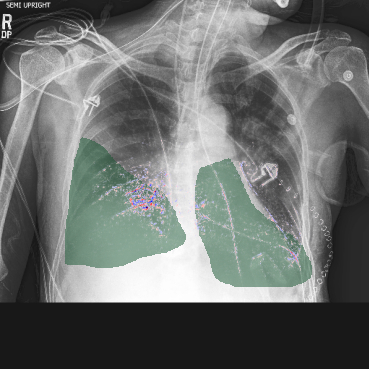}} &
        \frame{\includegraphics[width=0.21\textwidth]{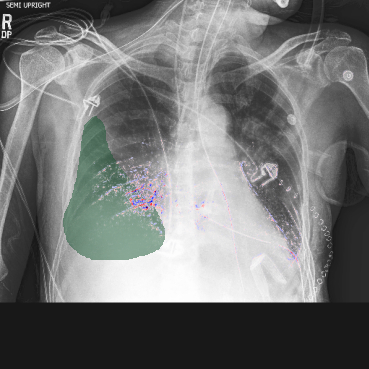}} & 
        \frame{\includegraphics[width=0.21\textwidth]{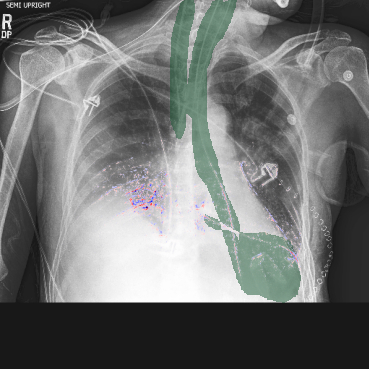}}
     \\
     0.94323 &  
     0.94379 &  
     0.94084  
    \end{tabular}

    \caption{Class-level chexpert trained with five concepts annotated as present}
    \label{fig:CBM class-level chexpert saliency group-2}
    \vspace{0.5cm}
    
\end{subfigure}

\caption{Concept saliency maps for class-level CheXpert shows the models are using similar input features to predict all present concepts for a given input image. The number beneath each saliency map is the concept prediction made by the model.}
\label{fig:CBM class-level chexpert saliency}
\vspace{0.5cm}
\end{figure}

To evaluate CBMs on a more challenging dataset, we have trained CBMs on CheXpert. This dataset contained chest X-ray images with concepts representing visual observations. As uncertain or missing values in the dataset are set to present, some concept annotations will be inaccurate and we may assume there are some additional inaccurate concept annotations caused by the annotations originally being generated using an automated labeller \citep{chexpert}. For our results in Figure~\ref{fig:CBM chexpert concept saliency}, we are using our models trained on instance-level CheXpert and the saliency mapping technique Guided Grad-CAM \citep{gradCAM} as Grad-CAM techniques have been shown to outperform other techniques with this dataset \citep{cheXlocalize}. To validate that the models have mapped concepts to intended input features we are using ground truth segmentations from \citep{cheXlocalize}. These were created by two board-certified radiologists and ensure our conclusions are made w.r.t. expert opinion. Our results show concepts trained on instance-level CheXpert can map concepts to semantic input features for models trained with both the independent/sequential method in Figure~\ref{fig:chexpert_saliency_sequential_12} and joint method in Figure~\ref{fig:chexpert_saliency_joint_12}. The concepts for ``lung opacity'', ``atelectasis'' and ``pleural effusion'' all should be observable with observations in the lung, ``cardiomegaly'' observed by an enlarged heart, and ``support devices'' by the observation of an object that is not part of the body (e.g. a pacemaker). From the samples we show, most concepts map to features within the ground truth segmentation such as with the saliency maps for the concepts ``atelectasis'' and ``pleural effusion''. The concept ``cardiomegaly'' is localised to a portion of the segmentation, while ``support devices'' missed the ground truth segmentation. In the case of ``support devices'', the model may have missed the semantic input features as they are hard to spot in this sample compared to a pacemaker which would also be annotated as the same concept. It's also worth pointing out the concept ``support devices'' was not predicted as present as seen by the value beneath the saliency map being lower than 0.5. The main takeaway from Figure~\ref{fig:CBM chexpert concept saliency} is the saliency maps for Instance-level CheXpert are distinctly different to each other and do not appear to be using the same input features for every concept prediction, unlike class-level CheXpert as we see in Figure~\ref{fig:CBM class-level chexpert saliency} where all concept saliency maps highlight similar input features irrespective of the concept being predicted.

\begin{figure}[ht!]
\centering
\includegraphics[width=0.45\textwidth]{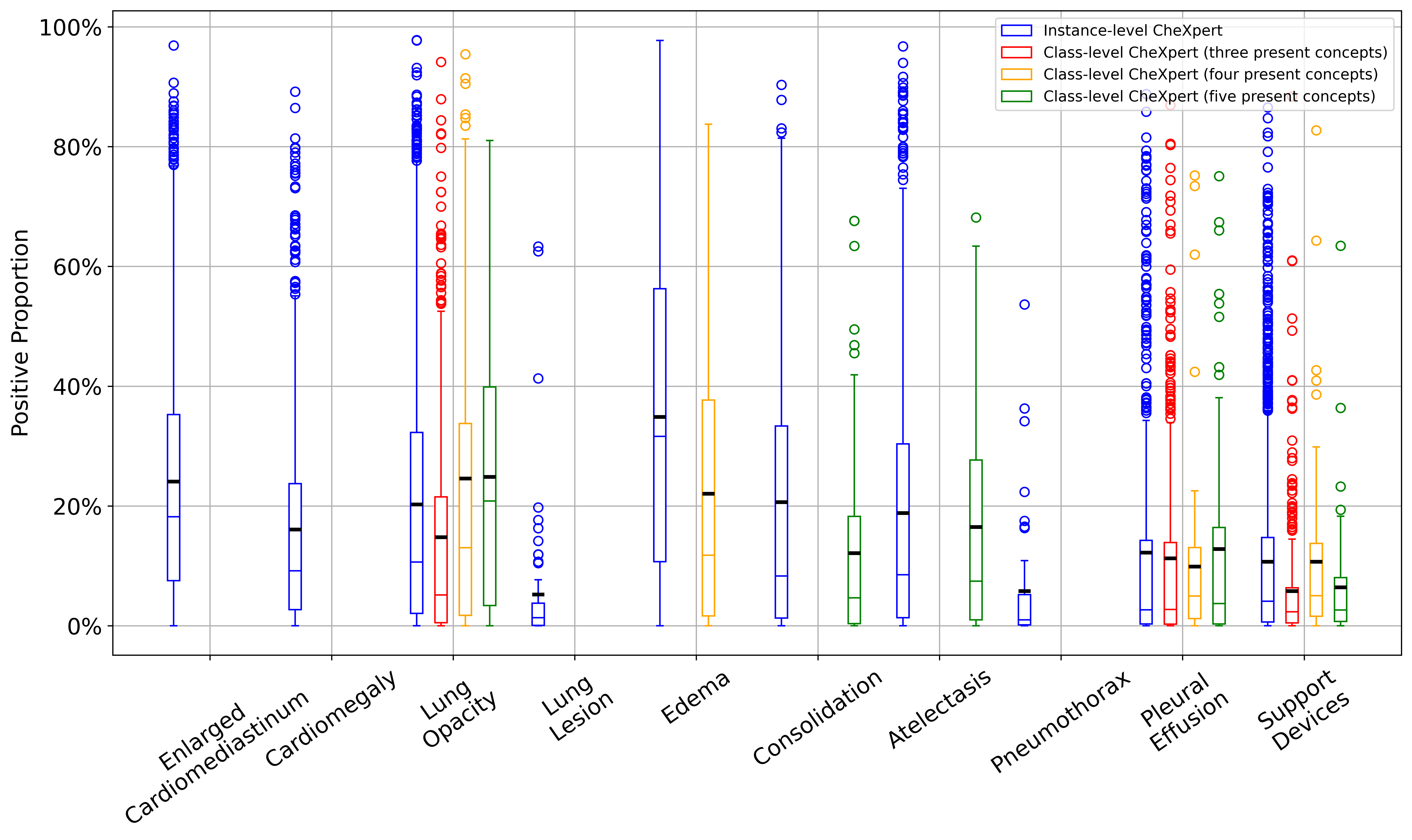}
\caption{Instance-level CheXpert has more positive relevancy attributed to ground truth segmentations than class-level CheXpert for most concepts.}
\label{fig:relevancy proportion chexpert}
\vspace{0.5cm}
\end{figure}

The proportion of relevance using our sequentially trained CheXpert models is shown in Figure~\ref{fig:relevancy proportion chexpert} which compares positive relevance applied to the ground truth segmentations to the entire image. As the proportion of positive and negative relevance was almost the same we have excluded negative relevance proportions from this plot. The averaged maximum positive proportion of relevance, the black lines in the figure, tops out just under 40\%. Instance-level CheXpert outperforms class-level CheXpert at applying relevance to input features in ground truth segmentations for some concepts such as ``consolidation'' and ``atelectasis'', while class-level CheXpert outperforms instance-level CheXpert for other concepts such as ``lung opacity''. For the cases where class-level CheXpert outperforms instance-level CheXpert, we can attribute this to if the model learning the same input features to accurately predict multiple concepts then this will inflate the relevance proportion for some concepts, but harm the proportion of relevance for other concepts. For instance, if the model can predict all concepts using the pixels that represent the concept ``lung opacity'' then we would expect the proportion of relevance for that concept to be high in comparison to the other concepts if the ground truth segmentation's are in a different region of the chest. This is overall what we see and as the saliency maps for class-level CheXpert models confirm the same input features are used to predict all present concepts in a given sample, it is clear the class-level CheXpert dataset did not constrain our CBMs to learn semantically meaningful input features to predict concepts. Although instance-level CheXpert cannot consistently predict concepts using semantically meaningful input features, because class-level CheXpert uses the same input features for all concept prediction, we can still conclude a dataset should include a clear link between input features and concept annotations.

\subsection{Concept Purity}

If we can train CBMs to learn semantically meaningful input features we may also expect learned concepts to be disentangled such that one concept cannot be used to predict another concept if there is no correlation between them. We can measure this using the Oracle Impurity Score (OIS) \citep{zarlenga2023towards} which is a measurement of whether a learned concept representation has the predictive power to predict other concepts compared to the expected predictability from ground truth labels. This score is defined as the concept purity, and is the divergence of two matrices; the predictability of ground truth concepts w.r.t. one another, the oracle matrix, and the predictability of ground truth concepts from learned concepts, the purity matrix. These matrices are created by training a series of helper models which receive either ground truth concepts or predicted concepts and minimise the loss of predicting ground truth concepts. Our configuration uses helper models formed of a two-layer ReLU multi-layer perceptron with 32 activations in the hidden layer. If OIS results in a value of 0 then learned concepts do not encode any more or less information than the ground truth concepts whereas a value of 1 means each learned concept can predict all other concepts.

OIS results for poker cards, random cards (both tested with poker cards), and class-level poker cards are shown in Table~\ref{tab:ois}. Our OIS results are averaged using the 5 training repeats we created for each dataset, with each model weight used to generate an OIS value 3 times for a total of 15 per dataset. Models trained on poker cards show a marginal improvement over random cards with class-level poker cards having the highest OIS value. Because OIS is w.r.t. the expected impurities that exist in the dataset, if several ground truth concepts have a high correlation and the concept representations have the same correlation then the OIS metric will show low impurity. If however the concept representations capture different correlations then the OIS metric will show higher impurities. Random cards having a higher OIS value than Poker cards could show poker card models have encoded inter-concept predictability that exists in the dataset which random cards has not.

\begin{table}[ht!]
\vspace{0.5cm}
\caption{OIS is a metric for measuring additional or lacking information in learned concepts compared to the ground truth concepts. Models trained on poker cards encode the least difference in information from learned concepts compared to ground truth concepts which is closely followed by random cards.}
\vspace{0.2cm}
\centering
\hspace{8mm}
\begin{tabular}{lll}
\toprule
Dataset & OIS & Standard deviation \\
\midrule
Random cards & 0.21 & 0.03 \\
Poker cards & 0.19 & 0.02 \\
Class-level poker cards & 0.26 & 0.02 \\
\bottomrule
\end{tabular}
\label{tab:ois}
\end{table}

\begin{figure*}[ht!]
     \centering
     \begin{subfigure}[b]{0.161\textwidth}
         \centering
         \includegraphics[width=1\textwidth]{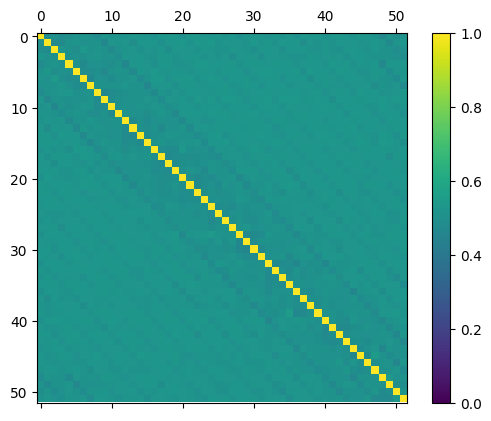}
         \caption{Poker cards oracle matrix}
         \label{fig:poker_oracle_matrix}
     \end{subfigure}
     \hfill
     \begin{subfigure}[b]{0.161\textwidth}
         \centering
         \includegraphics[width=1\textwidth]{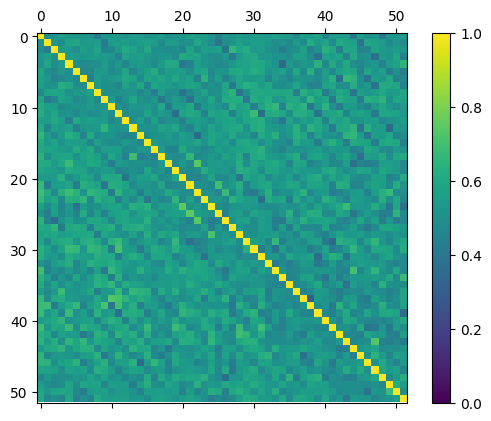}
         \caption{Poker cards purity matrix}
         \label{fig:poker_pred_matrix}
     \end{subfigure}
     \hfill
     \begin{subfigure}[b]{0.161\textwidth}
         \centering
         \includegraphics[width=1\textwidth]{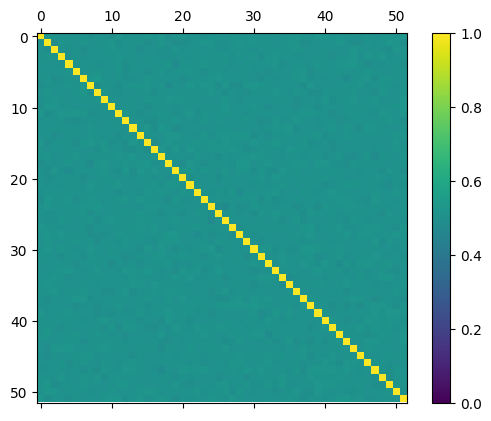}
         \caption{Random cards oracle matrix}
         \label{fig:three_oracle_matrix}
     \end{subfigure}
     \hfill
     \begin{subfigure}[b]{0.161\textwidth}
         \centering
         \includegraphics[width=1\textwidth]{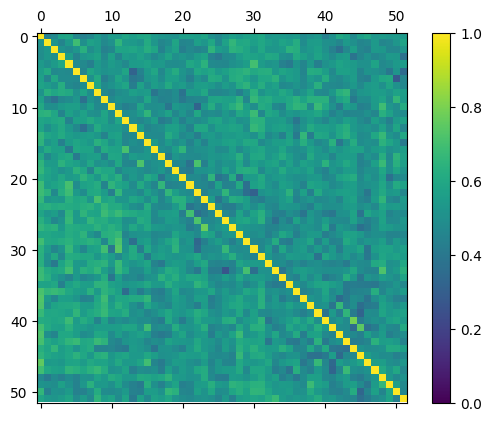}
         \caption{Random cards purity matrix}
         \label{fig:three_pred_matrix}
     \end{subfigure}
     \hfill
     \begin{subfigure}[b]{0.161\textwidth}
         \centering
         \includegraphics[width=1\textwidth]{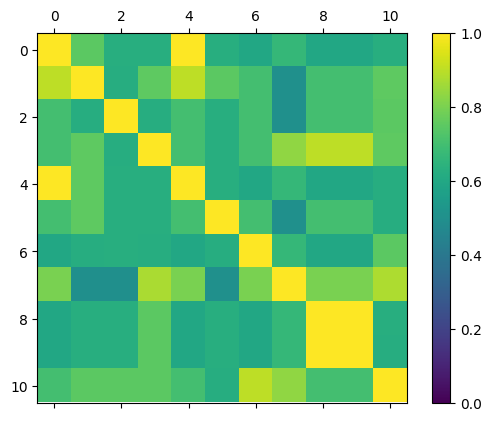}
         \caption{Class-level poker cards oracle matrix}
         \label{fig:class_level_poker_oracle_matrix}
     \end{subfigure}
     \hfill
     \begin{subfigure}[b]{0.161\textwidth}
         \centering
         \includegraphics[width=1\textwidth]{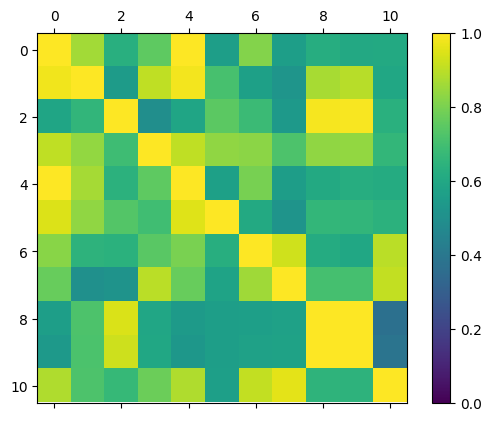}
         \caption{Class-level poker cards purity matrix}
         \label{fig:class_level_poker_pred_matrix}
     \end{subfigure}
        \vspace{0.5cm}
        \caption{Oracle and purity matrices for random cards, poker cards and class-level poker cards. The oracle matrices show inter-concept predictability of ground truth concepts while the purity matrices show inter-concept predictability of learned concepts w.r.t. ground truth concepts. Poker cards show numerous diagonals of non-random inter-concept predictability. Class-level poker cards show a high level of inter-concept predictability for a few concepts within learned concept representations that did not exist in the oracle matrix.}
        \label{fig:oracle_pred_matrix}
    \vspace{0.5cm}
\end{figure*}

We have displayed the two matrices for each dataset and predicted concepts in Figure~\ref{fig:oracle_pred_matrix}. Each coordinate in these matrices shows the AUC value for each concept on the y axis predicting each concept on the x axis. All oracle matrices show a strong predictability for each concept predicting itself, which is to be expected. The oracle matrix for Figure~\ref{fig:poker_oracle_matrix} shows additional diagonals with AUC values lower than 0.5, where concepts in the dataset have information for non-random inter-concept predictability. These diagonals are not present in the random concepts dataset, Figure~\ref{fig:three_oracle_matrix}. Looking at the purity matrices, Figure~\ref{fig:poker_pred_matrix} and Figure~\ref{fig:three_pred_matrix}, the additional diagonals of inter-concept predictability continues for poker cards but there is a distinct reduction with random cards. This confirms the higher OIS value for random cards in Table~\ref{tab:ois} is showing a lack of expected information compared to ground truth concepts for poker cards. As for class-level poker cards, the oracle matrix, Figure~\ref{fig:class_level_poker_oracle_matrix}, shows mostly random inter-concept predictability apart from a few concepts. These concepts are those that are only used for a single task label and as such, these concepts will only appear together. The purity matrix in Figure~\ref{fig:class_level_poker_pred_matrix} has generally higher AUC values which shows the learned concepts are encoding extra information than the ground truth concepts. This is enabled as ground truth concepts are either 0 or 1, but the predicted concept values are between 0 and 1 which allows more information to be encoded. This can be seen by the concept ``Four of Clubs'' (index 2) being able to accurately predict the presence of the concepts Six and Nine of Diamonds (index 8 and 9) in Figure~\ref{fig:class_level_poker_pred_matrix} despite no direct link between them.

The OIS oracle and purity matrices further reinforce the need for a dataset to provide enough constraints such that a CBM cannot learn unintended inter-concept correlations. Essentially, a CBM will learn similar inter-concept correlations as the dataset it is trained on, so to reduce unintended inter-concept correlations we need to ensure their presence in the training data is reduced as much as possible.

\section{Discussion and Conclusion}

In this paper, we realise the original promise of CBMs inherent interpretability by training models where concepts are predicted from semantically meaningful input features. This benefits inherent interpretability as we can be sure the model is predicting concepts for the right reasons. If a CBM makes concept predictions using input features with the same meaning, we can argue the model will be easier to build trust with as concept predictions will use the expected input features from a human perspective. We achieved training CBMs to learn concepts that apply relevance to semantically meaningful input features by training CBMs with accurate concept annotations and with a high variety of concepts that are present together. This differs from the previously analysed CUB dataset with class-level concepts which does not account for visual changes in the input images.

To achieve semantically meaningful concept mappings the following points should be followed:

\begin{itemize}
    \item Concept correlation: Concept annotations should not have any correlation between concepts unless that correlation is intended. As shown by the oracle and purity matrices in Figure~\ref{fig:oracle_pred_matrix}, unintended concept correlation can lead to unrelated concepts accurately predicting each other. It can also obscure which input features are semantically meaningful, as seen in Figure~\ref{fig:CBM class-level concept saliency} and Figure~\ref{fig:CBM class-level chexpert saliency}.
    \item Ensuring concept annotations and visualisations are consistent: Unlike previous studies, we restricted our datasets so we could ensure concepts' visual representations were present in sample images. Following this requirement helps to ensure there is a clear training signal for CBMs to learn semantically meaningful concept representations.
\end{itemize}

We also recommend the use of instance-level concept annotations over class-level concept annotations. Although we show training a CBM to map input features to concepts semantically does not come from the use of class or instance-level concepts. We demonstrate it's far easier to achieve semantically meaningful concept mappings with instance-level concept annotations. In any case, we need to ensure if a concept is annotated as present in a dataset then it is visually present in the corresponding image, and the occurrence of concepts in the dataset does not create unintended correlations. 

Future research regarding the representations CBMs learn should focus on CBMs trained on larger datasets. Although \citep{oikarinen2023labelfree} shows how large language models and CLIP can be used to train an iteration of CBMs, removing the need for manual concept annotations, the semantic mapping from concepts to input features remains unclear. Ensuring concept annotations have a semantic mapping, even for large or machine-generated datasets, or including another constraint to achieve semantically meaningful concept mappings, should be a priority for ensuring inherent interpretability and human trust can be achieved.

% \section*{Acknowledgements}

% This research is funded by the UK Engineering and Physical Sciences Research Council (EPSRC) and IBM UK via an Industrial CASE (ICASE) award.

\bibliography{refs}

\newpage

\clearpage

\appendix

\section{Experiments set-up}

In this section, we provide further details on datasets, model architectures and training methods used in this paper. All experiments were run on a workstation with a single 24GB NVIDIA Quadro RTX 6000 GPU, Intel(R) Core(TM) i9-10900K CPU @ 3.70GHz and 64GB of system memory. The machine runs Ubuntu 22.04.4 LTS. We estimate around 500 hours are required to train models and run all experiments, which includes training each model 5 times on random seeds. All random seeds were selected using the shuf command to select a number between 0 and 1000.

\subsection{Datasets} \label{datasets_appendix}

\subsubsection{Playing cards}

Playing cards is a synthetic image dataset we introduced to analyse CBMs. In each sample playing cards are placed onto a random background. The playing cards are placed in a line with their size, rotation, location and the background image varying from sample to sample. The dataset has annotations for which playing cards are present in each sample, a task label representing hand ranks in the game Three Card Poker, and the coordinates of each playing card in the sample. We use playing cards as concepts.

\begin{figure}[ht!]
  \centering
  \includegraphics[width=0.30\textwidth]{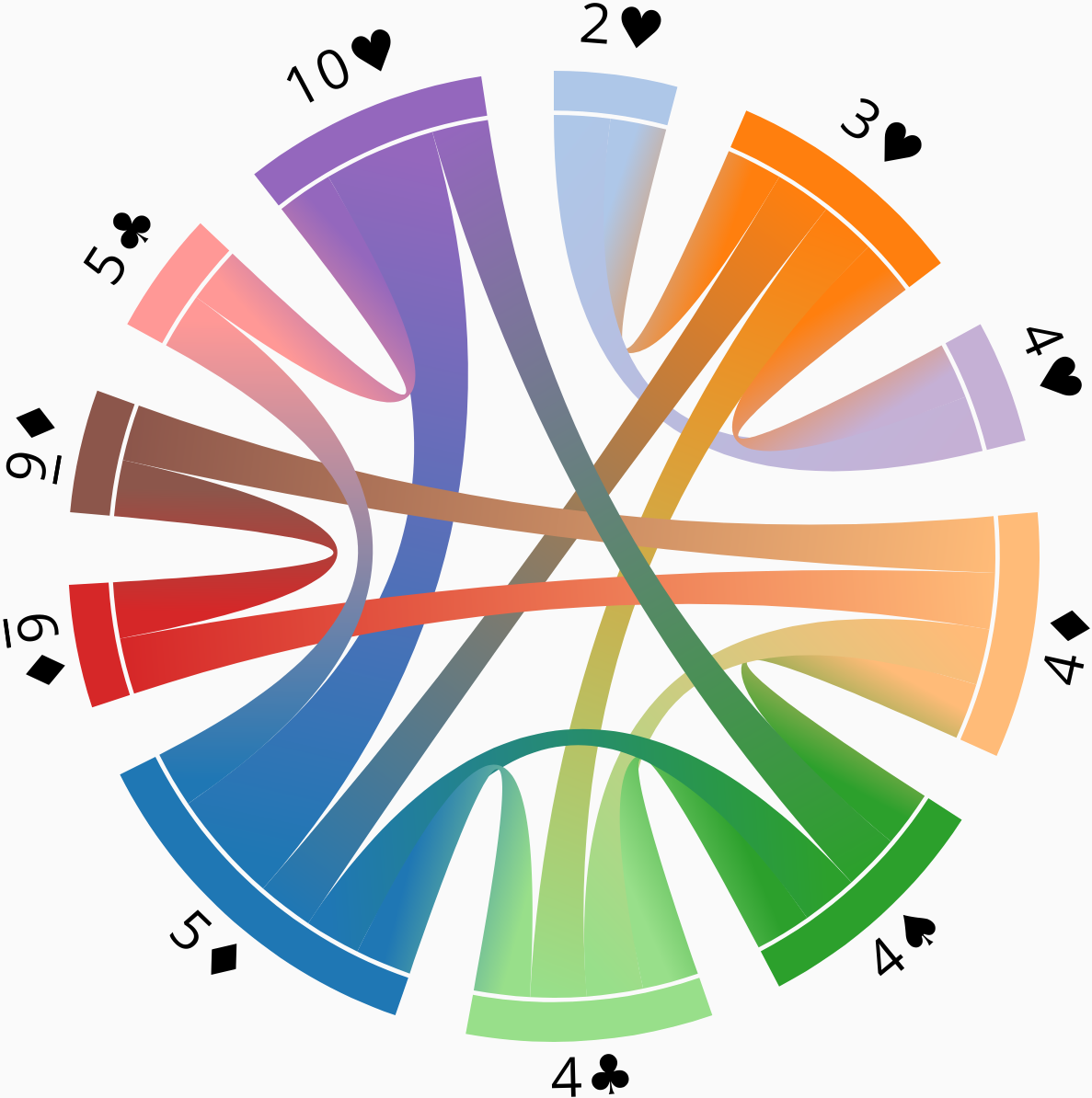}
  \caption{Class-level poker cards has concepts arranged such that some only appear in the same image with two other concepts (e.g. Two of Hearts, Three of Hearts and Four of Hearts) while some appear in the same image as many other concepts (e.g. Five of Diamonds). This diagram links together the concepts that appear together.}
  \label{fig:class-level-poker-cards}
  \vspace{0.5cm}
\end{figure}

The dataset consists of multiple image variations, of which we use three in this paper: \textit{Random cards}, \textit{poker cards} and \textit{class-level poker cards}. Each image variation has 10,000 images where 70\% are training images and 30\% are testing/validation images. Random cards have concepts selected at random in each image to ensure there is no correlation between concepts that appear together, however, this means the downstream class occurrences are unbalanced as most samples belong to the class ``high card'' while 20 have the class "straight flush". Poker cards tackle the unbalanced downstream classes but this means there are some inherent correlations between concepts as there are only a few unique triplets of cards for some downstream classes. The most extreme of these is ``straight flush'', where there are only 48 unique triplets of cards. We duplicate these triplets until the final number of downstream class instances is reached. Class-level poker cards have the same task classes as random cards and poker cards, with concepts selected such that some concepts are only used for one task class while others are used for many. This means class-level poker cards cannot use the full 52 concepts that can appear with the other image variations and instead use a subset of 11 concepts. We show which concepts appear together in Figure~\ref{fig:class-level-poker-cards} using a chord diagram. Each concept is linked to other concepts that appear together with the thickness of each link representing the number of samples. Code to generate the dataset is publicly available.\footnote{Playing cards dataset generator: Available in camera-ready submission}

% \url{https://github.com/JackFurby/playing-card-concept-generator}

In this paper, we transform training samples with a random flip, both horizontal and vertical, apply a colour jitter to the brightness, contrast, saturation and hue, and randomly convert to grey scale. Samples are scaled to 299 by 299 pixels.

\subsubsection{CheXpert}

CheXpert \citep{chexpert} is a real-world image dataset with visually represented observations. Each sample has 14 observations such as fracture and edema, of which 12 are pathologies. The other two observations are ``support devices'', e.g. a pacemaker, and ``no\_findings''. We use 13 of these observations as concepts while the 14th, ``no\_findings'', as the task. CheXpert has instance-level concepts and contains 224,316 chest X-ray images. We use the official dataset splits from \citep{chexpert} which has 223,414 training images, 234 validation images and 668 test images. We also created a modified version of the dataset to use class-level concepts with the most common concept vector for three, four and five concepts present. These have 44,974 samples, 21,760 samples, and 636 samples respectively. We refer to the original version of CheXpert as \textit{instance-level CheXpert} and the modified version as \textit{class-level CheXpert}.

During training, samples are randomly rotated by up to 15 degrees, translated by up to 5\% of the overall image width and scaled by up to 5\%. All samples are resized to 512 by 512 pixels.

\subsection{Models and training} \label{model_training_appendix}

\textbf{Playing cards}: All playing card models use a VGG-11 architecture with batch normalisation \citep{Simonyan15} for the concept encoder. The task predictor used two linear layers with a ReLU activation function. Between the two model parts, we used a sigmoid function. During training, we evaluate these models w.r.t. the combined concept and downstream loss. The standard NN uses the same architecture as our CBMs and is trained using the joint training method but with concept loss disabled. As random cards has a heavily unbalanced task, we train random card models on random cards for the concept encoder and poker cards for the task predictor. We detail average model accuracies in Table~\ref{playing-cards-model-accuracy-table-appendix}. For experiments that used a single model (e.g. saliency map generation), we selected the model with the highest concept accuracy. These are displayed in Table~\ref{playing-card-single-model-accuracy-table-appendix}.

\begin{table*}[p]
    \centering
    \caption{Playing card models averaged accuracy. All values are rounded to 3 decimal places.}
    \begin{tabular}{llp{2cm}p{2cm}p{2cm}p{2cm}p{2cm}}
        \toprule
        Training method & Dataset & Average \newline concept \newline accuracy (\%) & Concept \newline standard \newline deviation & Average task accuracy (\%) & Task standard deviation \\
        \midrule
        Independent & Random cards & 99.943 & 0.008 & 99.174 & 0.09     \\
        Sequential & Random cards & 99.921 & 0.044 & 97.457 & 0.76     \\
        Independent & Poker cards & 99.957 & 0.005 & 99.421 & 0.034     \\
        Sequential & Poker cards & 99.917 & 0.051 & 98.802 & 0.277     \\
        Joint & Poker cards & 99.867 & 0.046 & 96.005 & 0.213     \\
        Independent & Class-level poker cards & 99.98 & 0.014 & 99.96 & 0.039     \\
        Sequential & Class-level poker cards & 99.98 & 0.014 & 99.953 & 0.045     \\
        Joint & Class-level poker cards & 100 & 0 & 100 & 0     \\
        Standard NN & Poker cards & 50.25 & 1.31 & 67.137 & 0.584     \\
        \bottomrule
    \end{tabular}
  \label{playing-cards-model-accuracy-table-appendix}
\end{table*}

\begin{table*}[p]
    \centering
    \caption{Concept accuracy for models used to generate saliency maps}
    \begin{tabular}{lll}
        \toprule
        Model & Dataset & Concept accuracy (\%) \\
        \midrule
        Independent \& Sequential & Random cards & 99.961     \\
        Independent \& Sequential & Poker cards & 99.961     \\
        Independent \& Sequential & Class-level poker cards & 99.997     \\
        Joint & Poker cards & 99.936   \\
        Standard NN & Poker cards & 52.717     \\
        \bottomrule
    \end{tabular}
    \label{playing-card-single-model-accuracy-table-appendix}
\end{table*}

We used Weights and Biases Sweeps \citep{wandb} to find optimal hyperparameters for training each of our models on the playing cards dataset. This was configured with a Bayesian search method to optimise the parameters. These were; starting learning rate (between 0.1 and 0.001), optimizer (between Adam \citep{adam} and stochastic gradient descent (SGD)), learning rate patience (between 3, 5, 10 and 15 epochs of no improvement in loss) and $\lambda$ value (A value used with joint training to balance concept and task loss). Sweeps were run for each of the dataset variants and CBM training. Each sweep ran until we stopped seeing improvements in the model accuracy, about 30 iterations per sweep. The final hyperparameters we used are in Table~\ref{card-hyperparams-table}.

\begin{table*}[p]
  \centering
  \caption{Playing cards training hyperparameters}
  \begin{tabular}{p{4cm}p{2cm}p{1.5cm}p{1.7cm}p{2.2cm}p{1cm}p{1cm}}
    \toprule
    Training method \& Dataset & Learning rate & Optimizer & Batch size & Learning rate patience & $\lambda$ & Epochs \\
    \midrule
    Independent \& sequential concept encoder - Random cards & 0.03 & SGD & 32 & 15 & N/A & 200 \\
    Independent \& sequential concept encoder - Poker cards & 0.02 & SGD & 32 & 15 & N/A & 200\\
    Independent \& sequential concept encoder - Class-level poker cards & 0.0825 & SGD & 32 & 3 & N/A & 100\\
    Independent task predictor - Random cards & 0.01 & Adam & 32 & 5 & N/A & 200 \\
    Independent task predictor - Poker cards & 0.01 & Adam & 32 & 5 & N/A & 200\\
    Independent task predictor - Class-level poker cards & 0.064 & Adam & 32 & 5 & N/A & 100\\
    Sequential task predictor - Random cards & 0.059 & Adam & 32 & 5 & N/A & 200 \\
    Sequential task predictor - Poker cards & 0.046 & Adam & 32 & 15 & N/A & 200 \\
    Sequential task predictor - Class-level poker cards & 0.0846 & Adam & 32 & 10 & N/A & 100 \\
    Joint - Poker cards & 0.025 & SGD & 32 & 15 & 0.98 & 300\\
    Joint - Class-level poker cards & 0.0398 & SGD & 32 & 15 & 0.867 & 100\\
    Standard NN - Poker cards & 0.088 & SGD & 32 & 15 & 0 & 300\\
    \bottomrule
  \end{tabular}
  \label{card-hyperparams-table}
\end{table*}

\textbf{CheXpert \citep{chexpert}}: Our CheXpert models use a Densenet121 architecture \citep{densenet} for the concept encoder and pre-trained weights, trained on Imagenet. The task predictor used two linear layers with a ReLU activation function. Between the two model parts, we used a sigmoid function. We evaluated these models w.r.t. Area Under the receiver operating characteristic Curve (AUC) for concepts. Averaged model performance is shown in Table~\ref{chexpert-model-accuracy-table}. The best model accuracies are shown in Table~\ref{chexpert-single-model-accuracy-table} and AUC values in Table~\ref{chexpert-single-model-auc-table}. These models were selected as they had the highest AUC values. Model training hyperparameters are listed in Table~\ref{chexpert-hyperparmeters}.

\begin{table*}[p]
    \centering
    \caption{CheXpert models averaged accuracy. All values are rounded to 3 decimal places.}
    \begin{tabular}{p{2cm}p{4cm}p{2cm}p{2cm}p{2cm}p{2cm}}
        \toprule
        Training method & Dataset \newline version & Concept \newline accuracy (\%) & Concept \newline standard \newline deviation & Task accuracy (\%) & Task standard deviation \\
        \midrule
        Sequential & Instance-level & 75.765 & 1.123 & 84.701 & 0.631     \\
        Joint & Instance-level & 74.696 & 1.224 & 85.15 & 0.617     \\
        Sequential & Class-level with 3 present concepts & 59.179 & 8.798 & 95 & 0.689     \\
        Sequential & Class-level with 4 present concepts & 63.901 & 9.752 & 95.714 & 1.429      \\
        Sequential & Class-level with 5 present concepts & 65.282 & 9.053 & 96 & 1.333     \\
        Joint & Class-level with 3 present concepts & 56.473 & 2.894 & 96.757 & 1.162  \\
        Joint & Class-level with 4 present concepts & 60.385 & 4.157 & 97.143 & 2.673  \\
        Joint & Class-level with 5 present concepts & 61.948 & 4.174 & 95.333 & 1.633   \\
        \bottomrule
    \end{tabular}
  \label{chexpert-model-accuracy-table}
\end{table*}

\begin{table*}[p]
    \centering
    \caption{Concept accuracy for models used to generate saliency maps}
    \begin{tabular}{lll}
        \toprule
        Model & Dataset version & Concept accuracy (\%) \\
        \midrule
        Sequential & Instance-level CheXpert & 75.903     \\
        Joint & Instance-level CheXpert & 75.808   \\
        Sequential & Class-level CheXpert with 3 present concepts & 59.683     \\
        Sequential & Class-level CheXpert with 4 present concepts & 68.132    \\
        Sequential & Class-level CheXpert with 5 present concepts & 74.359    \\
        \bottomrule
    \end{tabular}
    \label{chexpert-single-model-accuracy-table}
\end{table*}

\begin{table*}[p]
    \centering
    \caption{Concept AUC for models used to generate saliency maps}
    \begin{tabular}{llllll}
        \toprule
        & \rotatebox[origin=m]{90}{Sequential instance-level CheXpert} & \rotatebox[origin=m]{90}{Joint instance-level CheXpert} & \rotatebox[origin=m]{90}{Joint class-Level CheXpert with 3 present concepts} & \rotatebox[origin=m]{90}{Joint class-Level CheXpert with 4 present concepts} & \rotatebox[origin=m]{90}{Joint class-Level CheXpert with 5 present concepts} \\
        \midrule
        Enlarged cardiomediastinum & 0.6087 & 0.7214 & N/A & N/A & N/A \\
        Cardiomegaly  & 0.8943 & 0.9200 & N/A & N/A & N/A \\
        Lung opacity   & 0.9069 & 0.9290 & 1.0 & 1.0 & 1.0 \\
        Lung lesion  & 0.5014 & 0.5453 & N/A & N/A & N/A \\
        Edema  & 0.9251 & 0.8458 & N/A & 1.0 & N/A \\
        Consolidation  & 0.7431 & 0.6445 & N/A & N/A & 1.0 \\
        Pneumonia  & 0.4466 & 0.2809 & N/A & N/A & N/A \\
        Atelectasis  & 0.8337 & 0.8432 & N/A & N/A & 1.0 \\
        Pneumothorax  & 0.3318 & 0.3645 & N/A & N/A & N/A \\
        Pleural effusion  & 0.9473 & 0.9554 & 1.0 & 1.0 & 1.0 \\
        Plaural other  & 0.3187 & 0.3078 & N/A & N/A & N/A \\
        Fracture & 0.2427 & 0.1874 & & & \\
        Support devices  & 0.9528 & 0.9467 & 1.0 & 1.0 & 1.0 \\
        \bottomrule
    \end{tabular}
    \label{chexpert-single-model-auc-table}
\end{table*}

\begin{table*}[p]
  \centering
  \caption{CheXpert training hyperparameters}
  \begin{tabular}{p{4cm}p{2cm}p{1.5cm}p{1.7cm}p{2.2cm}p{1cm}p{1cm}}
    \toprule
    Training method & Learning rate & Optimizer & Batch size & Learning rate patience & $\lambda$ & Epochs \\
    \midrule
    Independent \& sequential concept encoder & 0.0001 & Adam & 14 & 0.1 & N/A & 3 \\
    Sequential task predictor & 0.0001 & Adam & 14 & 0.1 & N/A & 3 \\
    Joint & 0.0001 & Adam & 14 & 0.1 & 0.99 & 3\\
    \bottomrule
  \end{tabular}
  \label{chexpert-hyperparmeters}
\end{table*}

\section{Experiments and results} \label{results_appendix}

\subsection{Playing cards} \label{playing_cards_saliency_appendix}

We use three explanation techniques LPR, IG with a smoothgrad noise tunnel, and IG with a smoothgrad squared noise tunnel. LRP defines rules that control how relevancy is propagated from neuron to neuron in the NN \citep{10.1371/journal.pone.0130140}. From the first layer of the concept encoder, we use the rules LRP-$\alpha \beta$, where $\alpha=1$ and $\beta=0$, for the first four convolutional layers, LRP-$\epsilon$ for the next four convolutional layers and LRP-0 for the top three linear layers. These rules were selected to be similar to those used by \citep{Montavon2019}. IG is combined with a smoothgrad noise tunnel configured to use 25 randomly generated samples and a standard deviation of 0.

\begin{figure*}[p]
     \centering
     \begin{subfigure}[b]{0.45\textwidth}
         \centering
\begin{tabular}{b{0.1\textwidth} m{0.2\textwidth} m{0.2\textwidth} m{0.2\textwidth} }
        &
        input &
        &
     \\
        &
        \frame{\includegraphics[width=0.24\textwidth]{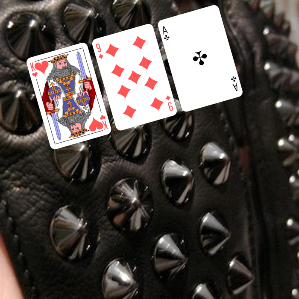}} & 
        &
     \\
        &
        King of Hearts &  
        Nine of Diamonds &  
        Ace of Clubs
     \\
        \rotatebox[origin=m]{90}{\parbox{0.24\textwidth}{\centering LRP}} &
        \frame{\includegraphics[width=0.24\textwidth]{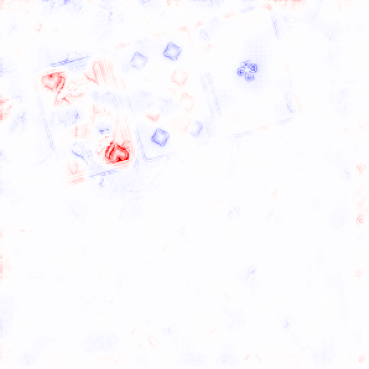}} &  
        \frame{\includegraphics[width=0.24\textwidth]{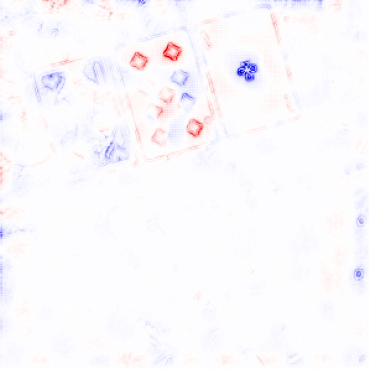}} &  
        \frame{\includegraphics[width=0.24\textwidth]{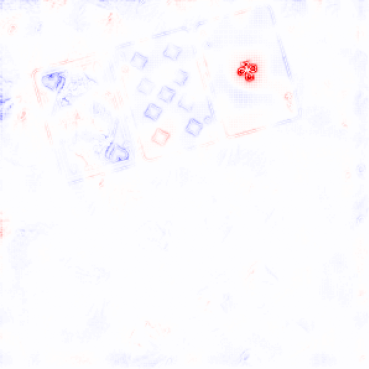}}
     \\
        \rotatebox[origin=m]{90}{\parbox{0.24\textwidth}{\centering IG with Smoothgrad}} &
        \frame{\includegraphics[width=0.24\textwidth]{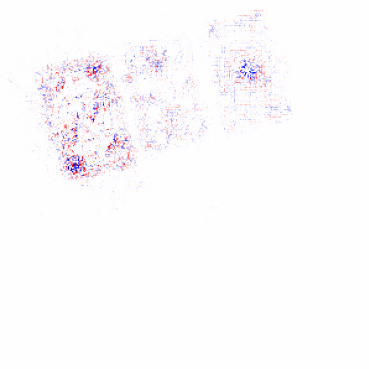}} &  
        \frame{\includegraphics[width=0.24\textwidth]{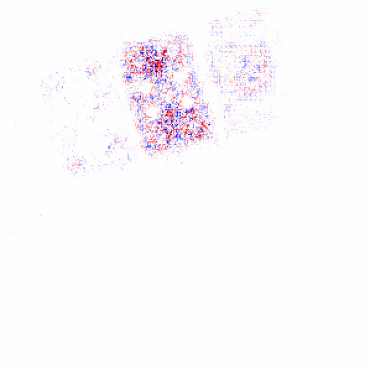}} &  
        \frame{\includegraphics[width=0.24\textwidth]{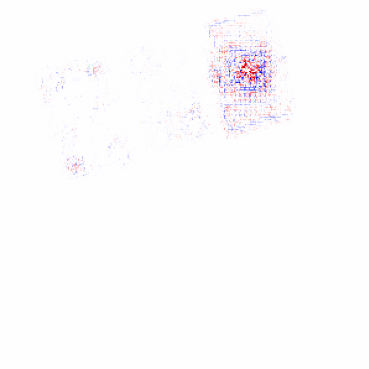}}  
     \\
        \rotatebox[origin=m]{90}{\parbox{0.24\textwidth}{\centering IG with Smoothgrad squared}} &  
        \frame{\includegraphics[width=0.24\textwidth]{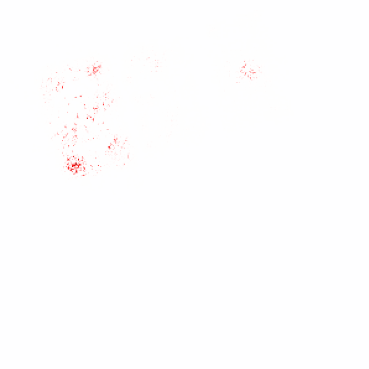}} &  
        \frame{\includegraphics[width=0.24\textwidth]{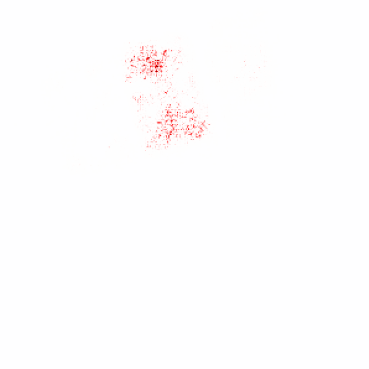}} &  
        \frame{\includegraphics[width=0.24\textwidth]{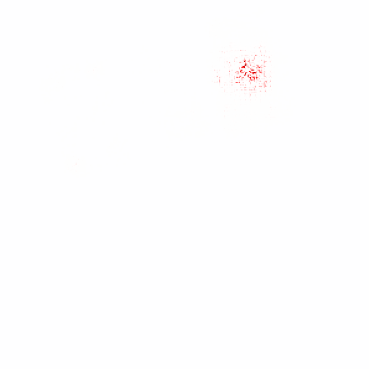}} 
     \\
    \end{tabular}
         \caption{High card task classification}
         \label{fig:indi_poker_salience_appendix_1}
     \end{subfigure}
     \hfill
     \begin{subfigure}[b]{0.45\textwidth}
         \centering
         \begin{tabular}{b{0.1\textwidth} m{0.2\textwidth} m{0.2\textwidth} m{0.2\textwidth} }
        &
        input &
        &
     \\
        &
        \frame{\includegraphics[width=0.24\textwidth]{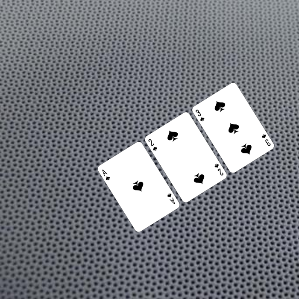}} & 
        &
     \\
        &
        Ace of Spades &  
        Two of Spades &  
        Three of Spades
     \\
        \rotatebox[origin=m]{90}{\parbox{0.24\textwidth}{\centering LRP}} &
        \frame{\includegraphics[width=0.24\textwidth]{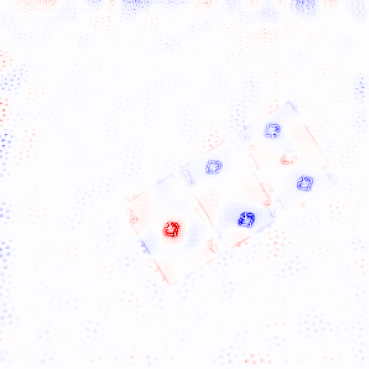}} &  
        \frame{\includegraphics[width=0.24\textwidth]{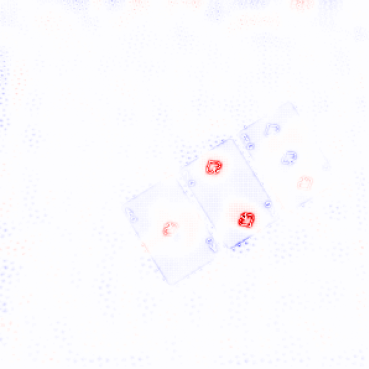}} &  
        \frame{\includegraphics[width=0.24\textwidth]{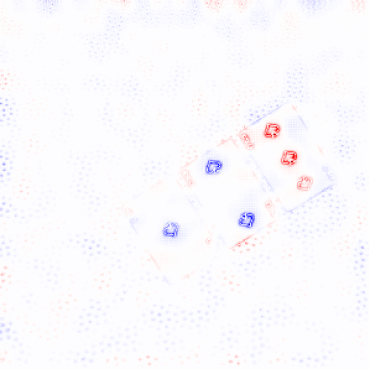}}
     \\
        \rotatebox[origin=m]{90}{\parbox{0.24\textwidth}{\centering IG with Smoothgrad}} &
        \frame{\includegraphics[width=0.24\textwidth]{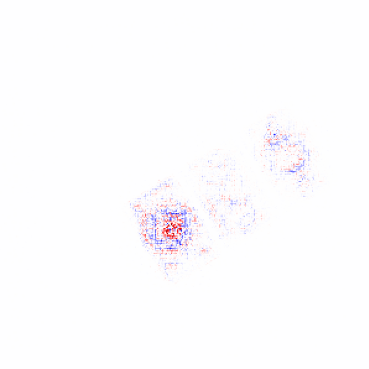}} &  
        \frame{\includegraphics[width=0.24\textwidth]{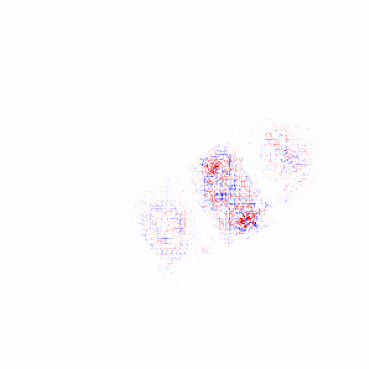}} &  
        \frame{\includegraphics[width=0.24\textwidth]{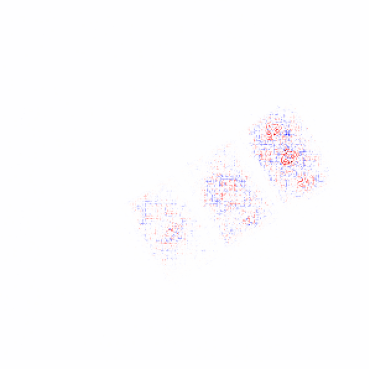}}  
     \\
        \rotatebox[origin=m]{90}{\parbox{0.24\textwidth}{\centering IG with Smoothgrad squared}} &  
        \frame{\includegraphics[width=0.24\textwidth]{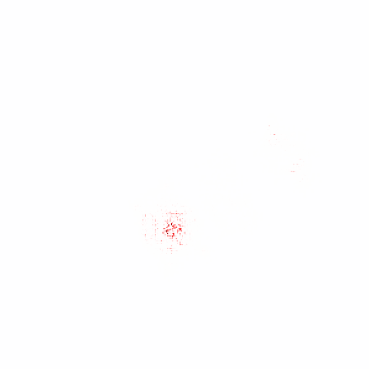}} &  
        \frame{\includegraphics[width=0.24\textwidth]{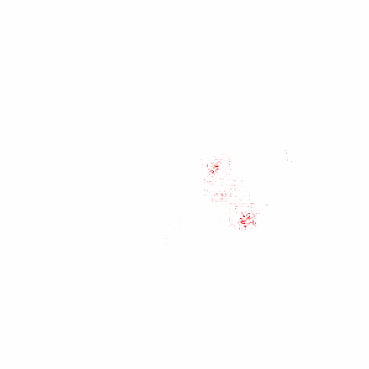}} &  
        \frame{\includegraphics[width=0.24\textwidth]{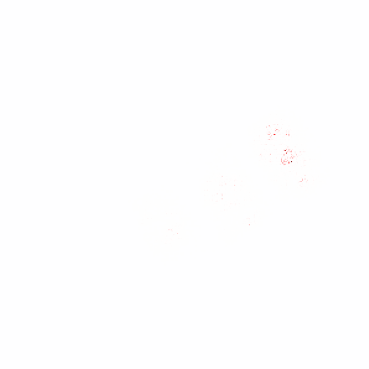}} 
     \\
    \end{tabular}
         \caption{Straight flush task classification}
         \label{fig:indi_poker_salience_appendix_2}
     \end{subfigure}

    \vspace{0.5cm}
    \caption{Independent / sequential model trained on poker cards}
\label{fig:indi_poker_salience_appendix}
\end{figure*}

\begin{figure*}[p]
     \centering
     \begin{subfigure}[b]{0.45\textwidth}
         \centering
         \begin{tabular}{b{0.10\textwidth} m{0.2\textwidth} m{0.2\textwidth} m{0.2\textwidth} }
        &
        input &
        &
     \\
        &
        \frame{\includegraphics[width=0.24\textwidth]{media/XtoC/high_card_2508/input.png}} & 
        &
     \\
        &
        King of Hearts &  
        Nine of Diamonds &  
        Ace of Clubs
     \\
        \rotatebox[origin=m]{90}{\parbox{0.24\textwidth}{\centering LRP}} &
        \frame{\includegraphics[width=0.24\textwidth]{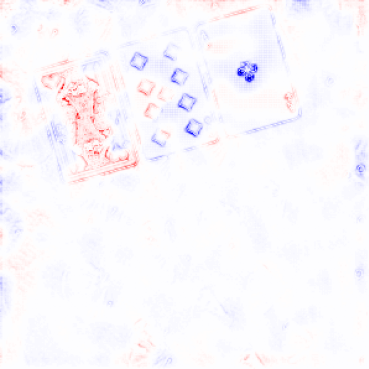}} &  
        \frame{\includegraphics[width=0.24\textwidth]{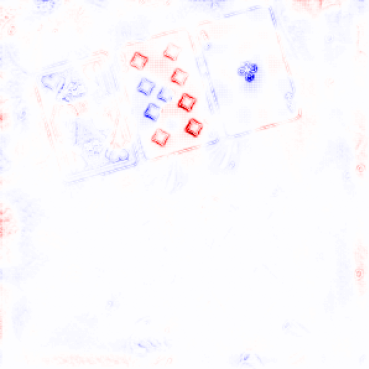}} &  
        \frame{\includegraphics[width=0.24\textwidth]{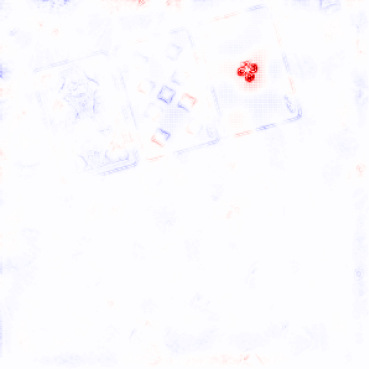}}
     \\
        \rotatebox[origin=m]{90}{\parbox{0.24\textwidth}{\centering IG with Smoothgrad}} &
        \frame{\includegraphics[width=0.24\textwidth]{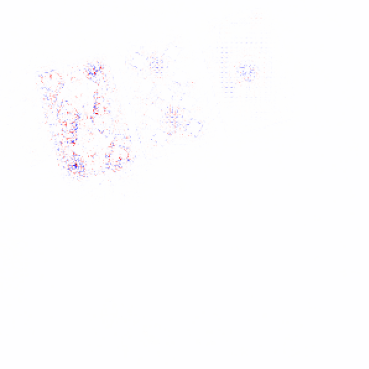}} &  
        \frame{\includegraphics[width=0.24\textwidth]{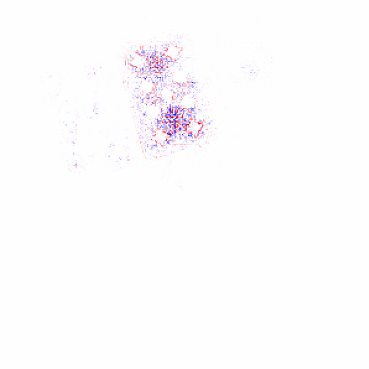}} &  
        \frame{\includegraphics[width=0.24\textwidth]{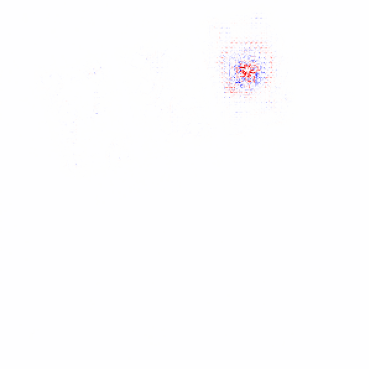}}  
     \\
        \rotatebox[origin=m]{90}{\parbox{0.24\textwidth}{\centering IG with Smoothgrad squared}} &  
        \frame{\includegraphics[width=0.24\textwidth]{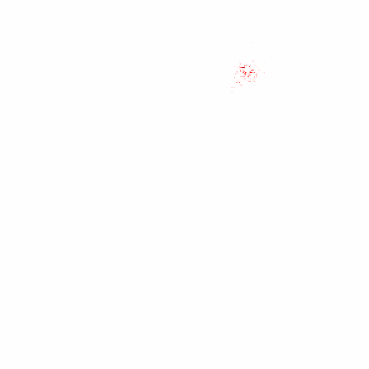}} &  
        \frame{\includegraphics[width=0.24\textwidth]{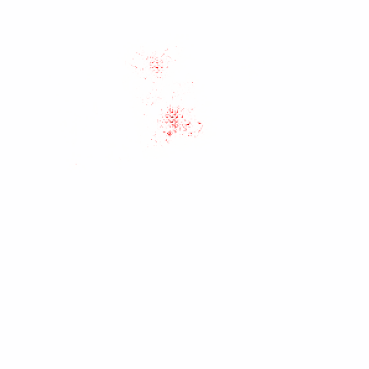}} &  
        \frame{\includegraphics[width=0.24\textwidth]{media/XtoC/high_card_2508/AC_three_ig_sg_sq.png}} 
     \\
    \end{tabular}
         \caption{High card task classification high card}
         \label{fig:indi_three_salience_appendix_1}
     \end{subfigure}
     \hfill
     \begin{subfigure}[b]{0.45\textwidth}
         \centering
         \begin{tabular}{b{0.10\textwidth} m{0.2\textwidth} m{0.2\textwidth} m{0.2\textwidth} }
        &
        input &
        &
     \\
        &
        \frame{\includegraphics[width=0.24\textwidth]{media/XtoC/straight_flush_3/input.png}} & 
        &
     \\
        &
        Ace of Spades &  
        Two of Spades &  
        Three of Spades
     \\
        \rotatebox[origin=m]{90}{\parbox{0.24\textwidth}{\centering LRP}} &
        \frame{\includegraphics[width=0.24\textwidth]{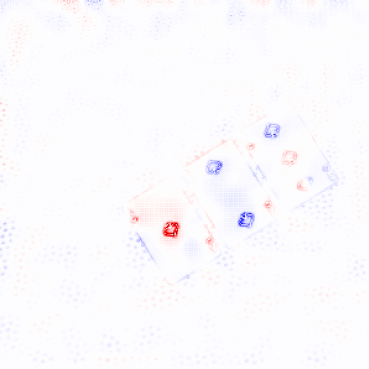}} &  
        \frame{\includegraphics[width=0.24\textwidth]{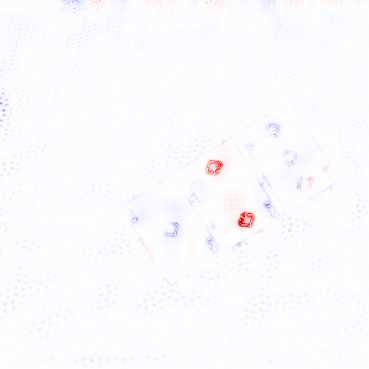}} &  
        \frame{\includegraphics[width=0.24\textwidth]{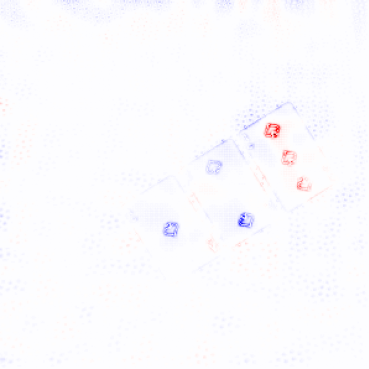}}
     \\
        \rotatebox[origin=m]{90}{\parbox{0.24\textwidth}{\centering IG with Smoothgrad}} &
        \frame{\includegraphics[width=0.24\textwidth]{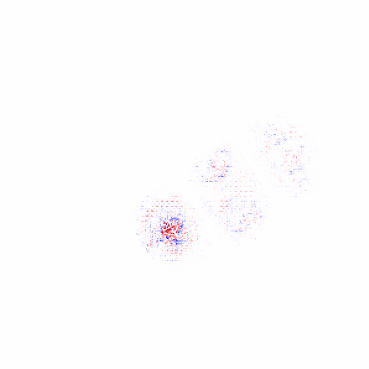}} &  
        \frame{\includegraphics[width=0.24\textwidth]{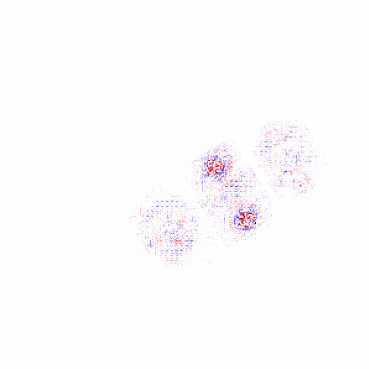}} &  
        \frame{\includegraphics[width=0.24\textwidth]{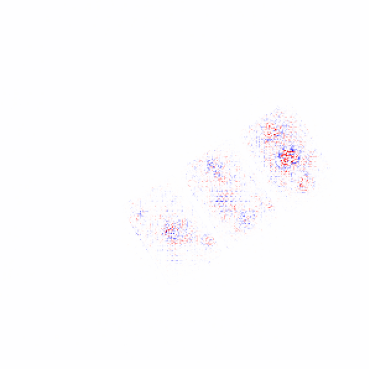}}  
     \\
        \rotatebox[origin=m]{90}{\parbox{0.24\textwidth}{\centering IG with Smoothgrad squared}} &  
        \frame{\includegraphics[width=0.24\textwidth]{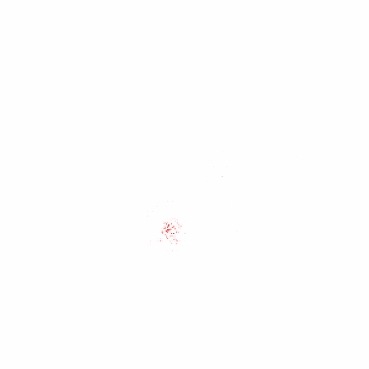}} &  
        \frame{\includegraphics[width=0.24\textwidth]{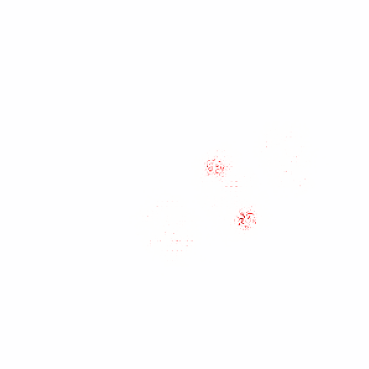}} &  
        \frame{\includegraphics[width=0.24\textwidth]{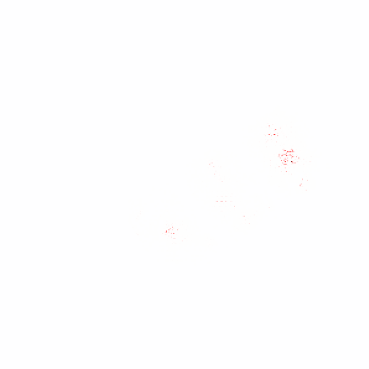}} 
     \\
    \end{tabular}
         \caption{Straight flush task classification}
         \label{fig:indi_three_salience_appendix_2}
     \end{subfigure}

    \vspace{0.5cm}
    \caption{Independent / sequential model trained on random cards}
\label{fig:indi_three_salience_appendix}
\end{figure*}

\begin{figure*}[p]
     \centering
     \begin{subfigure}[b]{0.45\textwidth}
         \centering
         \begin{tabular}{b{0.1\textwidth} m{0.2\textwidth} m{0.2\textwidth} m{0.2\textwidth} }
        &
        input &
        &
     \\
        &
        \frame{\includegraphics[width=0.24\textwidth]{media/XtoC/high_card_2508/input.png}} & 
        &
     \\
        &
        King of Hearts &  
        Nine of Diamonds &  
        Ace of Clubs
     \\
        \rotatebox[origin=m]{90}{\parbox{0.24\textwidth}{\centering LRP}} &
        \frame{\includegraphics[width=0.24\textwidth]{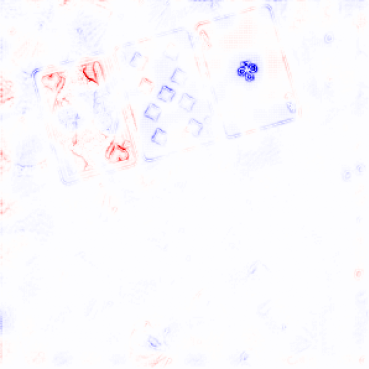}} &  
        \frame{\includegraphics[width=0.24\textwidth]{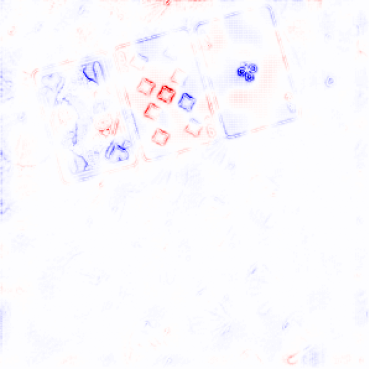}} &  
        \frame{\includegraphics[width=0.24\textwidth]{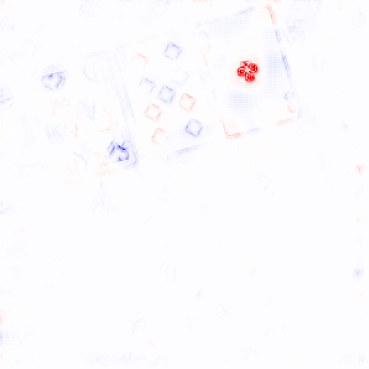}}
     \\
        \rotatebox[origin=m]{90}{\parbox{0.24\textwidth}{\centering IG with Smoothgrad}} &
        \frame{\includegraphics[width=0.24\textwidth]{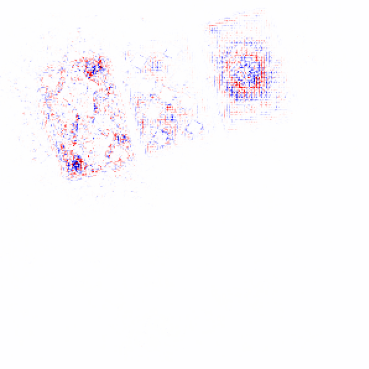}} &  
        \frame{\includegraphics[width=0.24\textwidth]{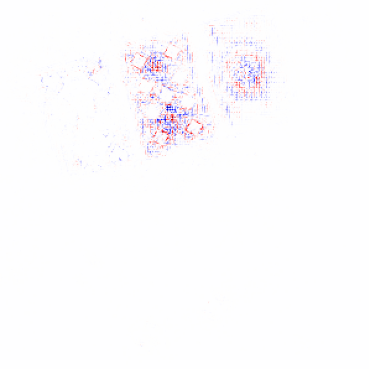}} &  
        \frame{\includegraphics[width=0.24\textwidth]{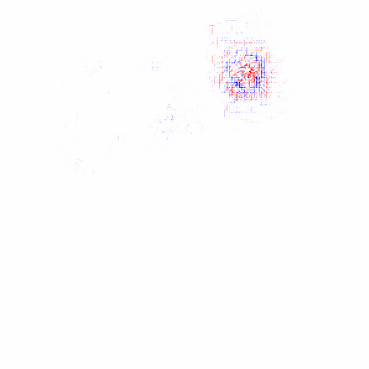}}  
     \\
        \rotatebox[origin=m]{90}{\parbox{0.24\textwidth}{\centering IG with Smoothgrad squared}} &  
        \frame{\includegraphics[width=0.24\textwidth]{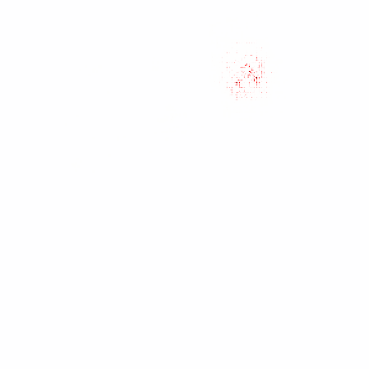}} &  
        \frame{\includegraphics[width=0.24\textwidth]{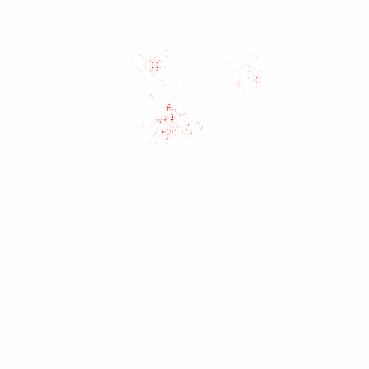}} &  
        \frame{\includegraphics[width=0.24\textwidth]{media/XtoC/high_card_2508/AC_joint_poker_ig_sg_sq.png}}  
     \\
    \end{tabular}
         \caption{High card task classification}
         \label{fig:joint_poker_salience_appendix_1}
     \end{subfigure}
     \hfill
     \begin{subfigure}[b]{0.45\textwidth}
         \centering
         \begin{tabular}{b{0.1\textwidth} m{0.2\textwidth} m{0.2\textwidth} m{0.2\textwidth} }
        &
        input &
        &
     \\
        &
        \frame{\includegraphics[width=0.24\textwidth]{media/XtoC/straight_flush_3/input.png}} & 
        &
     \\
        &
        Ace of Spades &  
        Two of Spades &  
        Three of Spades
     \\
        \rotatebox[origin=m]{90}{\parbox{0.24\textwidth}{\centering LRP}} &
        \frame{\includegraphics[width=0.24\textwidth]{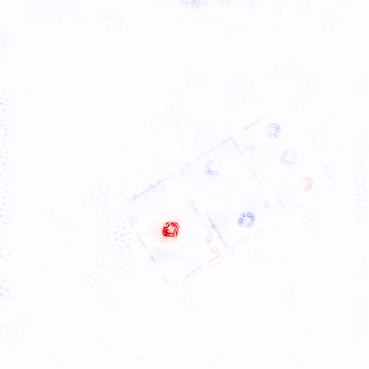}} &  
        \frame{\includegraphics[width=0.24\textwidth]{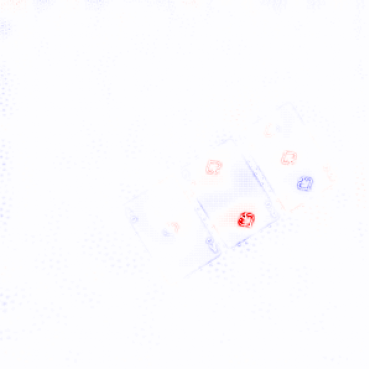
        }} &  
        \frame{\includegraphics[width=0.24\textwidth]{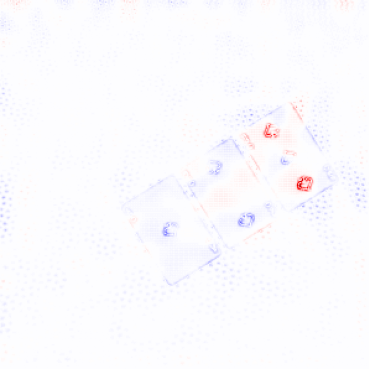}}
     \\
        \rotatebox[origin=m]{90}{\parbox{0.24\textwidth}{\centering IG with Smoothgrad}} &
        \frame{\includegraphics[width=0.24\textwidth]{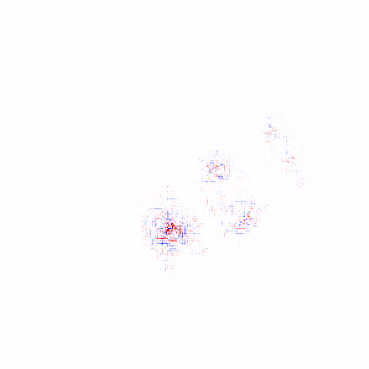}} &  
        \frame{\includegraphics[width=0.24\textwidth]{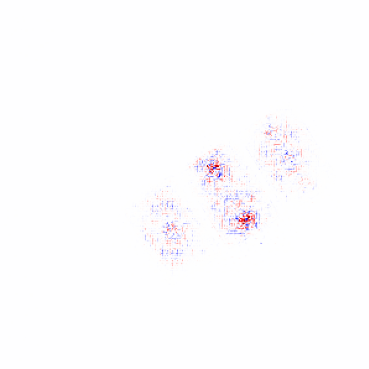}} &  
        \frame{\includegraphics[width=0.24\textwidth]{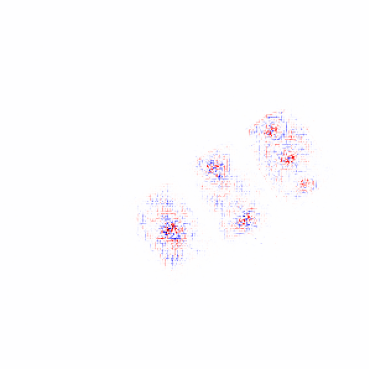}}  
     \\
        \rotatebox[origin=m]{90}{\parbox{0.24\textwidth}{\centering IG with Smoothgrad squared}} &  
        \frame{\includegraphics[width=0.24\textwidth]{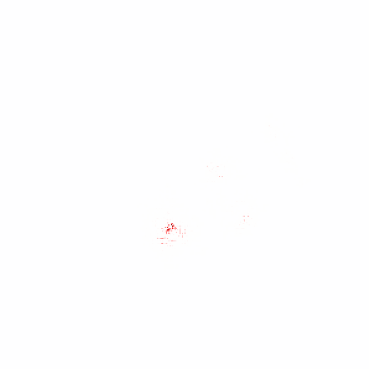}} &  
        \frame{\includegraphics[width=0.24\textwidth]{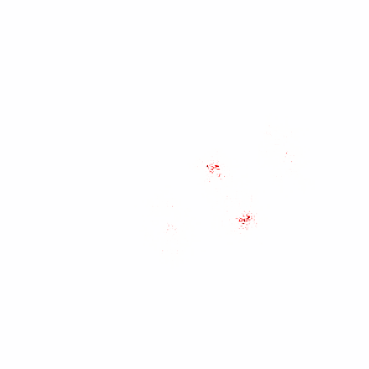}} &  
        \frame{\includegraphics[width=0.24\textwidth]{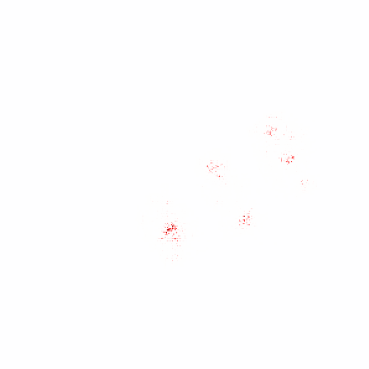}} 
     \\
    \end{tabular}
         \caption{Straight flush task classification}
         \label{fig:joint_poker_salience_appendix_2}
     \end{subfigure}
     
    \vspace{0.5cm}
    \caption{Joint model trained on poker cards}
\label{fig:joint_poker_salience_appendix}
\end{figure*}

\begin{figure*}[p]
     \centering
     \begin{subfigure}[b]{0.45\textwidth}
         \centering
         \begin{tabular}{b{0.1\textwidth} m{0.2\textwidth} m{0.2\textwidth} m{0.2\textwidth} }
        &
        input &
        &
     \\
        &
        \frame{\includegraphics[width=0.24\textwidth]{media/XtoC/high_card_2508/input.png}} & 
        &
     \\
        &
        King of Hearts &  
        Nine of Diamonds &  
        Ace of Clubs
     \\
        \rotatebox[origin=m]{90}{\parbox{0.24\textwidth}{\centering LRP}} &
        \frame{\includegraphics[width=0.24\textwidth]{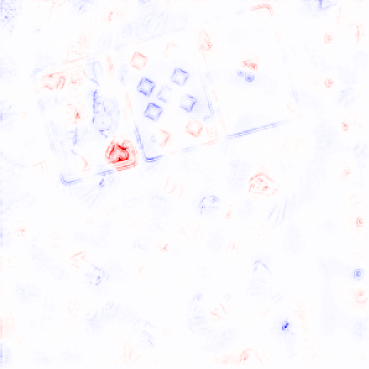}} &  
        \frame{\includegraphics[width=0.24\textwidth]{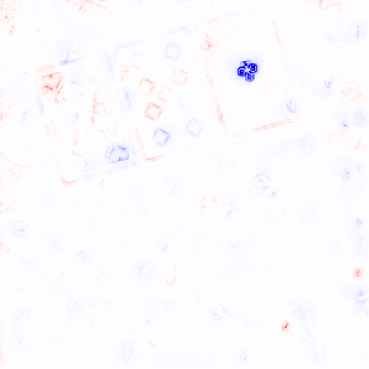}} &  
        \frame{\includegraphics[width=0.24\textwidth]{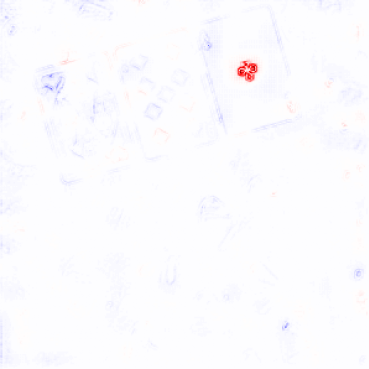}}
     \\
        \rotatebox[origin=m]{90}{\parbox{0.24\textwidth}{\centering IG with Smoothgrad}} &
        \frame{\includegraphics[width=0.24\textwidth]{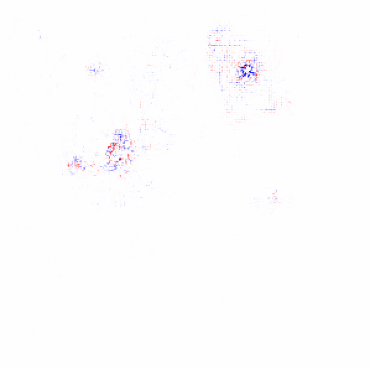}} &  
        \frame{\includegraphics[width=0.24\textwidth]{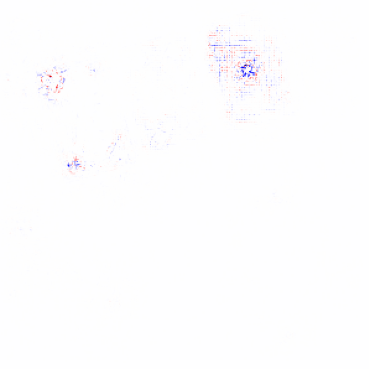}} &  
        \frame{\includegraphics[width=0.24\textwidth]{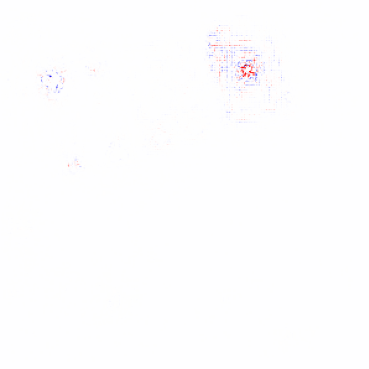}}  
     \\
        \rotatebox[origin=m]{90}{\parbox{0.24\textwidth}{\centering IG with Smoothgrad squared}} &  
        \frame{\includegraphics[width=0.24\textwidth]{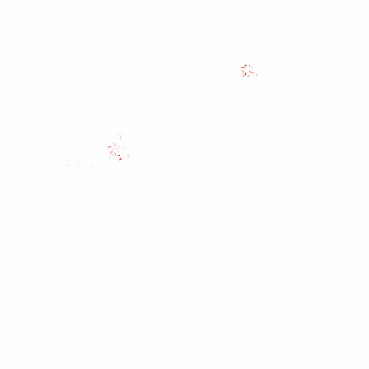}} &  
        \frame{\includegraphics[width=0.24\textwidth]{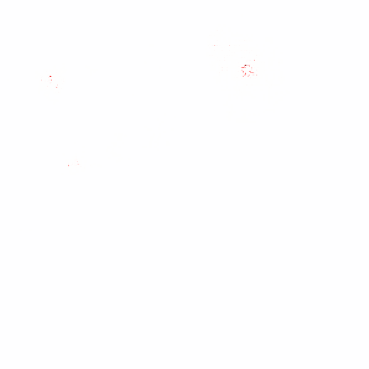}} &  
        \frame{\includegraphics[width=0.24\textwidth]{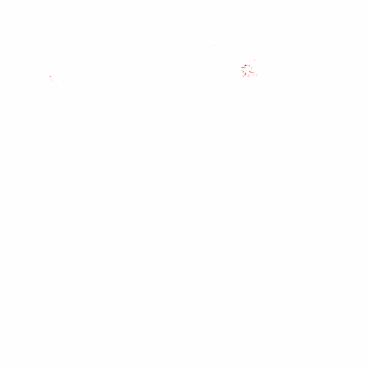}} 
     \\
    \end{tabular}
         \caption{High card task classification}
         \label{fig:standard_poker_salience_appendix_1}
     \end{subfigure}
     \hfill
     \begin{subfigure}[b]{0.45\textwidth}
         \centering
         \begin{tabular}{b{0.1\textwidth} m{0.2\textwidth} m{0.2\textwidth} m{0.2\textwidth} }
        &
        input &
        &
     \\
        &
        \frame{\includegraphics[width=0.24\textwidth]{media/XtoC/straight_flush_3/input.png}} & 
        &
     \\
        &
        Ace of Spades &  
        Two of Spades &  
        Three of Spades
     \\
        \rotatebox[origin=m]{90}{\parbox{0.24\textwidth}{\centering LRP}} &
        \frame{\includegraphics[width=0.24\textwidth]{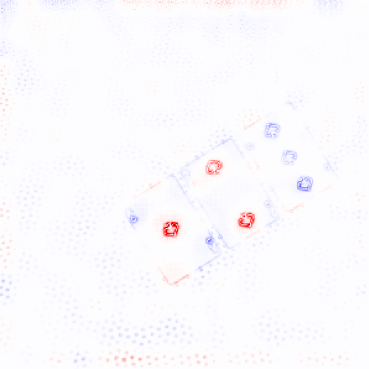}} &  
        \frame{\includegraphics[width=0.24\textwidth]{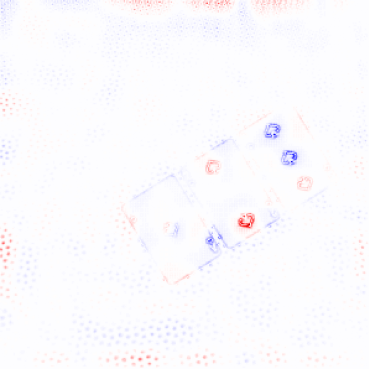}} &  
        \frame{\includegraphics[width=0.24\textwidth]{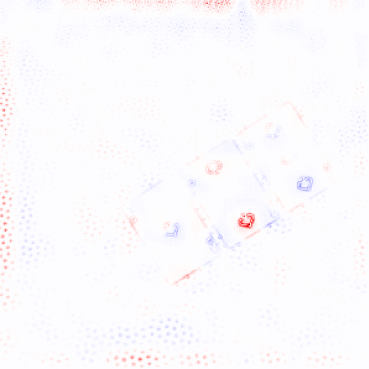}}
     \\
        \rotatebox[origin=m]{90}{\parbox{0.24\textwidth}{\centering IG with Smoothgrad}} &
        \frame{\includegraphics[width=0.24\textwidth]{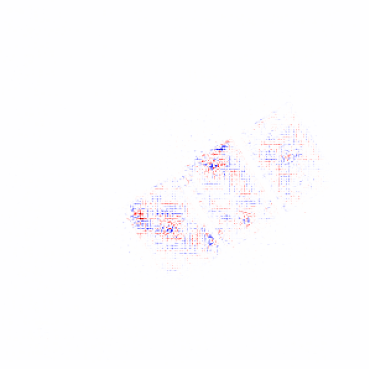}} &  
        \frame{\includegraphics[width=0.24\textwidth]{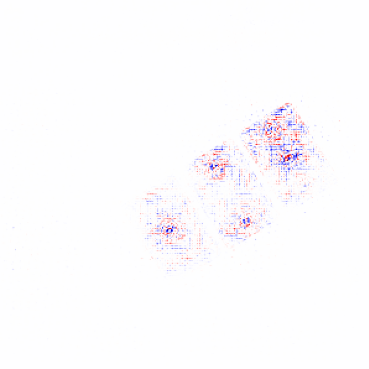}} &  
        \frame{\includegraphics[width=0.24\textwidth]{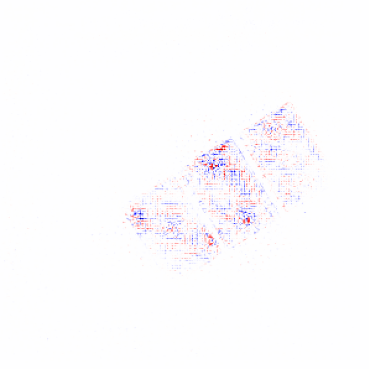}}  
     \\
        \rotatebox[origin=m]{90}{\parbox{0.24\textwidth}{\centering IG with Smoothgrad squared}} &  
        \frame{\includegraphics[width=0.24\textwidth]{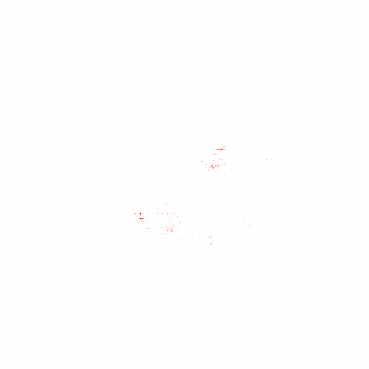}} &  
        \frame{\includegraphics[width=0.24\textwidth]{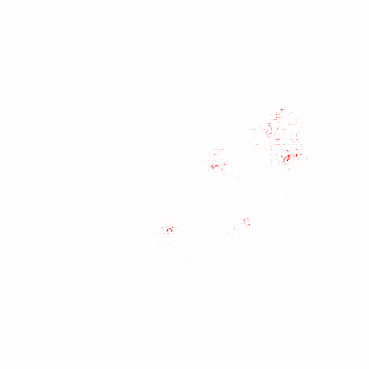}} &  
        \frame{\includegraphics[width=0.24\textwidth]{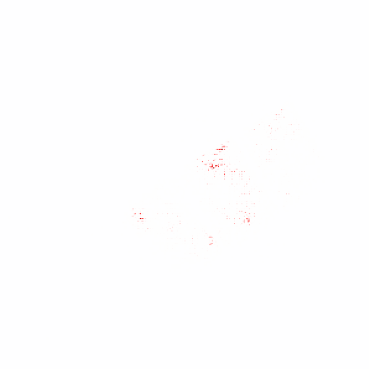}} 
     \\
    \end{tabular}
         \caption{Straight flush task classification}
         \label{fig:standard_poker_salience_appendix_2}
     \end{subfigure}

    \vspace{0.5cm}
    \caption{Standard NN model trained on poker cards}
\label{fig:standard_poker_salience_appendix}
\end{figure*}

We show additional concept saliency map results than those in section~\ref{results_section} for our instance-level playing card models in Figure~\ref{fig:indi_poker_salience_appendix}, \ref{fig:indi_three_salience_appendix}, \ref{fig:joint_poker_salience_appendix}, and \ref{fig:standard_poker_salience_appendix}. Most saliency maps focus on the playing card that are semantically meaningful to its corresponding concept. All saliency maps generated with LRP show positive relevancy attributed to the correct playing card for each concept to satisfy semantically meaningful feature mapping, while both IG variations show good localisation. IG with a smoothgrad squared noise tunnel applies little relevance in general and highlights a different card for the concept ``King of Hearts'' in Figure~\ref{fig:joint_poker_salience_appendix}.

\begin{figure*}[p]
     \centering
     \begin{subfigure}[b]{0.45\textwidth}
         \centering
         \begin{tabular}{b{0.1\textwidth} m{0.2\textwidth} m{0.2\textwidth} m{0.2\textwidth} }
        &
        input &
        &
     \\
        &
        \frame{\includegraphics[width=0.24\textwidth]{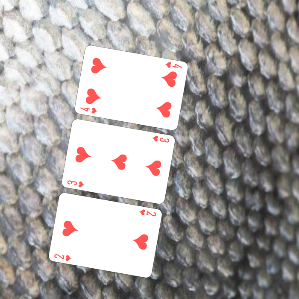}} & 
        &
     \\
        &
        Two of Hearts &  
        Three of Hearts &  
        Four of Hearts
     \\
        \rotatebox[origin=m]{90}{\parbox{0.24\textwidth}{\centering LRP}} &
        \frame{\includegraphics[width=0.24\textwidth]{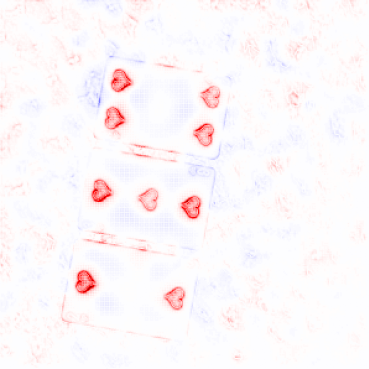}} &  
        \frame{\includegraphics[width=0.24\textwidth]{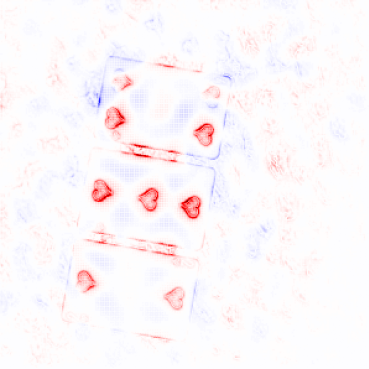}} &  
        \frame{\includegraphics[width=0.24\textwidth]{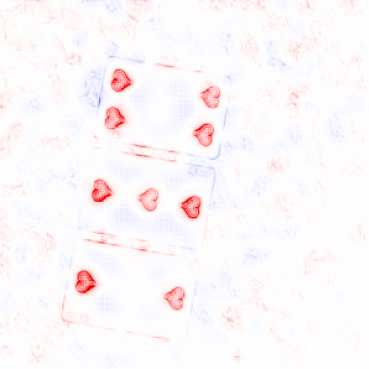}}
     \\
        \rotatebox[origin=m]{90}{\parbox{0.24\textwidth}{\centering IG with Smoothgrad}} &
        \frame{\includegraphics[width=0.24\textwidth]{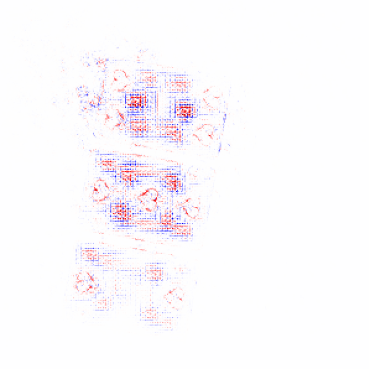}} &  
        \frame{\includegraphics[width=0.24\textwidth]{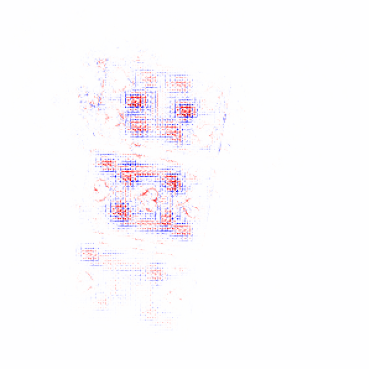}} &  
        \frame{\includegraphics[width=0.24\textwidth]{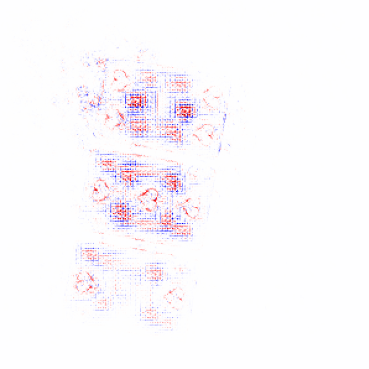}}  
     \\
        \rotatebox[origin=m]{90}{\parbox{0.24\textwidth}{\centering IG with Smoothgrad squared}} &  
        \frame{\includegraphics[width=0.24\textwidth]{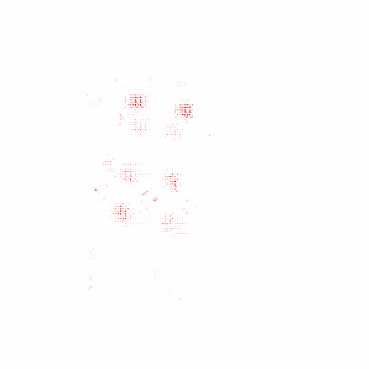}} &  
        \frame{\includegraphics[width=0.24\textwidth]{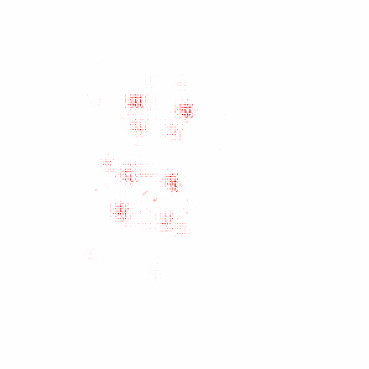}} &  
        \frame{\includegraphics[width=0.24\textwidth]{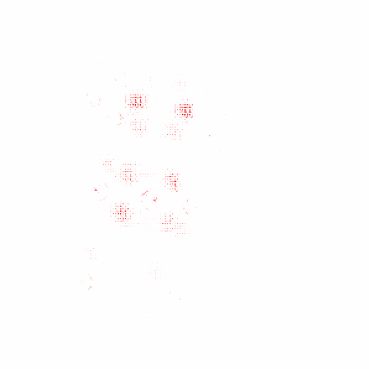}} 
     \\
    \end{tabular}
         \caption{}
         \label{fig:class_level_poker_salience_appendix_1}
     \end{subfigure}
     \hfill
     \begin{subfigure}[b]{0.45\textwidth}
         \centering
         \begin{tabular}{b{0.1\textwidth} m{0.2\textwidth} m{0.2\textwidth} m{0.2\textwidth} }
        &
        input &
        &
     \\
        &
        \frame{\includegraphics[width=0.24\textwidth]{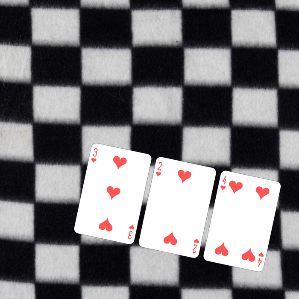}} & 
        &
     \\
        &
        Two of Hearts &  
        Three of Hearts &  
        Four of Hearts
     \\
        \rotatebox[origin=m]{90}{\parbox{0.24\textwidth}{\centering LRP}} &
        \frame{\includegraphics[width=0.24\textwidth]{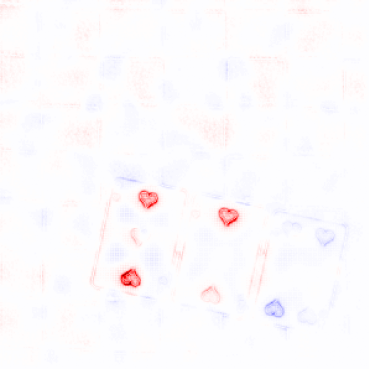}} &  
        \frame{\includegraphics[width=0.24\textwidth]{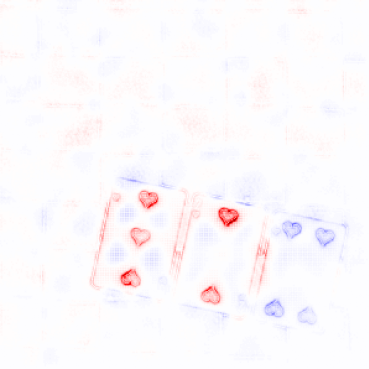}} &  
        \frame{\includegraphics[width=0.24\textwidth]{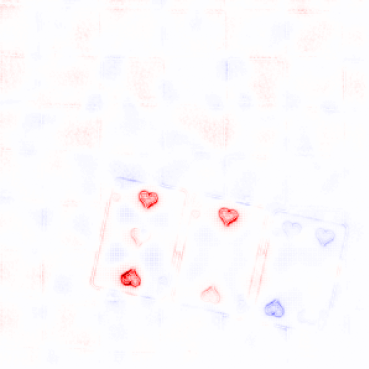}}
     \\
        \rotatebox[origin=m]{90}{\parbox{0.24\textwidth}{\centering IG with Smoothgrad}} &
        \frame{\includegraphics[width=0.24\textwidth]{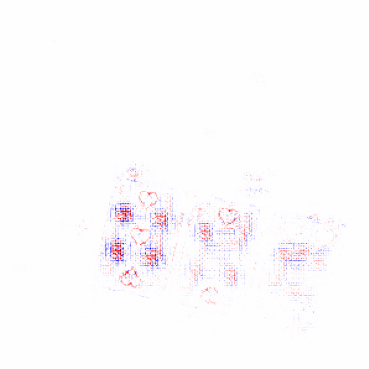}} &  
        \frame{\includegraphics[width=0.24\textwidth]{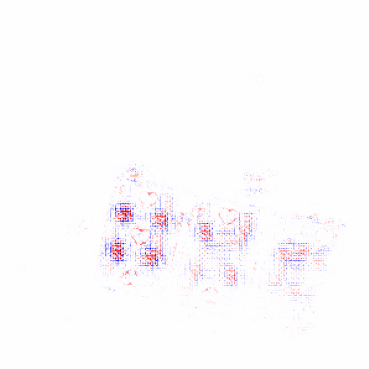}} &  
        \frame{\includegraphics[width=0.24\textwidth]{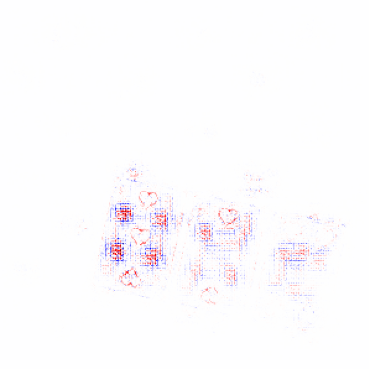}}  
     \\
        \rotatebox[origin=m]{90}{\parbox{0.24\textwidth}{\centering IG with Smoothgrad squared}} &  
        \frame{\includegraphics[width=0.24\textwidth]{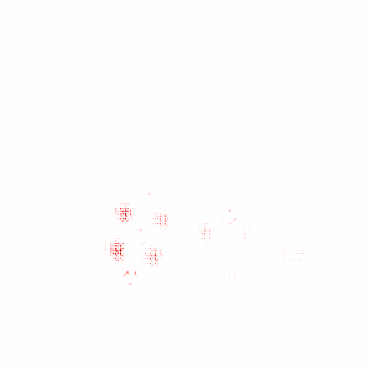}} &  
        \frame{\includegraphics[width=0.24\textwidth]{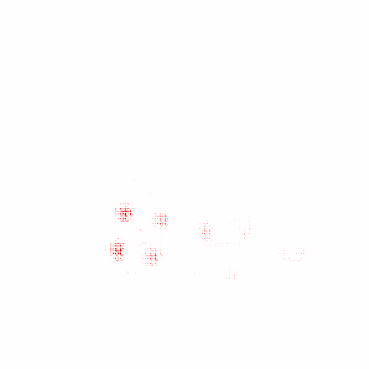}} &  
        \frame{\includegraphics[width=0.24\textwidth]{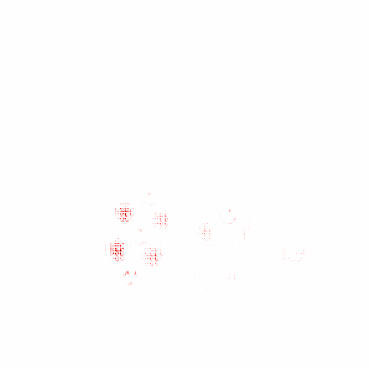}} 
     \\
    \end{tabular}
         \caption{}
         \label{fig:class_level_poker_salience_appendix_2}
     \end{subfigure}
     
    \vspace{0.5cm}
    \caption{Class-level poker cards predicting concepts with the task classification of Straight flush}
\label{fig:class_level_poker_sf_salience_appendix}
\end{figure*}

\begin{figure*}[p]
     \centering
     \begin{subfigure}[b]{0.45\textwidth}
         \centering
         \begin{tabular}{b{0.1\textwidth} m{0.2\textwidth} m{0.2\textwidth} m{0.2\textwidth} }
        &
        input &
        &
     \\
        &
        \frame{\includegraphics[width=0.24\textwidth]{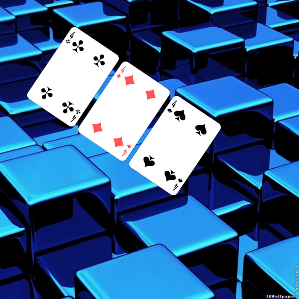}} & 
        &
     \\
        &
        Four of Clubs &  
        Four of Diamonds &  
        Four of Spades
     \\
        \rotatebox[origin=m]{90}{\parbox{0.24\textwidth}{\centering LRP}} &
        \frame{\includegraphics[width=0.24\textwidth]{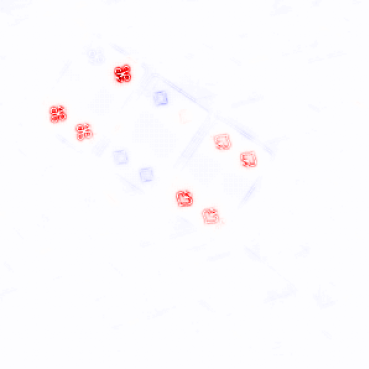}} &  
        \frame{\includegraphics[width=0.24\textwidth]{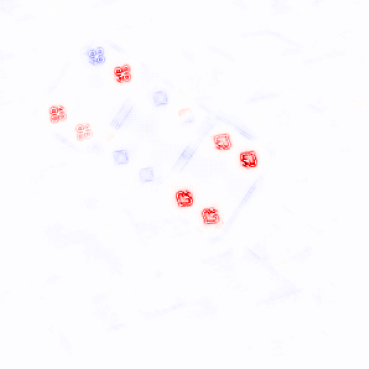}} &  
        \frame{\includegraphics[width=0.24\textwidth]{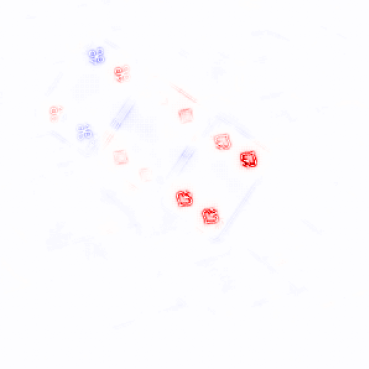}}
     \\
        \rotatebox[origin=m]{90}{\parbox{0.24\textwidth}{\centering IG with Smoothgrad}} &
        \frame{\includegraphics[width=0.24\textwidth]{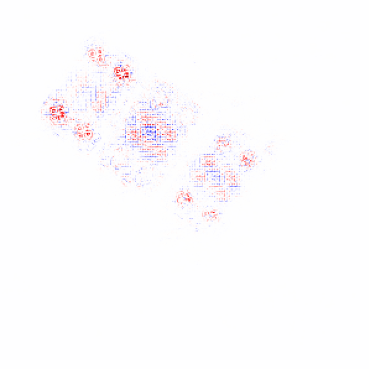}} &  
        \frame{\includegraphics[width=0.24\textwidth]{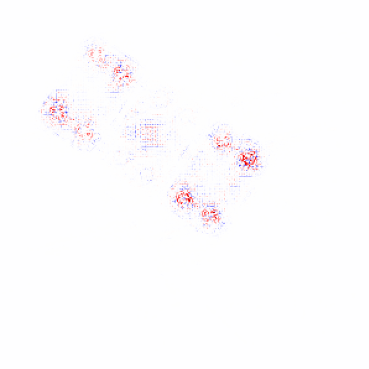}} &  
        \frame{\includegraphics[width=0.24\textwidth]{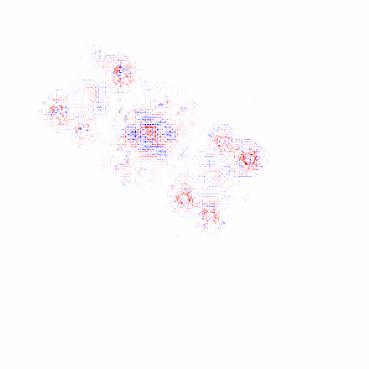}}  
     \\
        \rotatebox[origin=m]{90}{\parbox{0.24\textwidth}{\centering IG with Smoothgrad squared}} &  
        \frame{\includegraphics[width=0.24\textwidth]{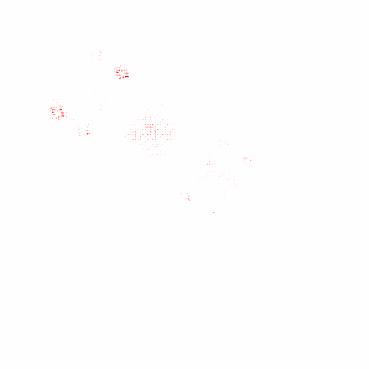}} &  
        \frame{\includegraphics[width=0.24\textwidth]{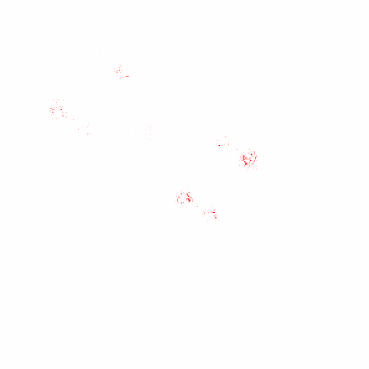}} &  
        \frame{\includegraphics[width=0.24\textwidth]{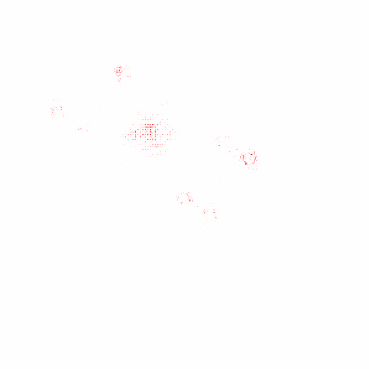}} 
     \\
    \end{tabular}
         \caption{}
         \label{fig:class_level_poker_salience_appendix_3}
     \end{subfigure}
     \hfill
     \begin{subfigure}[b]{0.45\textwidth}
         \centering
         \begin{tabular}{b{0.1\textwidth} m{0.2\textwidth} m{0.2\textwidth} m{0.2\textwidth} }
        &
        input &
        &
     \\
        &
        \frame{\includegraphics[width=0.24\textwidth]{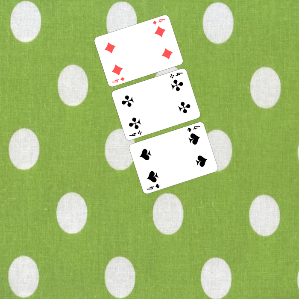}} & 
        &
     \\
        &
        Four of Clubs &  
        Four of Diamonds &  
        Four of Spades
     \\
        \rotatebox[origin=m]{90}{\parbox{0.24\textwidth}{\centering LRP}} &
        \frame{\includegraphics[width=0.24\textwidth]{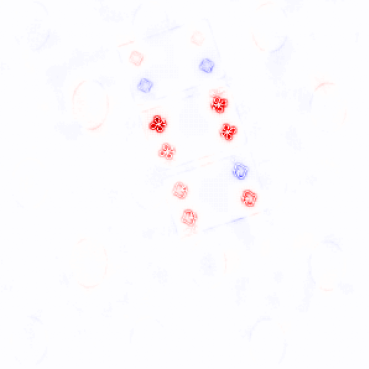}} &  
        \frame{\includegraphics[width=0.24\textwidth]{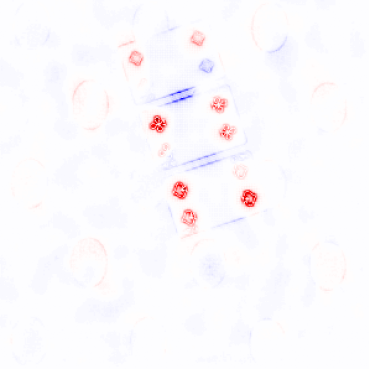}} &  
        \frame{\includegraphics[width=0.24\textwidth]{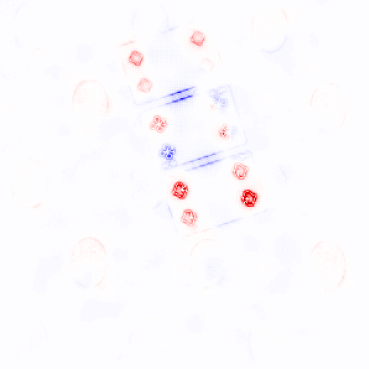}}
     \\
        \rotatebox[origin=m]{90}{\parbox{0.24\textwidth}{\centering IG with Smoothgrad}} &
        \frame{\includegraphics[width=0.24\textwidth]{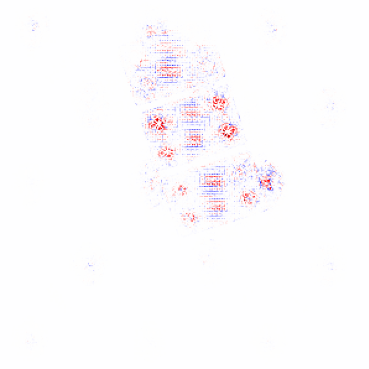}} &  
        \frame{\includegraphics[width=0.24\textwidth]{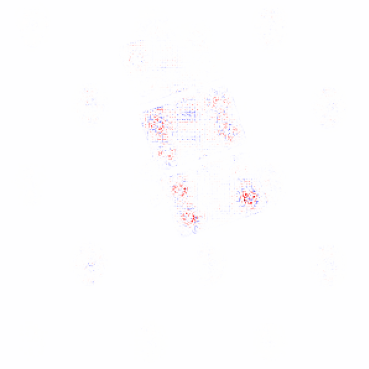}} &  
        \frame{\includegraphics[width=0.24\textwidth]{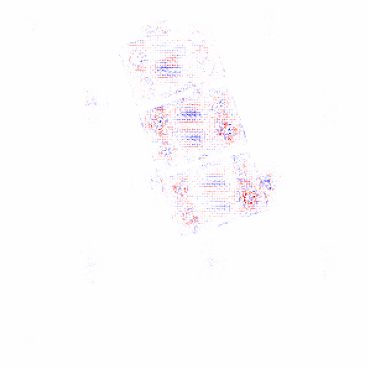}}  
     \\
        \rotatebox[origin=m]{90}{\parbox{0.24\textwidth}{\centering IG with Smoothgrad squared}} &  
        \frame{\includegraphics[width=0.24\textwidth]{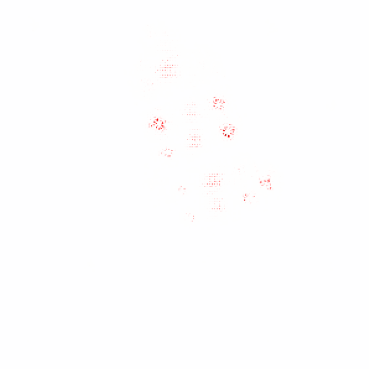}} &  
        \frame{\includegraphics[width=0.24\textwidth]{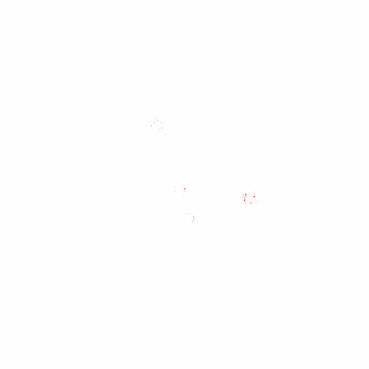}} &  
        \frame{\includegraphics[width=0.24\textwidth]{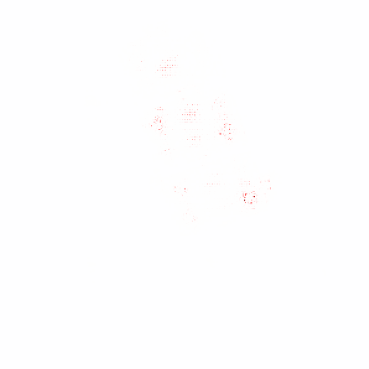}} 
     \\
    \end{tabular}
         \caption{}
         \label{fig:class_level_poker_salience_appendix_4}
     \end{subfigure}

    \vspace{0.5cm}
    \caption{Class-level poker cards predicting concepts with the task classification of Three of a kind}
\label{fig:class_level_poker_tk_salience_appendix}
\end{figure*}

\begin{figure*}[p]
     \centering
     \begin{subfigure}[b]{0.45\textwidth}
         \centering
         \begin{tabular}{b{0.1\textwidth} m{0.2\textwidth} m{0.2\textwidth} m{0.2\textwidth} }
        &
        input &
        &
     \\
        &
        \frame{\includegraphics[width=0.24\textwidth]{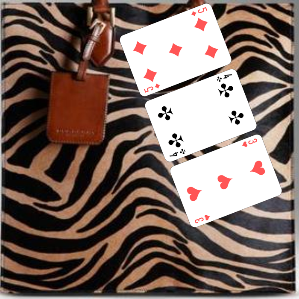}} & 
        &
     \\
        &
        Three of Hearts &  
        Four of Clubs &  
        Five of Diamonds
     \\
        \rotatebox[origin=m]{90}{\parbox{0.24\textwidth}{\centering LRP}} &
        \frame{\includegraphics[width=0.24\textwidth]{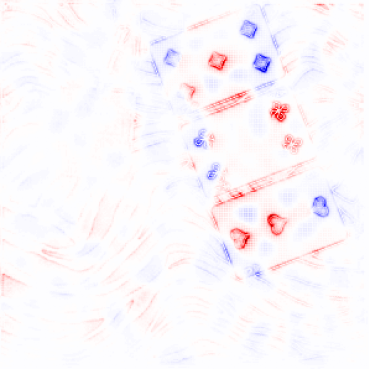}} &  
        \frame{\includegraphics[width=0.24\textwidth]{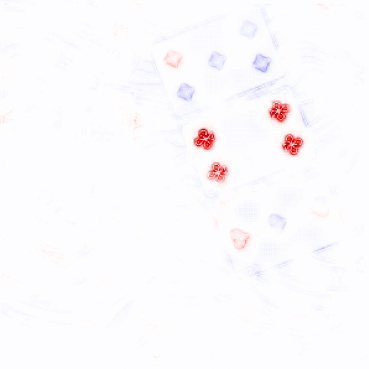}} &  
        \frame{\includegraphics[width=0.24\textwidth]{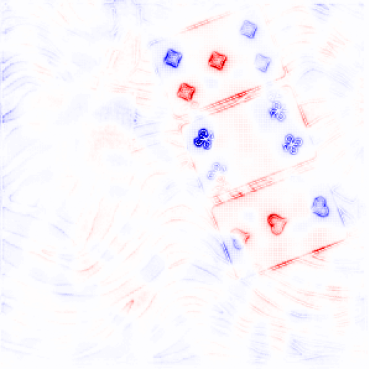}}
     \\
        \rotatebox[origin=m]{90}{\parbox{0.24\textwidth}{\centering IG with Smoothgrad}} &
        \frame{\includegraphics[width=0.24\textwidth]{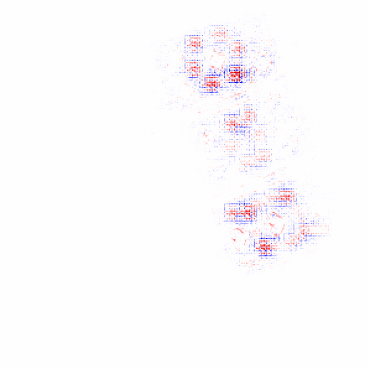}} &  
        \frame{\includegraphics[width=0.24\textwidth]{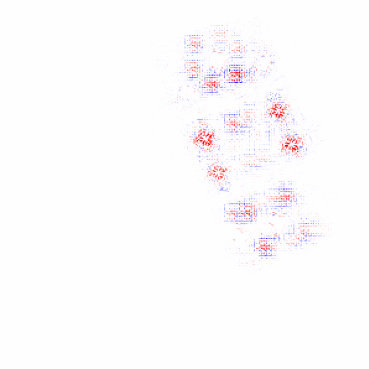}} &  
        \frame{\includegraphics[width=0.24\textwidth]{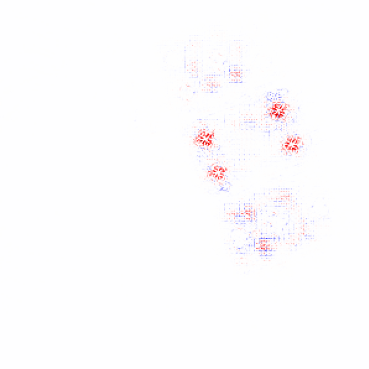}}  
     \\
        \rotatebox[origin=m]{90}{\parbox{0.24\textwidth}{\centering IG with Smoothgrad squared}} &  
        \frame{\includegraphics[width=0.24\textwidth]{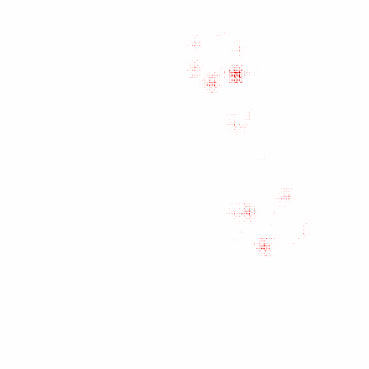}} &  
        \frame{\includegraphics[width=0.24\textwidth]{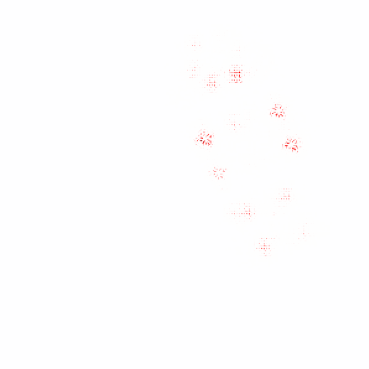}} &  
        \frame{\includegraphics[width=0.24\textwidth]{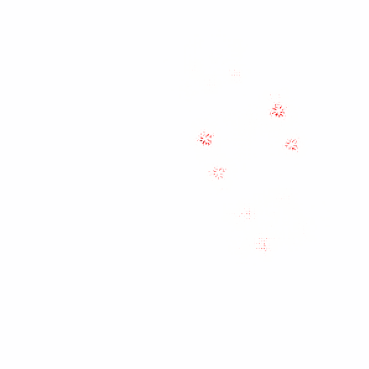}} 
     \\
    \end{tabular}
         \caption{}
         \label{fig:class_level_poker_salience_appendix_5}
     \end{subfigure}
     \hfill
     \begin{subfigure}[b]{0.45\textwidth}
         \centering
         \begin{tabular}{b{0.1\textwidth} m{0.2\textwidth} m{0.2\textwidth} m{0.2\textwidth} }
        &
        input &
        &
     \\
        &
        \frame{\includegraphics[width=0.24\textwidth]{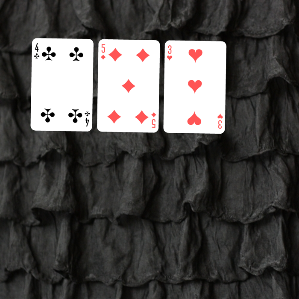}} & 
        &
     \\
        &
        Three of Hearts &  
        Four of Clubs &  
        Five of Diamonds
     \\
        \rotatebox[origin=m]{90}{\parbox{0.24\textwidth}{\centering LRP}} &
        \frame{\includegraphics[width=0.24\textwidth]{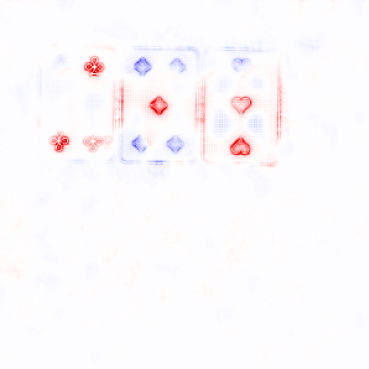}} &  
        \frame{\includegraphics[width=0.24\textwidth]{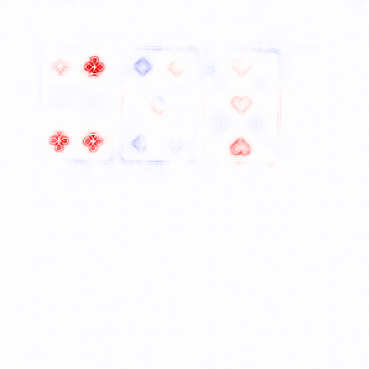}} &  
        \frame{\includegraphics[width=0.24\textwidth]{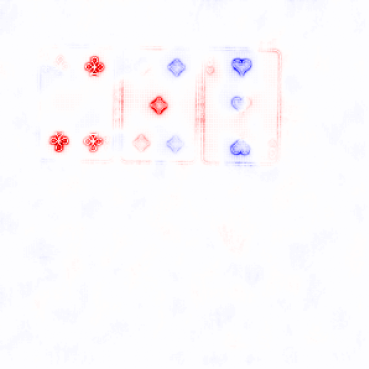}}
     \\
        \rotatebox[origin=m]{90}{\parbox{0.24\textwidth}{\centering IG with Smoothgrad}} &
        \frame{\includegraphics[width=0.24\textwidth]{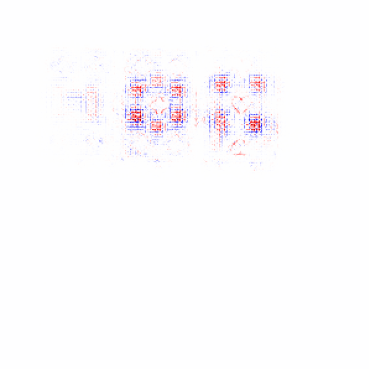}} &  
        \frame{\includegraphics[width=0.24\textwidth]{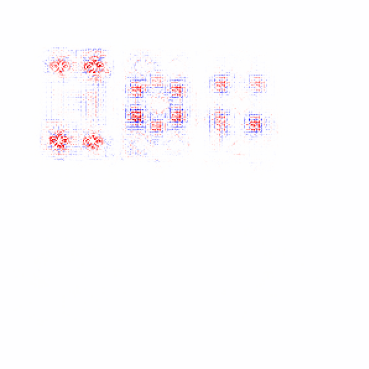}} &  
        \frame{\includegraphics[width=0.24\textwidth]{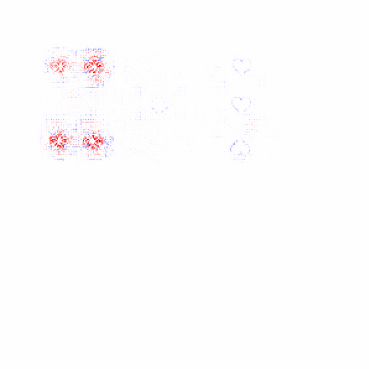}}  
     \\
        \rotatebox[origin=m]{90}{\parbox{0.24\textwidth}{\centering IG with Smoothgrad squared}} &  
        \frame{\includegraphics[width=0.24\textwidth]{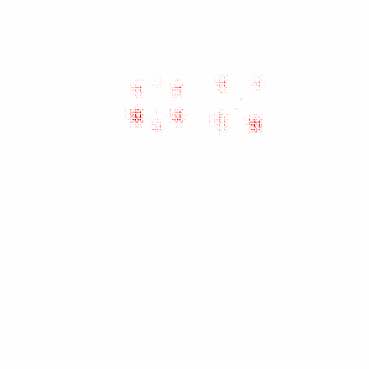}} &  
        \frame{\includegraphics[width=0.24\textwidth]{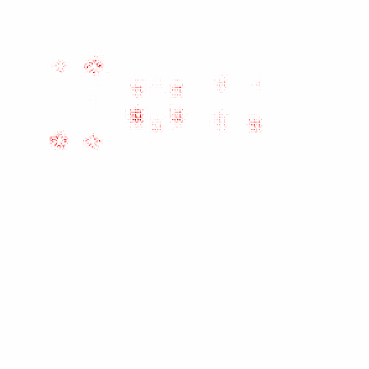}} &  
        \frame{\includegraphics[width=0.24\textwidth]{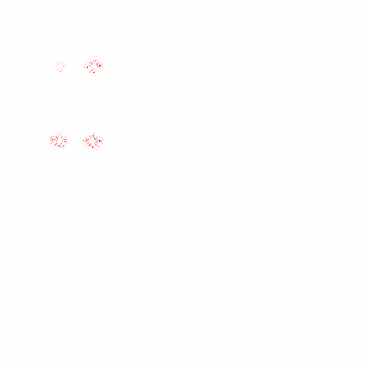}} 
     \\
    \end{tabular}
         \caption{}
         \label{fig:class_level_poker_salience_appendix_6}
     \end{subfigure}

    \vspace{0.5cm}
    \caption{Class-level poker cards predicting concepts with the task classification of Straight}
\label{fig:class_level_poker_s_salience_appendix}
\end{figure*}

\begin{figure*}[p]
     \centering
     \begin{subfigure}[b]{0.45\textwidth}
         \centering
         \begin{tabular}{b{0.1\textwidth} m{0.2\textwidth} m{0.2\textwidth} m{0.2\textwidth} }
        &
        input &
        &
     \\
        &
        \frame{\includegraphics[width=0.24\textwidth]{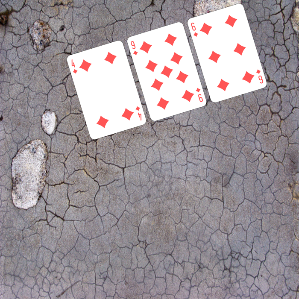}} & 
        &
     \\
        &
        Four of Diamonds &  
        Six of Diamonds &  
        Nine of Diamonds
     \\
        \rotatebox[origin=m]{90}{\parbox{0.24\textwidth}{\centering LRP}} &
        \frame{\includegraphics[width=0.24\textwidth]{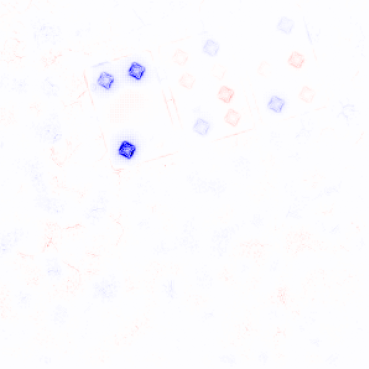}} &  
        \frame{\includegraphics[width=0.24\textwidth]{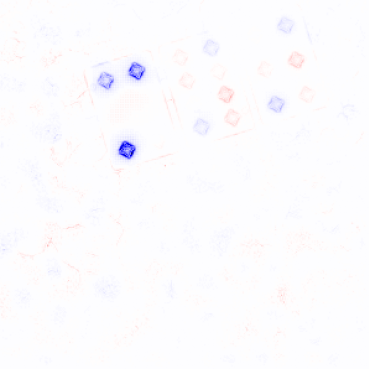}} &  
        \frame{\includegraphics[width=0.24\textwidth]{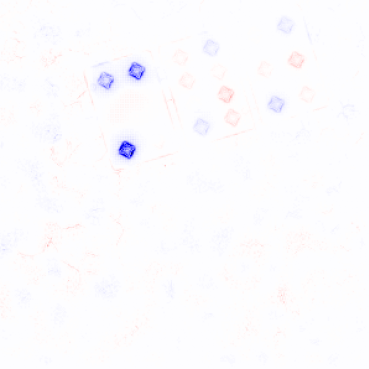}}
     \\
        \rotatebox[origin=m]{90}{\parbox{0.24\textwidth}{\centering IG with Smoothgrad}} &
        \frame{\includegraphics[width=0.24\textwidth]{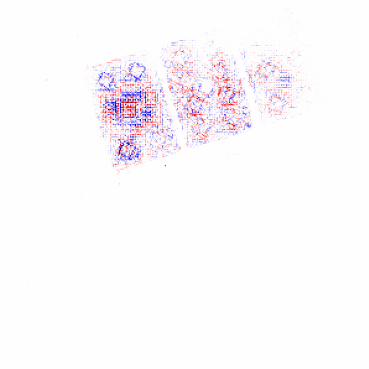}} &  
        \frame{\includegraphics[width=0.24\textwidth]{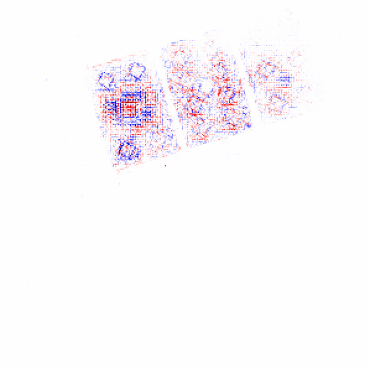}} &  
        \frame{\includegraphics[width=0.24\textwidth]{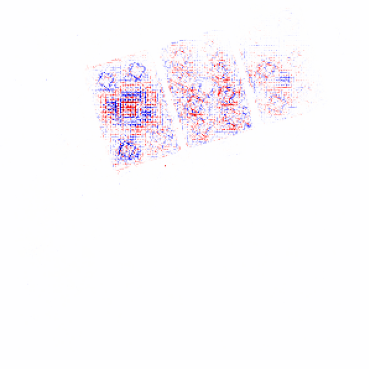}}  
     \\
        \rotatebox[origin=m]{90}{\parbox{0.24\textwidth}{\centering IG with Smoothgrad squared}} &  
        \frame{\includegraphics[width=0.24\textwidth]{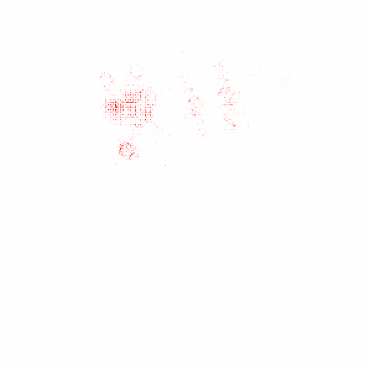}} &  
        \frame{\includegraphics[width=0.24\textwidth]{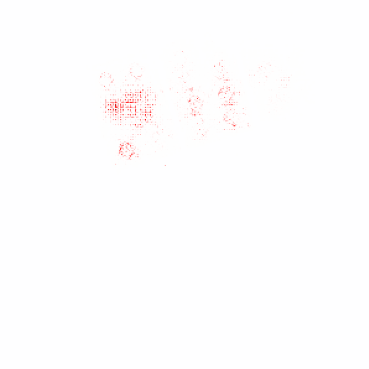}} &  
        \frame{\includegraphics[width=0.24\textwidth]{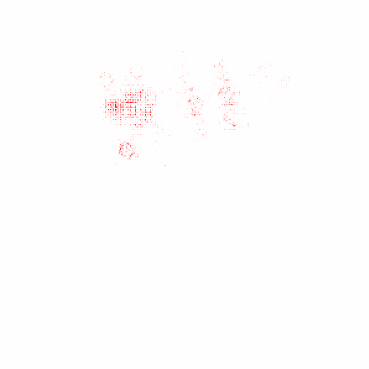}} 
     \\
    \end{tabular}
         \caption{}
         \label{fig:class_level_poker_salience_appendix_7}
     \end{subfigure}
     \hfill
     \begin{subfigure}[b]{0.45\textwidth}
         \centering
         \begin{tabular}{b{0.1\textwidth} m{0.2\textwidth} m{0.2\textwidth} m{0.2\textwidth} }
        &
        input &
        &
     \\
        &
        \frame{\includegraphics[width=0.24\textwidth]{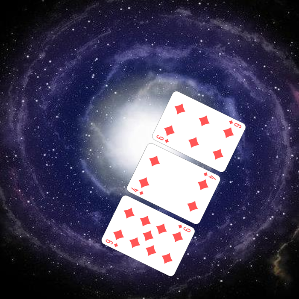}} & 
        &
     \\
        &
        Four of Diamonds &  
        Six of Diamonds &  
        Nine of Diamonds
     \\
        \rotatebox[origin=m]{90}{\parbox{0.24\textwidth}{\centering LRP}} &
        \frame{\includegraphics[width=0.24\textwidth]{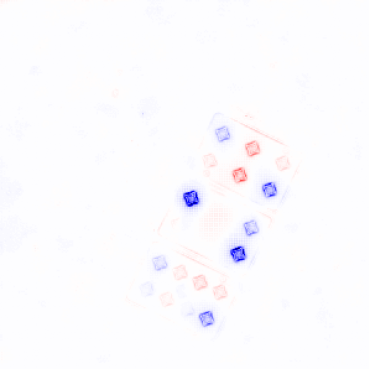}} &  
        \frame{\includegraphics[width=0.24\textwidth]{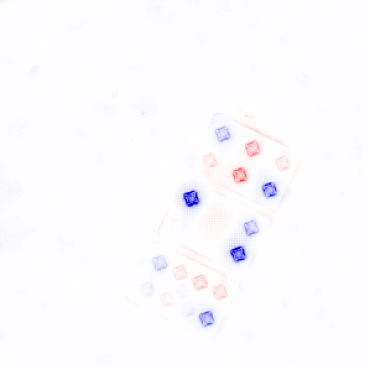}} &  
        \frame{\includegraphics[width=0.24\textwidth]{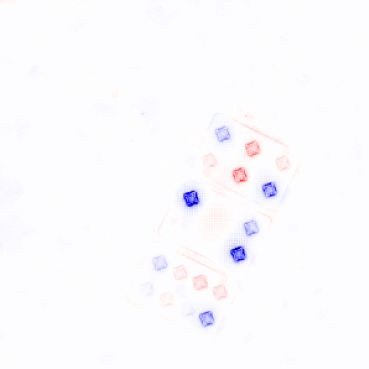}}
     \\
        \rotatebox[origin=m]{90}{\parbox{0.24\textwidth}{\centering IG with Smoothgrad}} &
        \frame{\includegraphics[width=0.24\textwidth]{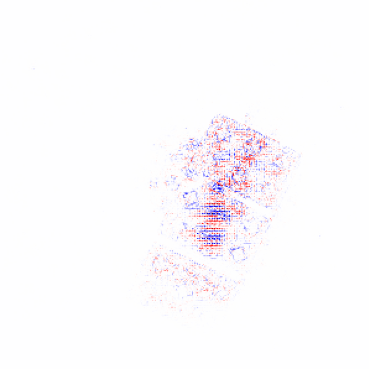}} &  
        \frame{\includegraphics[width=0.24\textwidth]{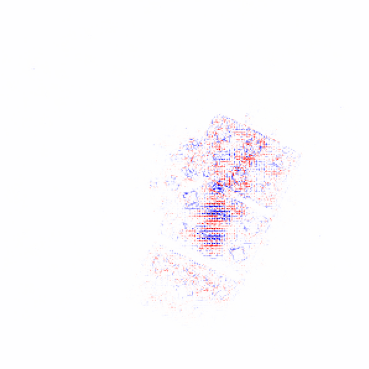}} &  
        \frame{\includegraphics[width=0.24\textwidth]{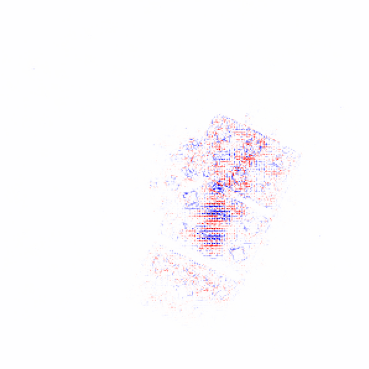}}  
     \\
        \rotatebox[origin=m]{90}{\parbox{0.24\textwidth}{\centering IG with Smoothgrad squared}} &  
        \frame{\includegraphics[width=0.24\textwidth]{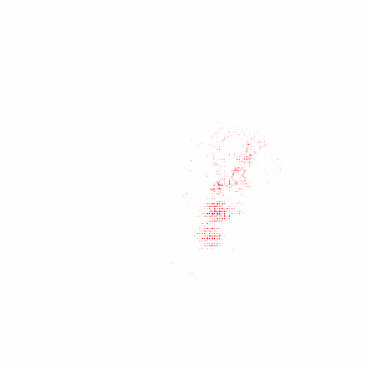}} &  
        \frame{\includegraphics[width=0.24\textwidth]{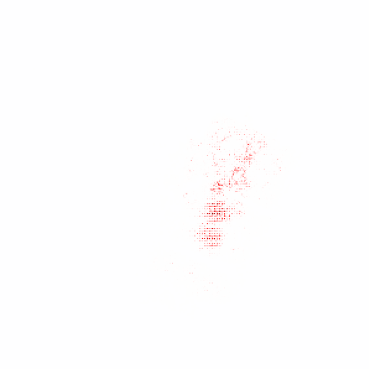}} &  
        \frame{\includegraphics[width=0.24\textwidth]{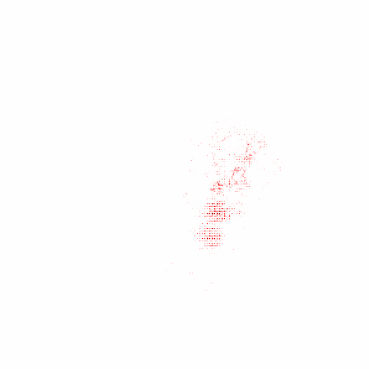}} 
     \\
    \end{tabular}
         \caption{}
         \label{fig:class_level_poker_salience_appendix_8}
     \end{subfigure}

    \vspace{0.5cm}
    \caption{Class-level poker cards predicting concepts with the task classification of Flush}
\label{fig:class_level_poker_f_salience_appendix}
\end{figure*}

\begin{figure*}[p]
     \centering
     \begin{subfigure}[b]{0.45\textwidth}
         \centering
         \begin{tabular}{b{0.1\textwidth} m{0.2\textwidth} m{0.2\textwidth} m{0.2\textwidth} }
        &
        input &
        &
     \\
        &
        \frame{\includegraphics[width=0.24\textwidth]{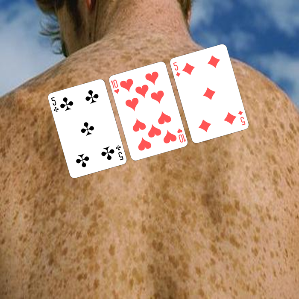}} & 
        &
     \\
        &
        Ten of Hearts &  
        Five of Clubs &  
        Five of Diamonds
     \\
        \rotatebox[origin=m]{90}{\parbox{0.24\textwidth}{\centering LRP}} &
        \frame{\includegraphics[width=0.24\textwidth]{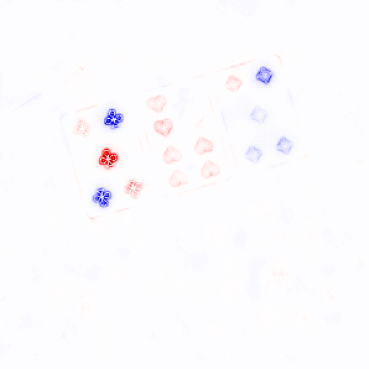}} &  
        \frame{\includegraphics[width=0.24\textwidth]{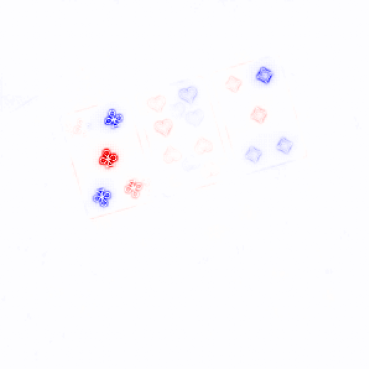}} &  
        \frame{\includegraphics[width=0.24\textwidth]{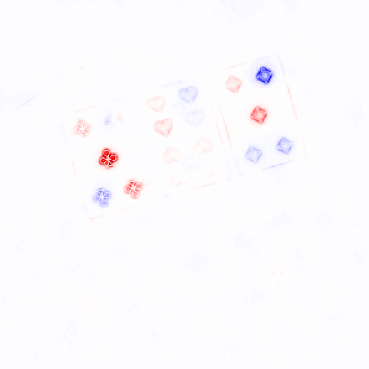}}
     \\
        \rotatebox[origin=m]{90}{\parbox{0.24\textwidth}{\centering IG with Smoothgrad}} &
        \frame{\includegraphics[width=0.24\textwidth]{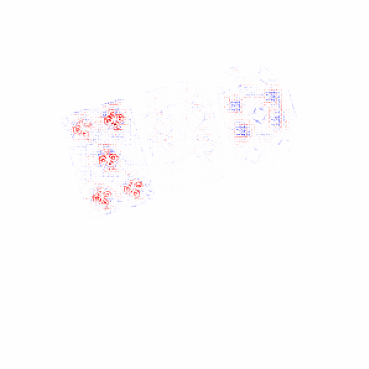}} &  
        \frame{\includegraphics[width=0.24\textwidth]{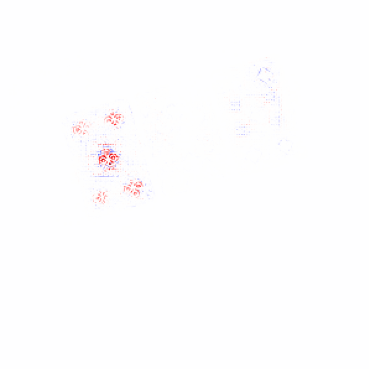}} &  
        \frame{\includegraphics[width=0.24\textwidth]{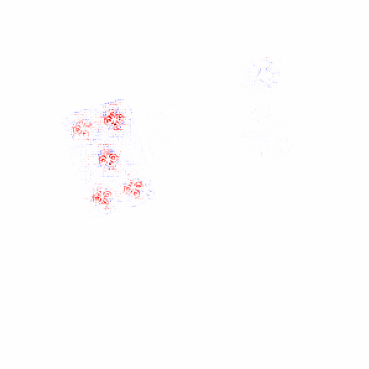}}  
     \\
        \rotatebox[origin=m]{90}{\parbox{0.24\textwidth}{\centering IG with Smoothgrad squared}} &  
        \frame{\includegraphics[width=0.24\textwidth]{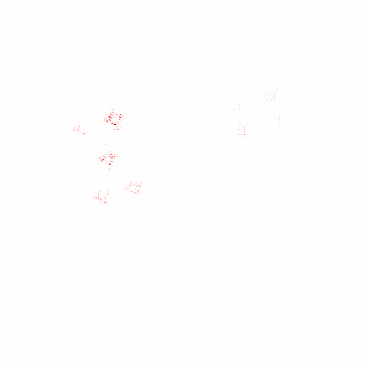}} &  
        \frame{\includegraphics[width=0.24\textwidth]{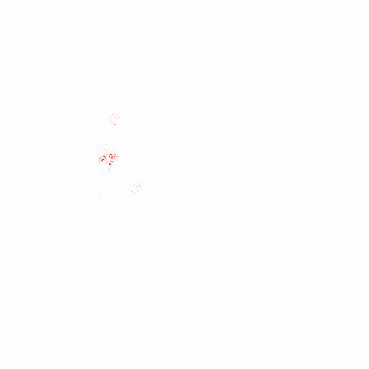}} &  
        \frame{\includegraphics[width=0.24\textwidth]{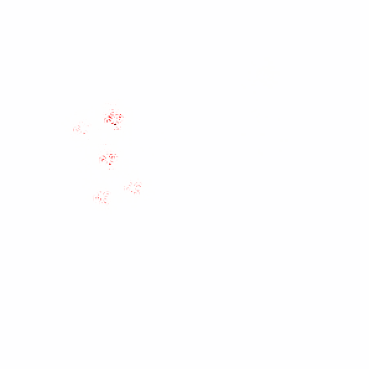}} 
     \\
    \end{tabular}
         \caption{}
         \label{fig:class_level_poker_salience_appendix_9}
     \end{subfigure}
     \hfill
     \begin{subfigure}[b]{0.45\textwidth}
         \centering
         \begin{tabular}{b{0.1\textwidth} m{0.2\textwidth} m{0.2\textwidth} m{0.2\textwidth} }
        &
        input &
        &
     \\
        &
        \frame{\includegraphics[width=0.24\textwidth]{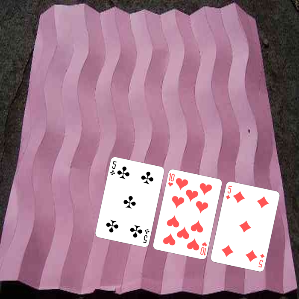}} & 
        &
     \\
        &
        Ten of Hearts &  
        Five of Clubs &  
        Five of Diamonds
     \\
        \rotatebox[origin=m]{90}{\parbox{0.24\textwidth}{\centering LRP}} &
        \frame{\includegraphics[width=0.24\textwidth]{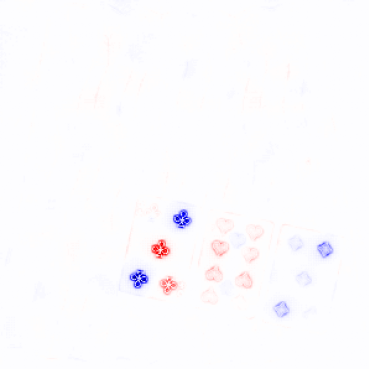}} &  
        \frame{\includegraphics[width=0.24\textwidth]{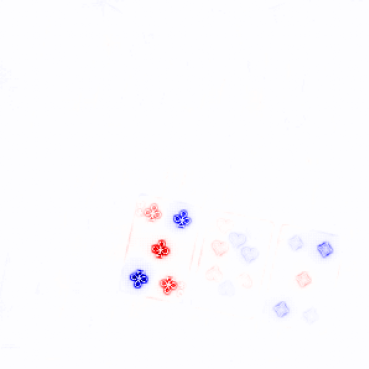}} &  
        \frame{\includegraphics[width=0.24\textwidth]{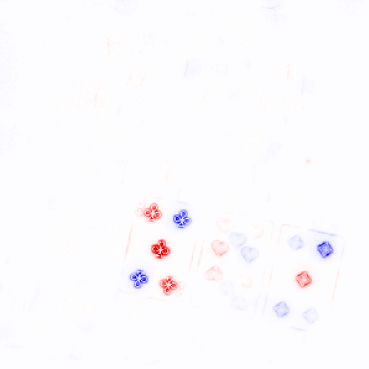}}
     \\
        \rotatebox[origin=m]{90}{\parbox{0.24\textwidth}{\centering IG with Smoothgrad}} &
        \frame{\includegraphics[width=0.24\textwidth]{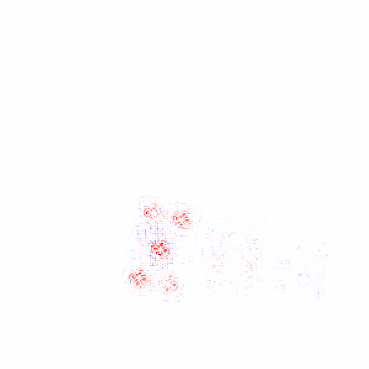}} &  
        \frame{\includegraphics[width=0.24\textwidth]{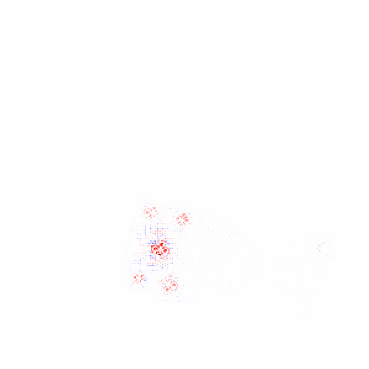}} &  
        \frame{\includegraphics[width=0.24\textwidth]{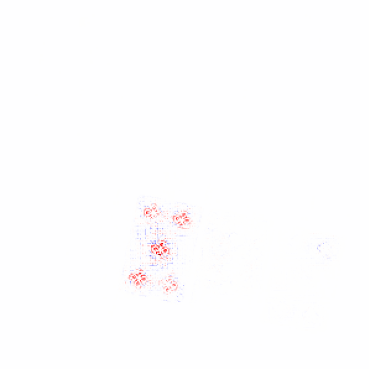}}  
     \\
        \rotatebox[origin=m]{90}{\parbox{0.24\textwidth}{\centering IG with Smoothgrad squared}} &  
        \frame{\includegraphics[width=0.24\textwidth]{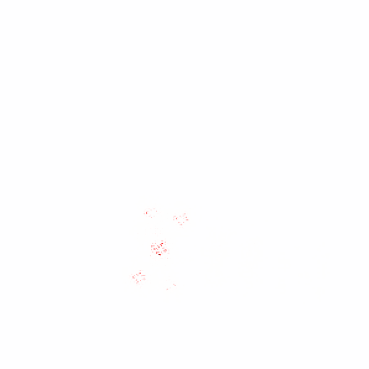}} &  
        \frame{\includegraphics[width=0.24\textwidth]{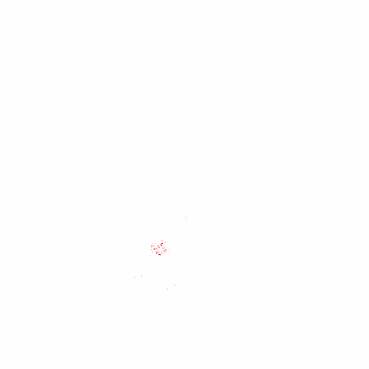}} &  
        \frame{\includegraphics[width=0.24\textwidth]{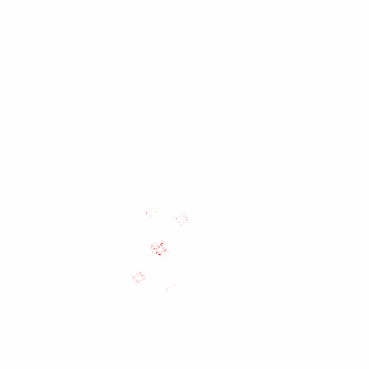}} 
     \\
    \end{tabular}
         \caption{}
         \label{fig:class_level_poker_salience_appendix_10}
     \end{subfigure}

    \vspace{0.5cm}
    \caption{Class-level poker cards predicting concepts with the task classification of Pair}
\label{fig:class_level_poker_p_salience_appendix}
\end{figure*}

\begin{figure*}[p]
     \centering
     \begin{subfigure}[b]{0.45\textwidth}
         \centering
         \begin{tabular}{b{0.1\textwidth} m{0.2\textwidth} m{0.2\textwidth} m{0.2\textwidth} }
        &
        input &
        &
     \\
        &
        \frame{\includegraphics[width=0.24\textwidth]{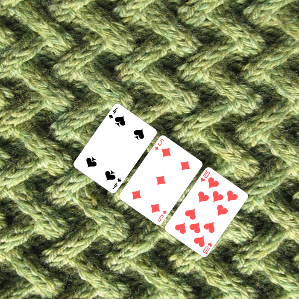}} & 
        &
     \\
        &
        Ten of Hearts &  
        Four of Spades &  
        Five of Diamonds
     \\
        \rotatebox[origin=m]{90}{\parbox{0.24\textwidth}{\centering LRP}} &
        \frame{\includegraphics[width=0.24\textwidth]{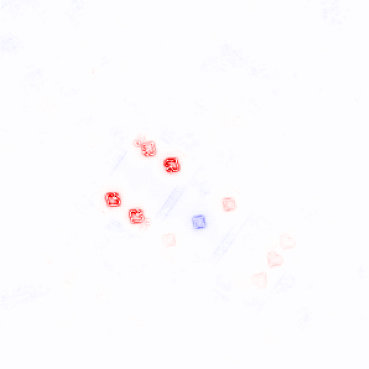}} &  
        \frame{\includegraphics[width=0.24\textwidth]{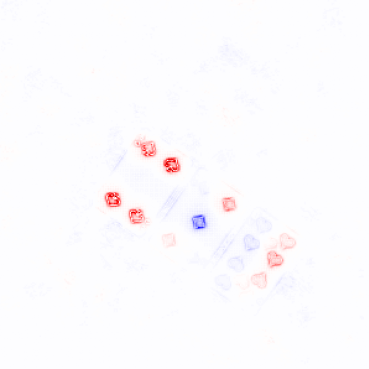}} &  
        \frame{\includegraphics[width=0.24\textwidth]{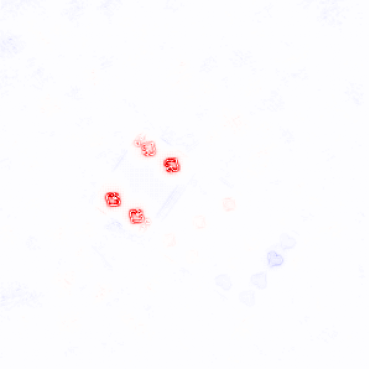}}
     \\
        \rotatebox[origin=m]{90}{\parbox{0.24\textwidth}{\centering IG with Smoothgrad}} &
        \frame{\includegraphics[width=0.24\textwidth]{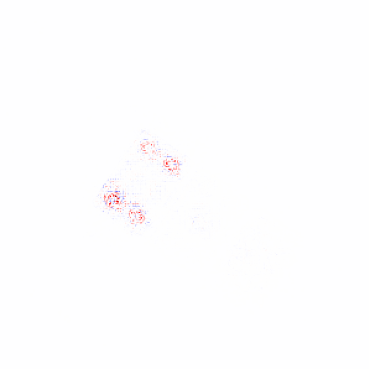}} &  
        \frame{\includegraphics[width=0.24\textwidth]{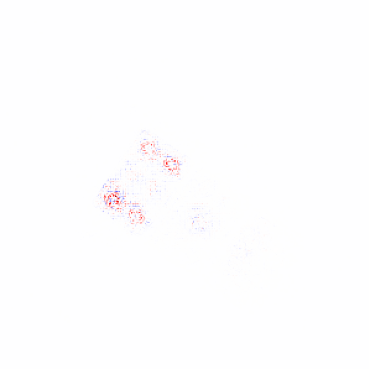}} &  
        \frame{\includegraphics[width=0.24\textwidth]{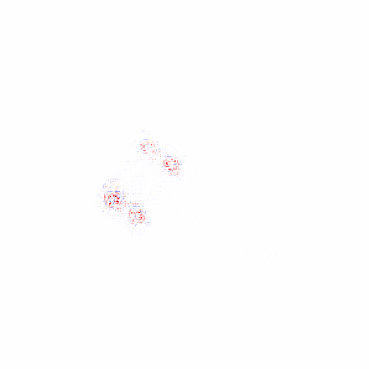}}  
     \\
        \rotatebox[origin=m]{90}{\parbox{0.24\textwidth}{\centering IG with Smoothgrad squared}} &  
        \frame{\includegraphics[width=0.24\textwidth]{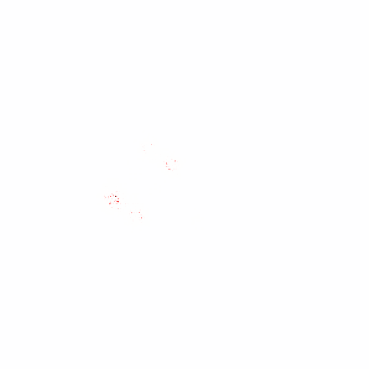}} &  
        \frame{\includegraphics[width=0.24\textwidth]{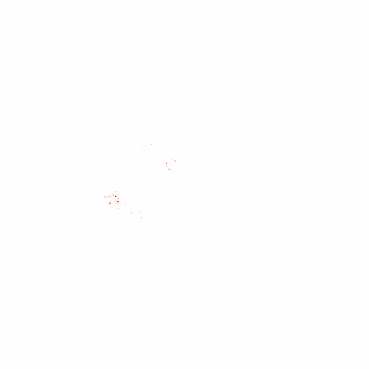}} &  
        \frame{\includegraphics[width=0.24\textwidth]{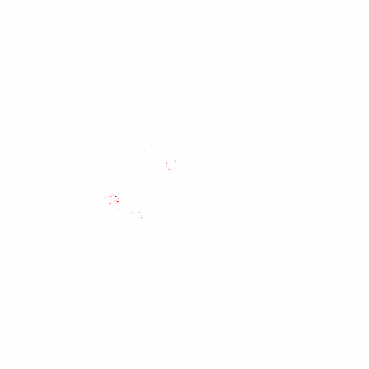}} 
     \\
    \end{tabular}
         \caption{}
         \label{fig:class_level_poker_salience_appendix_11}
     \end{subfigure}
     \hfill
     \begin{subfigure}[b]{0.45\textwidth}
         \centering
         \begin{tabular}{b{0.1\textwidth} m{0.2\textwidth} m{0.2\textwidth} m{0.2\textwidth} }
        &
        input &
        &
     \\
        &
        \frame{\includegraphics[width=0.24\textwidth]{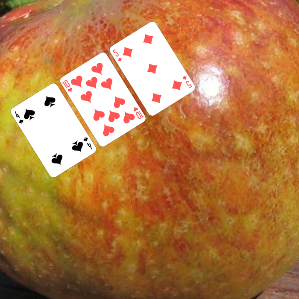}} & 
        &
     \\
        &
        Ten of Hearts &  
        Four of Spades &  
        Five of Diamonds
     \\
        \rotatebox[origin=m]{90}{\parbox{0.24\textwidth}{\centering LRP}} &
        \frame{\includegraphics[width=0.24\textwidth]{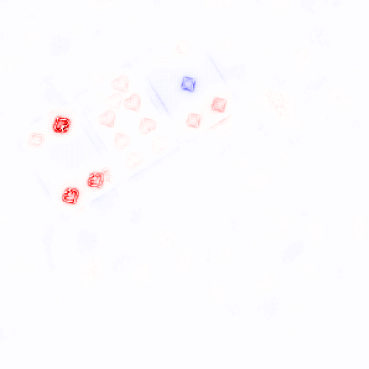}} &  
        \frame{\includegraphics[width=0.24\textwidth]{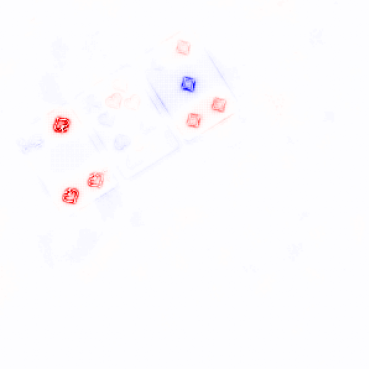}} &  
        \frame{\includegraphics[width=0.24\textwidth]{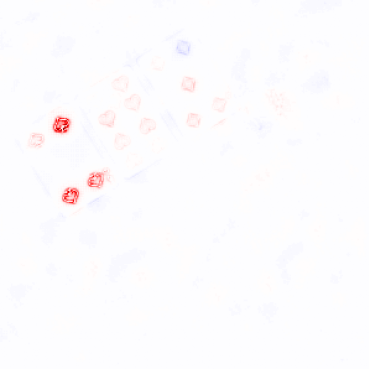}}
     \\
        \rotatebox[origin=m]{90}{\parbox{0.24\textwidth}{\centering IG with Smoothgrad}} &
        \frame{\includegraphics[width=0.24\textwidth]{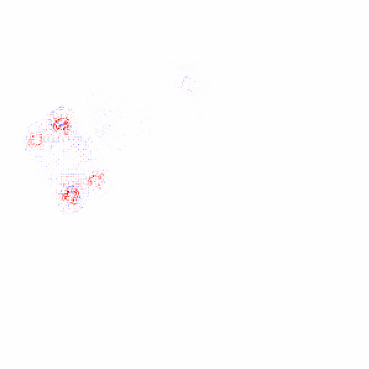}} &  
        \frame{\includegraphics[width=0.24\textwidth]{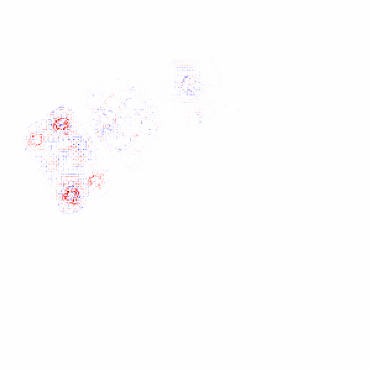}} &  
        \frame{\includegraphics[width=0.24\textwidth]{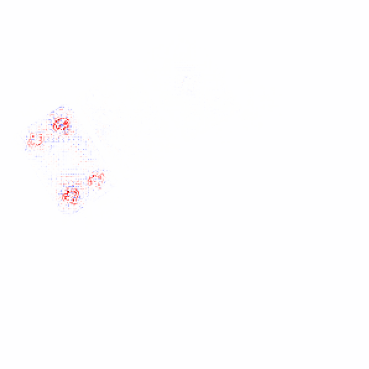}}  
     \\
        \rotatebox[origin=m]{90}{\parbox{0.24\textwidth}{\centering IG with Smoothgrad squared}} &  
        \frame{\includegraphics[width=0.24\textwidth]{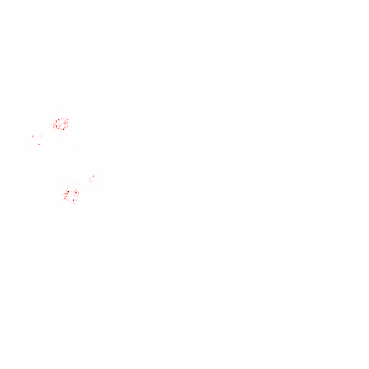}} &  
        \frame{\includegraphics[width=0.24\textwidth]{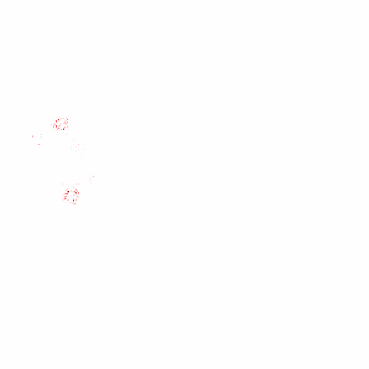}} &  
        \frame{\includegraphics[width=0.24\textwidth]{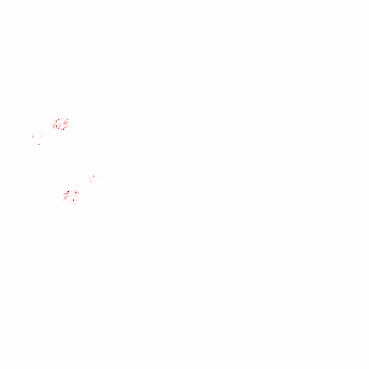}} 
     \\
    \end{tabular}
         \caption{}
         \label{fig:class_level_poker_salience_appendix_12}
     \end{subfigure}
     
    \vspace{0.5cm}
    \caption{Class-level poker cards predicting concepts with the task classification of High card}
\label{fig:class_level_poker_hc_salience_appendix}
\end{figure*}

We provide examples of all task classes for class-level poker card saliency maps to show variations between all concept predictions. These can be seen in Figure~\ref{fig:class_level_poker_sf_salience_appendix}, \ref{fig:class_level_poker_tk_salience_appendix}, \ref{fig:class_level_poker_s_salience_appendix}, \ref{fig:class_level_poker_f_salience_appendix}, \ref{fig:class_level_poker_p_salience_appendix} and \ref{fig:class_level_poker_hc_salience_appendix}. Most concepts do not apply relevance to the semantically meaningful input features. An exception is the concept Four of Clubs as seen in Figure~\ref{fig:class_level_poker_s_salience_appendix}.

\subsection{CheXpert} \label{CheXpert_saliency_appendix}

\begin{figure*}[p]

%sample: 15

\centering
\begin{tabular}{m{0.03\textwidth} m{0.15\textwidth} m{0.15\textwidth} m{0.15\textwidth} m{0.15\textwidth} m{0.15\textwidth} }
    &
    Cardiomegaly &
    Consolidation &
    Lung Opacity & 
    Plural Effusion &
    Support Devices
 \\
    \rotatebox[origin=l]{90}{\parbox{0.16\textwidth}{IG with Smoothgrad squared}} &
    \frame{\includegraphics[width=0.16\textwidth]{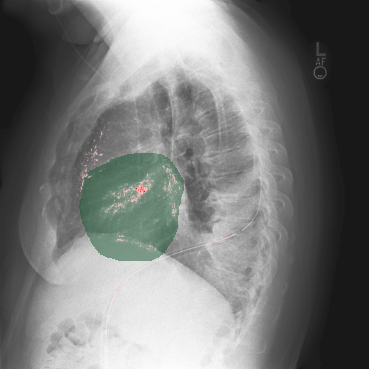}} &  
    \frame{\includegraphics[width=0.16\textwidth]{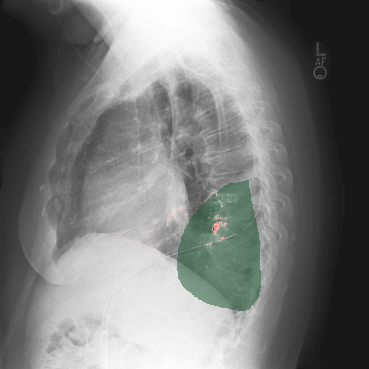}} &  
    \frame{\includegraphics[width=0.16\textwidth]{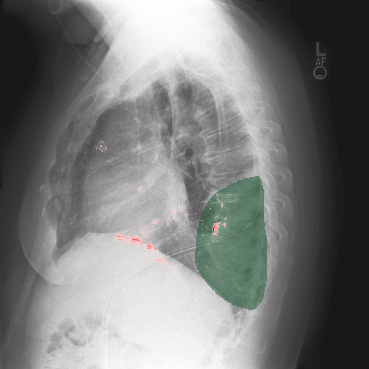}} &  
    \frame{\includegraphics[width=0.16\textwidth]{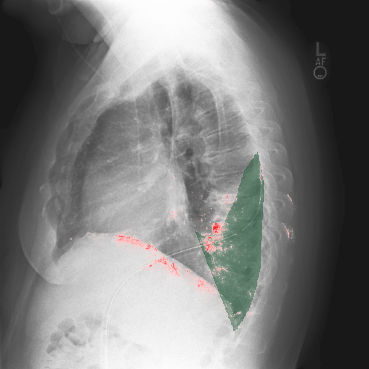}} &  
    \frame{\includegraphics[width=0.16\textwidth]{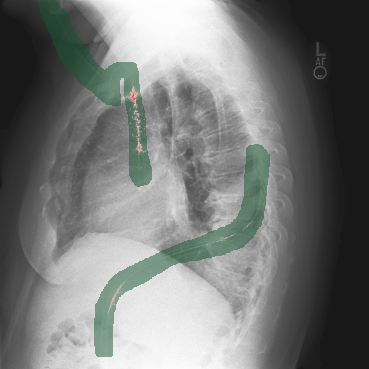}}
 \\
    \rotatebox[origin=l]{90}{\parbox{0.16\textwidth}{IG with Smoothgrad}} &
    \frame{\includegraphics[width=0.16\textwidth]{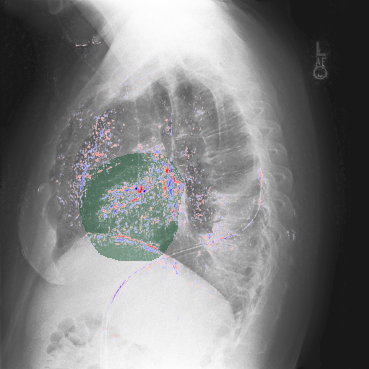}} &  
    \frame{\includegraphics[width=0.16\textwidth]{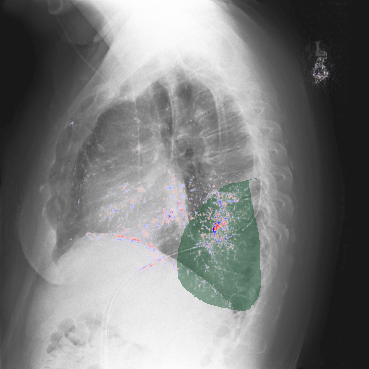}} &  
    \frame{\includegraphics[width=0.16\textwidth]{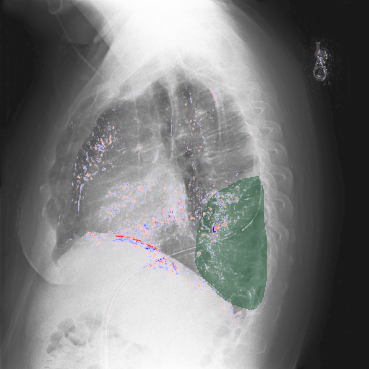}} &  
    \frame{\includegraphics[width=0.16\textwidth]{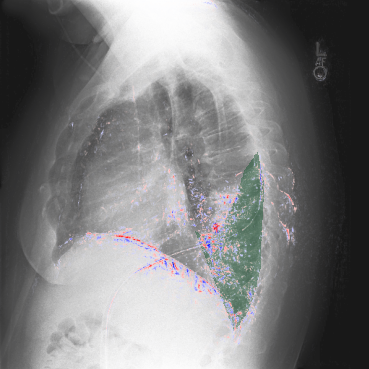}} &  
    \frame{\includegraphics[width=0.16\textwidth]{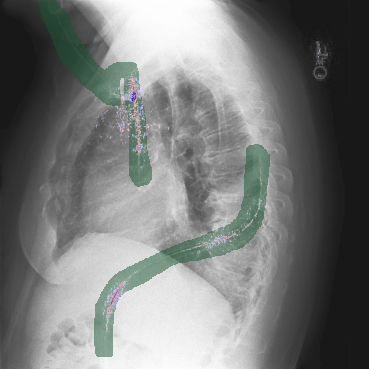}}
 \\
\end{tabular}

\caption{Saliency maps from a instance-level CheXpert trained using the joint CBM method. The number beneath the saliency map is the concept prediction which is in the range of 0 and 1.}
\label{fig:CBM chexpert concept saliency appendix 1}
\end{figure*}

\begin{figure*}[p]

%sample: 194

\centering
\begin{tabular}{m{0.03\textwidth} m{0.15\textwidth} m{0.15\textwidth} m{0.15\textwidth} m{0.15\textwidth} m{0.15\textwidth} }
    &
    Atelectasis &
    Cardiomegaly &
    Edema &  
    Lung Opacity &  
    Pleural Effusion
 \\
    \rotatebox[origin=l]{90}{\parbox{0.16\textwidth}{IG with Smoothgrad squared}} &
    \frame{\includegraphics[width=0.16\textwidth]{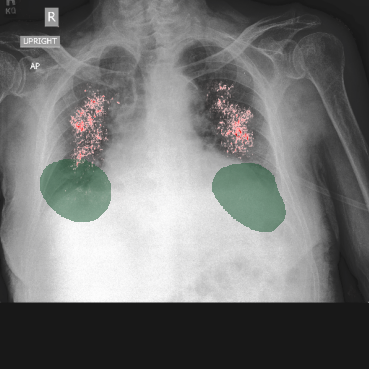}} &  
    \frame{\includegraphics[width=0.16\textwidth]{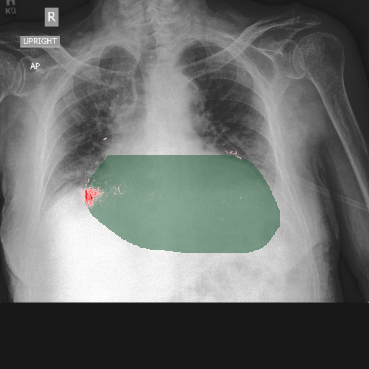}} &  
    \frame{\includegraphics[width=0.16\textwidth]{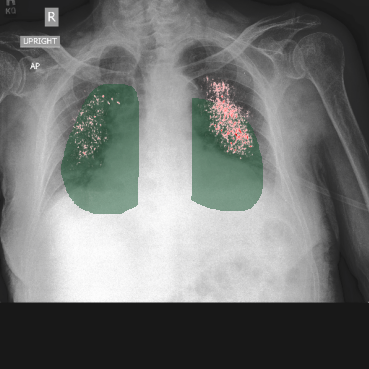}} &  
    \frame{\includegraphics[width=0.16\textwidth]{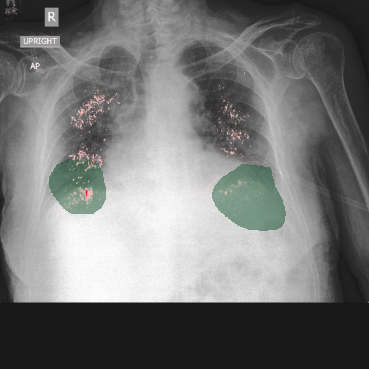}} &  
    \frame{\includegraphics[width=0.16\textwidth]{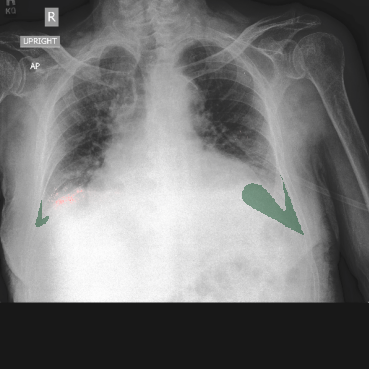}}
 \\
    \rotatebox[origin=l]{90}{\parbox{0.16\textwidth}{IG with Smoothgrad}} &
    \frame{\includegraphics[width=0.16\textwidth]{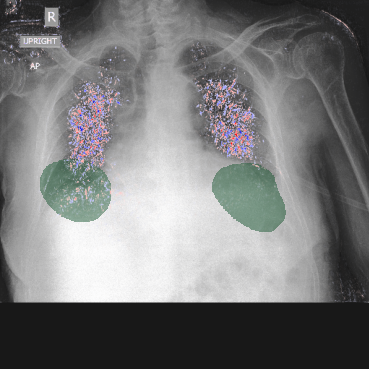}} &  
    \frame{\includegraphics[width=0.16\textwidth]{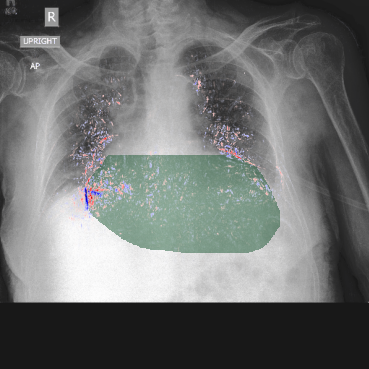}} &  
    \frame{\includegraphics[width=0.16\textwidth]{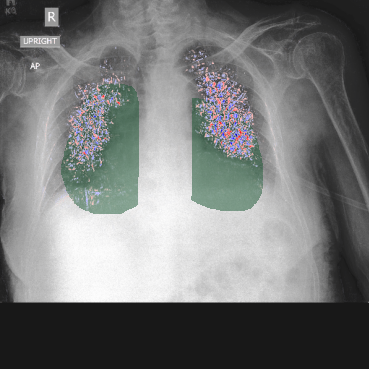}} &  
    \frame{\includegraphics[width=0.16\textwidth]{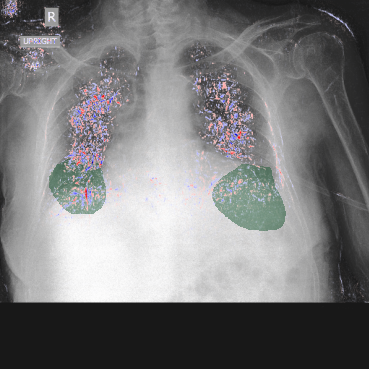}} &  
    \frame{\includegraphics[width=0.16\textwidth]{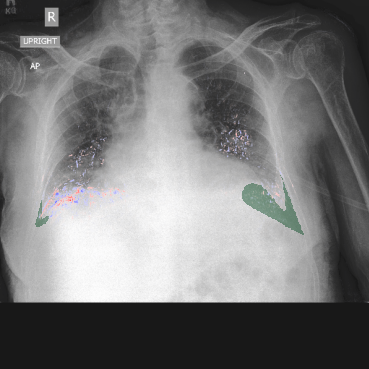}}
 \\
\end{tabular}

\caption{Saliency maps from a instance-level CheXpert trained using the joint CBM method. The number beneath the saliency map is the concept prediction which is in the range of 0 and 1.}
\label{fig:CBM chexpert concept saliency appendix 2}
\end{figure*}

\begin{figure*}[p]

\centering
\begin{tabular}{m{0.03\textwidth} m{0.15\textwidth} m{0.15\textwidth} m{0.15\textwidth} m{0.15\textwidth} m{0.15\textwidth} }
    &
    Cardiomegaly &
    Consolidation &
    Lung Opacity & 
    Plural Effusion &
    Support Devices
 \\
    \rotatebox[origin=l]{90}{\parbox{0.16\textwidth}{IG with Smoothgrad squared}} &
    \frame{\includegraphics[width=0.16\textwidth]{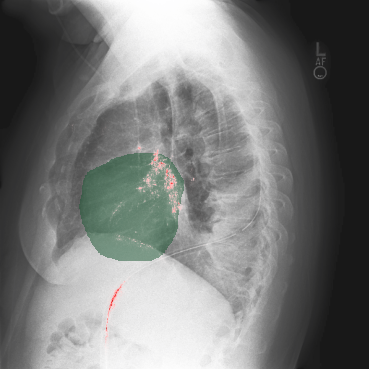}} &  
    \frame{\includegraphics[width=0.16\textwidth]{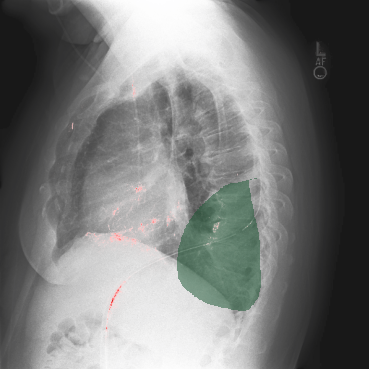}} &  
    \frame{\includegraphics[width=0.16\textwidth]{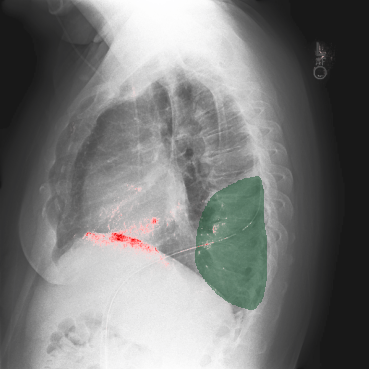}} &  
    \frame{\includegraphics[width=0.16\textwidth]{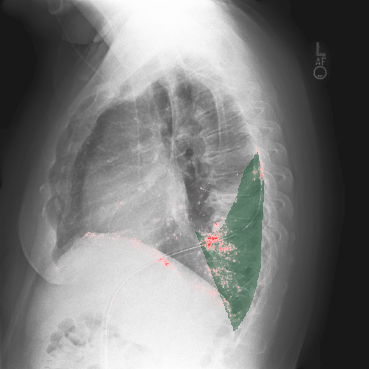}} &  
    \frame{\includegraphics[width=0.16\textwidth]{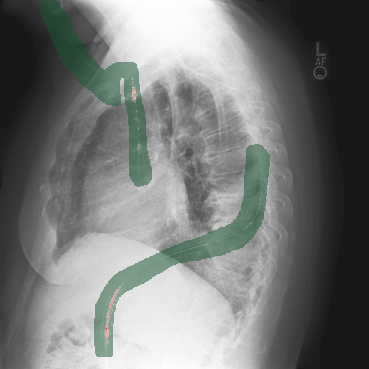}}
 \\
    \rotatebox[origin=l]{90}{\parbox{0.16\textwidth}{IG with Smoothgrad}} &
    \frame{\includegraphics[width=0.16\textwidth]{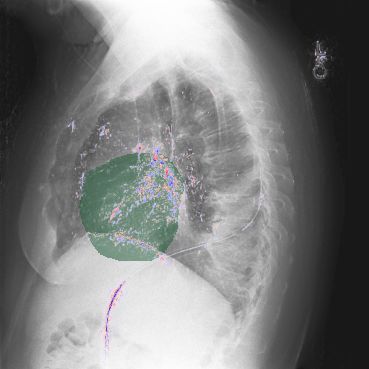}} &  
    \frame{\includegraphics[width=0.16\textwidth]{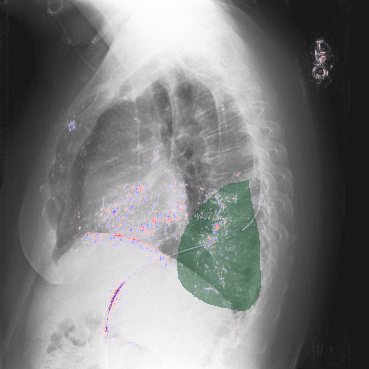}} &  
    \frame{\includegraphics[width=0.16\textwidth]{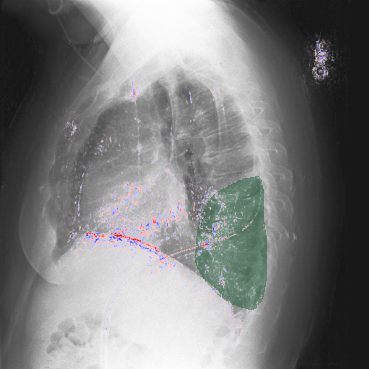}} &  
    \frame{\includegraphics[width=0.16\textwidth]{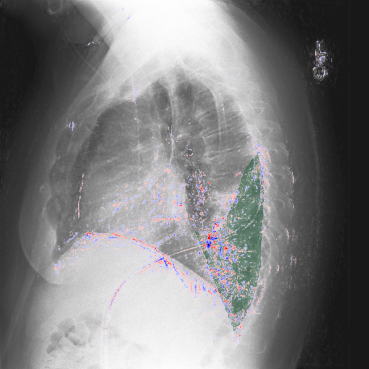}} &  
    \frame{\includegraphics[width=0.16\textwidth]{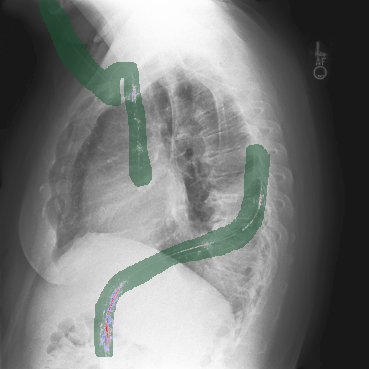}}
 \\
\end{tabular}

\caption{Saliency maps from a instance-level CheXpert trained using the sequential CBM method}
\label{fig:CBM chexpert concept saliency appendix 4}
\end{figure*}

\begin{figure*}[p]

\centering
\begin{tabular}{m{0.03\textwidth} m{0.15\textwidth} m{0.15\textwidth} m{0.15\textwidth} m{0.15\textwidth} m{0.15\textwidth} }
    &
    Atelectasis &
    Cardiomegaly &
    Edema &  
    Lung Opacity &  
    Pleural Effusion
 \\
    \rotatebox[origin=l]{90}{\parbox{0.16\textwidth}{IG with Smoothgrad squared}} &
    \frame{\includegraphics[width=0.16\textwidth]{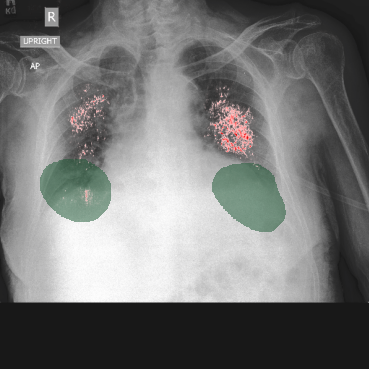}} &  
    \frame{\includegraphics[width=0.16\textwidth]{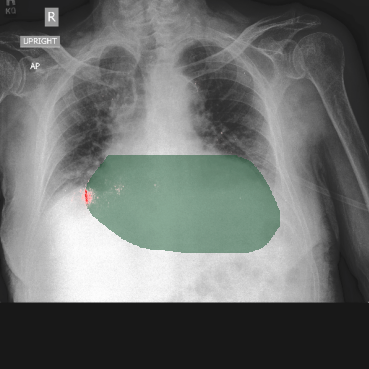}} &  
    \frame{\includegraphics[width=0.16\textwidth]{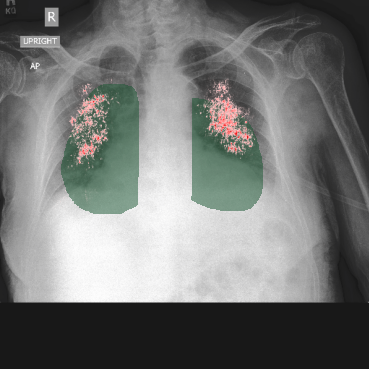}} &  
    \frame{\includegraphics[width=0.16\textwidth]{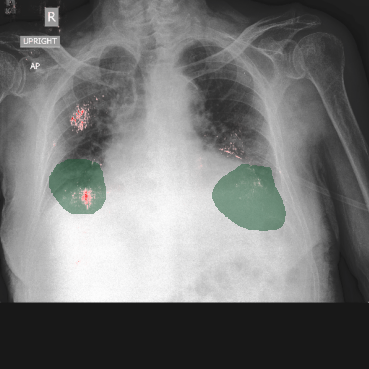}} &  
    \frame{\includegraphics[width=0.16\textwidth]{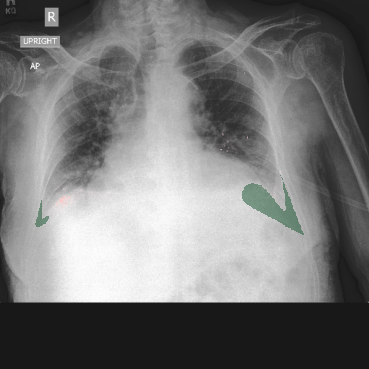}}
 \\
    \rotatebox[origin=l]{90}{\parbox{0.16\textwidth}{IG with Smoothgrad}} &
    \frame{\includegraphics[width=0.16\textwidth]{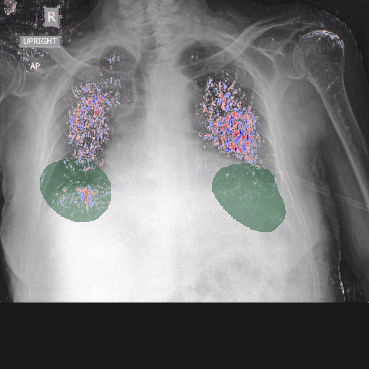}} &  
    \frame{\includegraphics[width=0.16\textwidth]{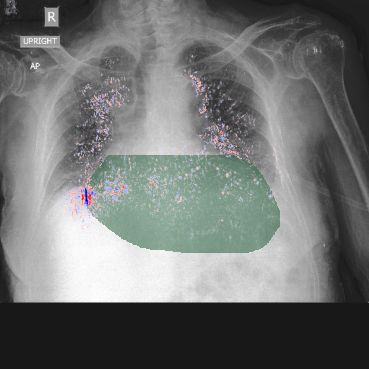}} &  
    \frame{\includegraphics[width=0.16\textwidth]{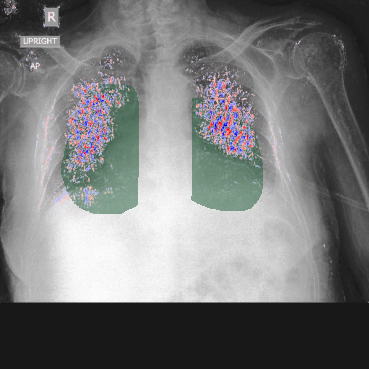}} &  
    \frame{\includegraphics[width=0.16\textwidth]{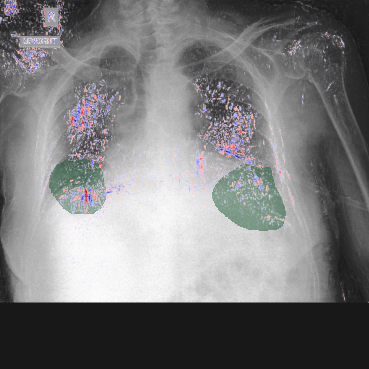}} &  
    \frame{\includegraphics[width=0.16\textwidth]{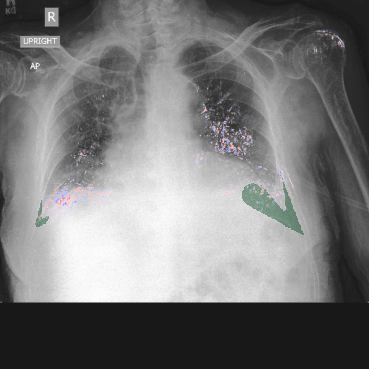}}
 \\
\end{tabular}

\caption{Saliency maps from a instance-level CheXpert trained using the sequential CBM method. The number beneath the saliency map is the concept prediction which is in the range of 0 and 1.}
\label{fig:CBM chexpert concept saliency appendix 3}
\end{figure*}

We show additional saliency maps results for instance-level CheXpert in Figure~\ref{fig:CBM chexpert concept saliency appendix 1}, \ref{fig:CBM chexpert concept saliency appendix 2}, \ref{fig:CBM chexpert concept saliency appendix 3} and \ref{fig:CBM chexpert concept saliency appendix 4} using three explanation techniques IG with a smoothgrad noise tunnel, and IG with a smoothgrad squared noise tunnel and Grad-CAM \citep{gradCAM}. IG with a smoothgrad noise tunnel and IG with a smoothgrad squared noise tunnel is configured to use 25 randomly generated samples and a standard deviation of 0.2.

\begin{figure*}[p]

%sample: 

\centering
\begin{tabular}{m{0.03\textwidth} m{0.15\textwidth} m{0.15\textwidth} m{0.15\textwidth} }
    &
    Lung Opacity &
    Pleural Effusion &
    Support Devices
 \\
    \rotatebox[origin=l]{90}{\parbox{0.16\textwidth}{IG with Smoothgrad squared}} &
    \frame{\includegraphics[width=0.16\textwidth]{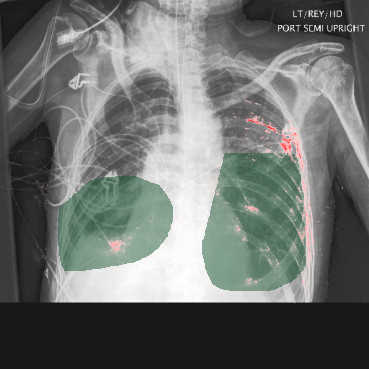}} &  
    \frame{\includegraphics[width=0.16\textwidth]{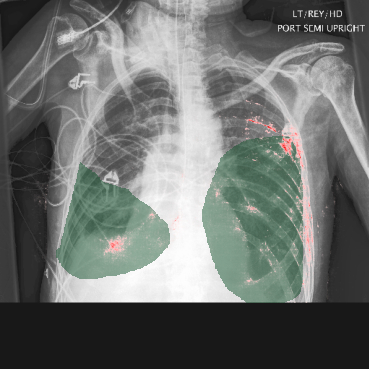}} &  
    \frame{\includegraphics[width=0.16\textwidth]{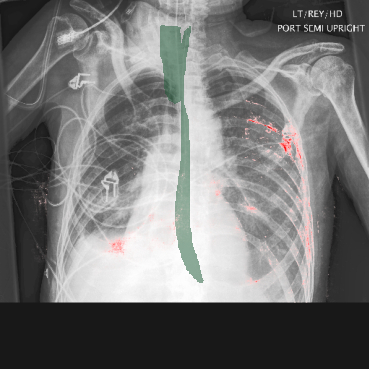}}
 \\
    \rotatebox[origin=l]{90}{\parbox{0.16\textwidth}{IG with Smoothgrad}} &
    \frame{\includegraphics[width=0.16\textwidth]{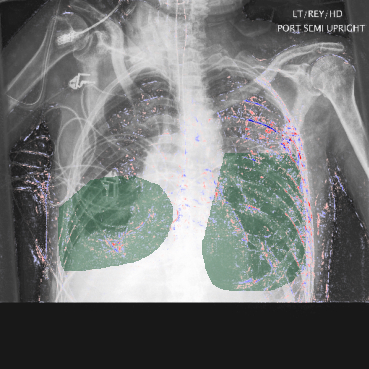}} &  
    \frame{\includegraphics[width=0.16\textwidth]{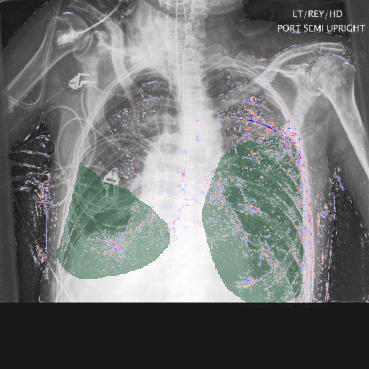}} &  
    \frame{\includegraphics[width=0.16\textwidth]{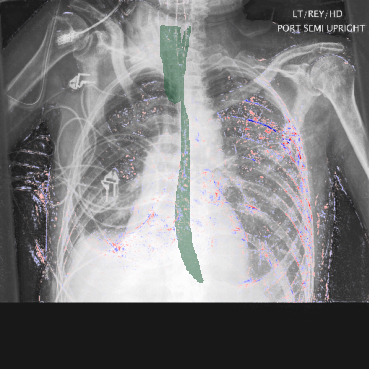}}
 \\
    \rotatebox[origin=l]{90}{\parbox{0.16\textwidth}{grad-CAM}} &
    \frame{\includegraphics[width=0.16\textwidth]{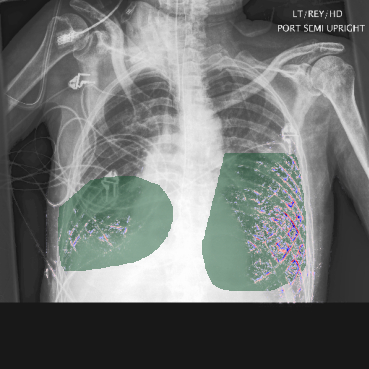}} &  
    \frame{\includegraphics[width=0.16\textwidth]{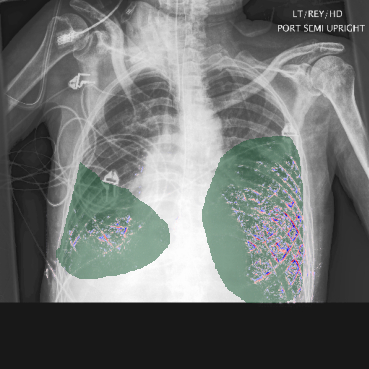}} &  
    \frame{\includegraphics[width=0.16\textwidth]{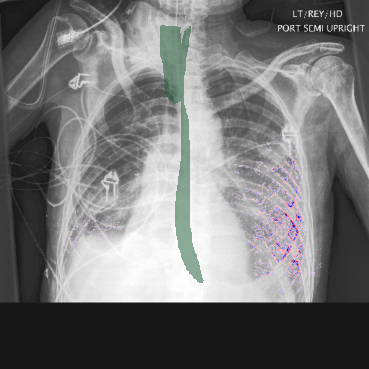}}
 \\
    &
     0.96053 &
     0.95868 &
     0.95689
\end{tabular}

\caption{Saliency maps from a class-level CheXpert with three concept present, trained using the sequential CBM method. The number beneath the saliency map is the concept prediction which is in the range of 0 and 1.}
\label{fig:CBM chexpert concept saliency appendix 5}
\end{figure*}

\begin{figure*}[p]

%sample: 

\centering
\begin{tabular}{m{0.03\textwidth} m{0.15\textwidth} m{0.15\textwidth} m{0.15\textwidth} }
    &
    Lung Opacity &
    Pleural Effusion &
    Support Devices
 \\
    \rotatebox[origin=l]{90}{\parbox{0.16\textwidth}{IG with Smoothgrad squared}} &
    \frame{\includegraphics[width=0.16\textwidth]{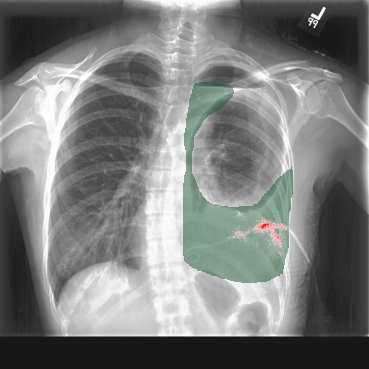}} &  
    \frame{\includegraphics[width=0.16\textwidth]{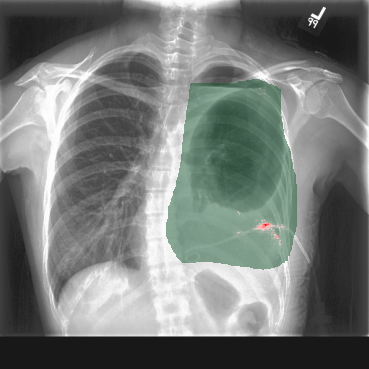}} &  
    \frame{\includegraphics[width=0.16\textwidth]{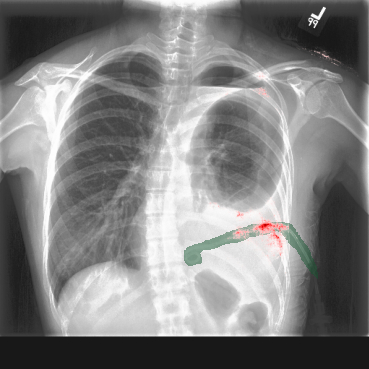}}
 \\
    \rotatebox[origin=l]{90}{\parbox{0.16\textwidth}{IG with Smoothgrad}} &
    \frame{\includegraphics[width=0.16\textwidth]{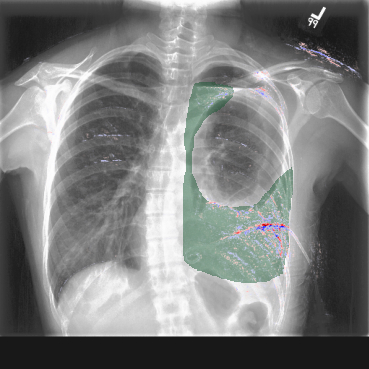}} &  
    \frame{\includegraphics[width=0.16\textwidth]{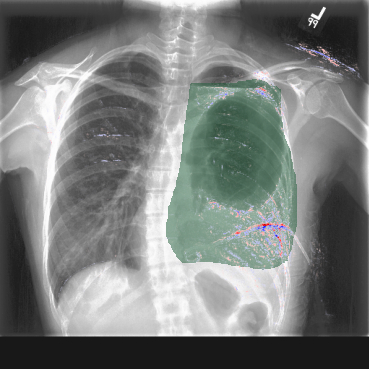}} &  
    \frame{\includegraphics[width=0.16\textwidth]{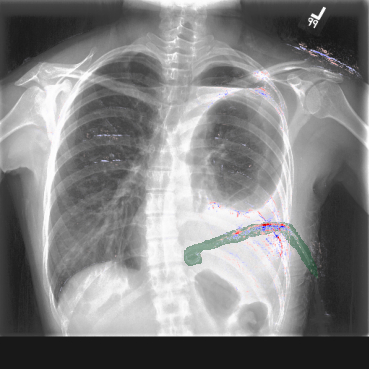}}
 \\
    \rotatebox[origin=l]{90}{\parbox{0.16\textwidth}{grad-CAM}} &
    \frame{\includegraphics[width=0.16\textwidth]{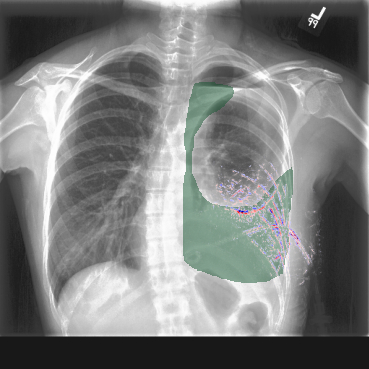}} &  
    \frame{\includegraphics[width=0.16\textwidth]{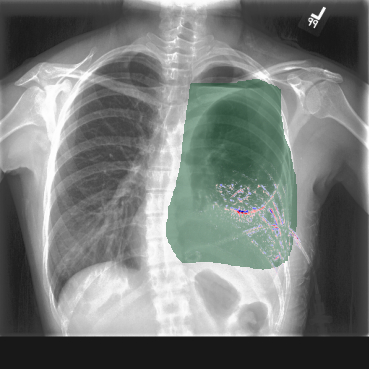}} &  
    \frame{\includegraphics[width=0.16\textwidth]{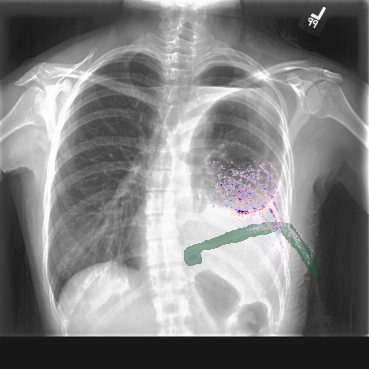}}
 \\
    &
     0.9152 &
     0.90924 &
     0.91351
\end{tabular}

\caption{Saliency maps from a class-level CheXpert with three concept present, trained using the sequential CBM method. The number beneath the saliency map is the concept prediction which is in the range of 0 and 1.}
\label{fig:CBM chexpert concept saliency appendix 6}
\end{figure*}

\begin{figure*}[p]

%sample: 

\centering
\begin{tabular}{m{0.03\textwidth} m{0.15\textwidth} m{0.15\textwidth} m{0.15\textwidth} m{0.15\textwidth} }
    &
    Edema &
    Lung Opacity &
    Pleural Effusion &
    Support Devices
 \\
    \rotatebox[origin=l]{90}{\parbox{0.16\textwidth}{IG with Smoothgrad squared}} &
    \frame{\includegraphics[width=0.16\textwidth]{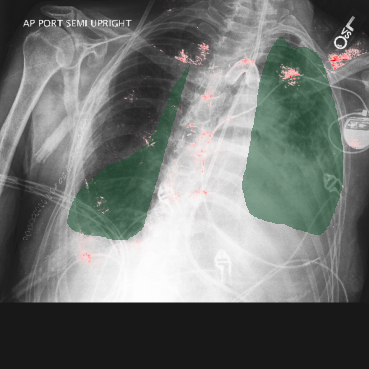}} &  
    \frame{\includegraphics[width=0.16\textwidth]{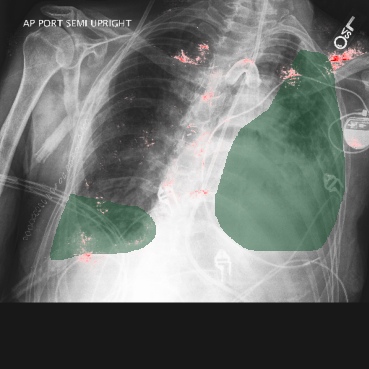}} &  
    \frame{\includegraphics[width=0.16\textwidth]{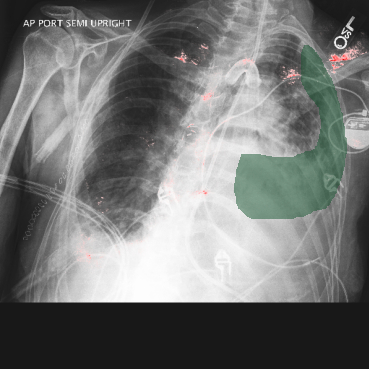}} &  
    \frame{\includegraphics[width=0.16\textwidth]{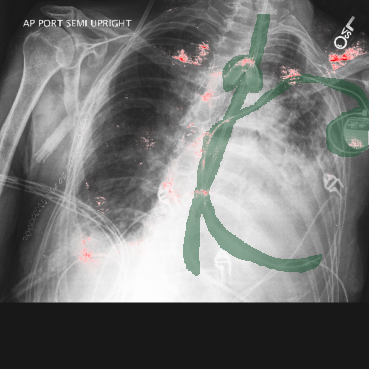}}
 \\
    \rotatebox[origin=l]{90}{\parbox{0.16\textwidth}{IG with Smoothgrad}} &
    \frame{\includegraphics[width=0.16\textwidth]{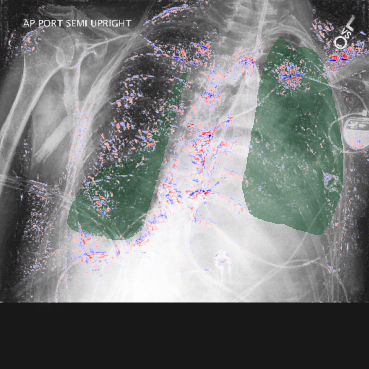}} &  
    \frame{\includegraphics[width=0.16\textwidth]{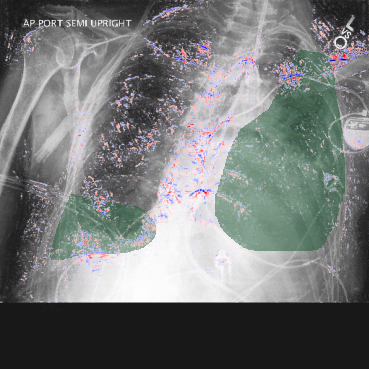}} &  
    \frame{\includegraphics[width=0.16\textwidth]{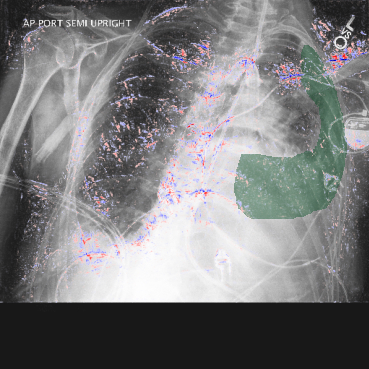}} &  
    \frame{\includegraphics[width=0.16\textwidth]{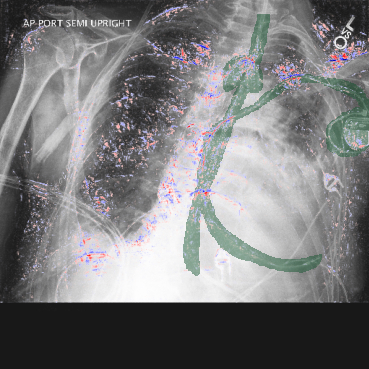}}
 \\
    \rotatebox[origin=l]{90}{\parbox{0.16\textwidth}{grad-CAM}} &
    \frame{\includegraphics[width=0.16\textwidth]{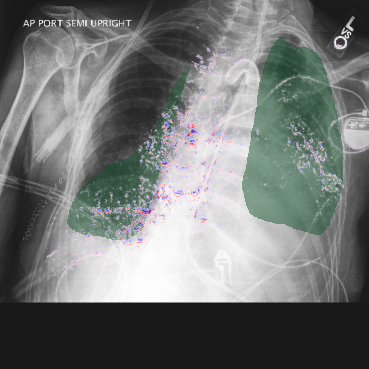}} &  
    \frame{\includegraphics[width=0.16\textwidth]{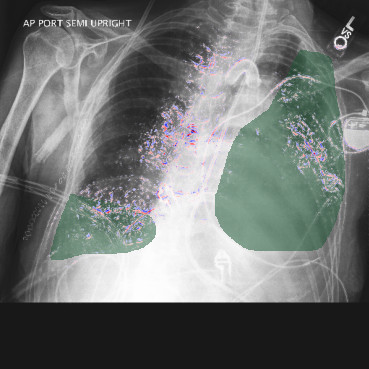}} &  
    \frame{\includegraphics[width=0.16\textwidth]{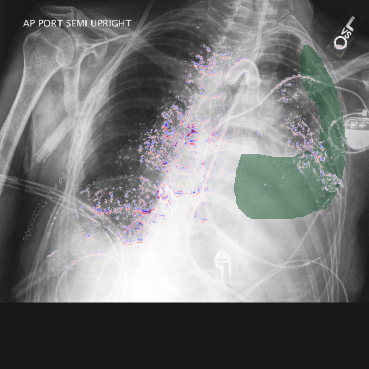}} &  
    \frame{\includegraphics[width=0.16\textwidth]{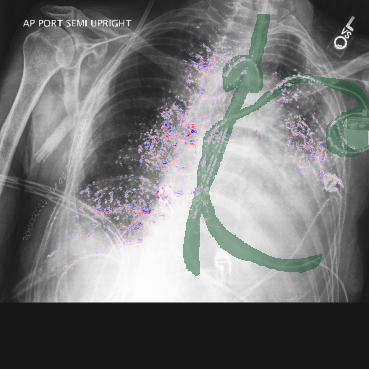}}
 \\
    &
     0.99165 &
     0.99156 &
     0.99143 &
     0.99158
\end{tabular}

\caption{Saliency maps from a class-level CheXpert with four concept present, trained using the sequential CBM method. The number beneath the saliency map is the concept prediction which is in the range of 0 and 1.}
\label{fig:CBM chexpert concept saliency appendix 7}
\end{figure*}

\begin{figure*}[p]

%sample: 

\centering
\begin{tabular}{m{0.03\textwidth} m{0.15\textwidth} m{0.15\textwidth} m{0.15\textwidth} m{0.15\textwidth} }
    &
    Edema &
    Lung Opacity &
    Pleural Effusion &
    Support Devices
 \\
    \rotatebox[origin=l]{90}{\parbox{0.16\textwidth}{IG with Smoothgrad squared}} &
    \frame{\includegraphics[width=0.16\textwidth]{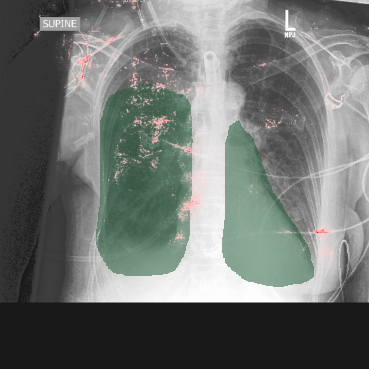}} &  
    \frame{\includegraphics[width=0.16\textwidth]{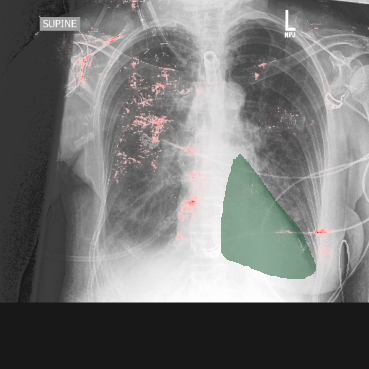}} &  
    \frame{\includegraphics[width=0.16\textwidth]{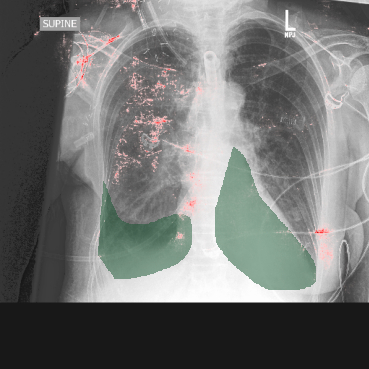}} &  
    \frame{\includegraphics[width=0.16\textwidth]{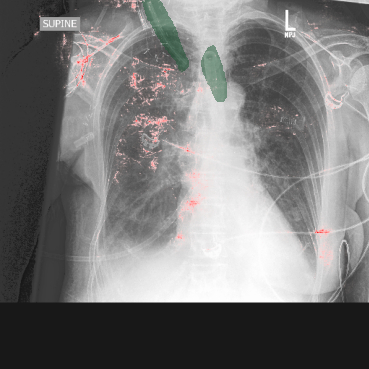}}
 \\
    \rotatebox[origin=l]{90}{\parbox{0.16\textwidth}{IG with Smoothgrad}} &
    \frame{\includegraphics[width=0.16\textwidth]{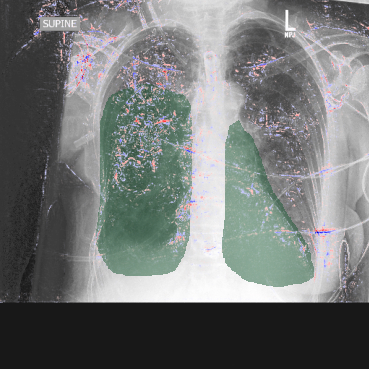}} &  
    \frame{\includegraphics[width=0.16\textwidth]{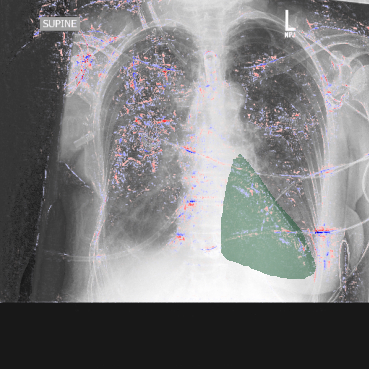}} &  
    \frame{\includegraphics[width=0.16\textwidth]{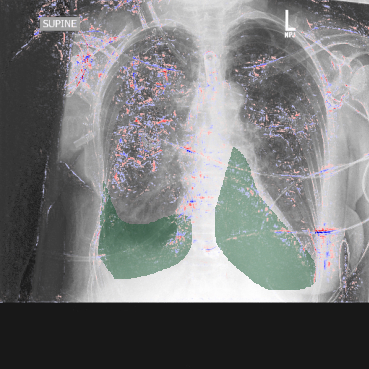}} &  
    \frame{\includegraphics[width=0.16\textwidth]{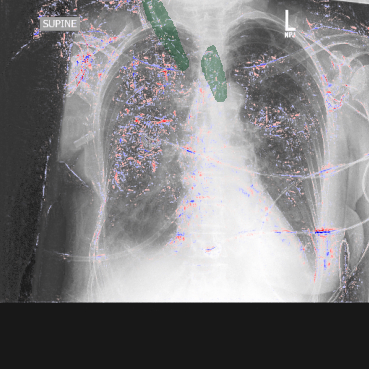}}
 \\
    \rotatebox[origin=l]{90}{\parbox{0.16\textwidth}{grad-CAM}} &
    \frame{\includegraphics[width=0.16\textwidth]{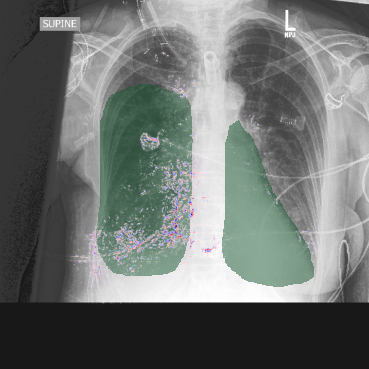}} &  
    \frame{\includegraphics[width=0.16\textwidth]{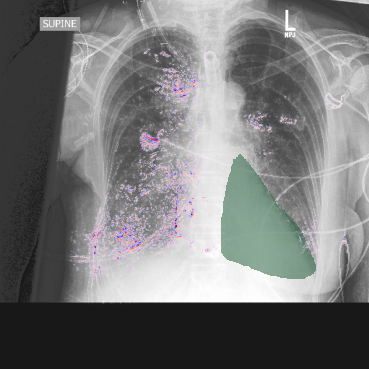}} &  
    \frame{\includegraphics[width=0.16\textwidth]{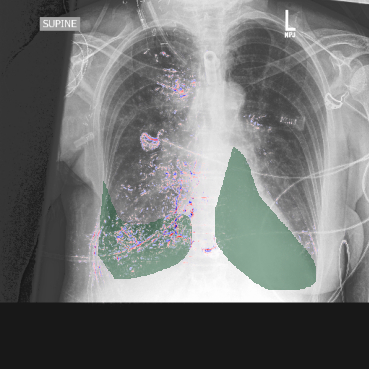}} &  
    \frame{\includegraphics[width=0.16\textwidth]{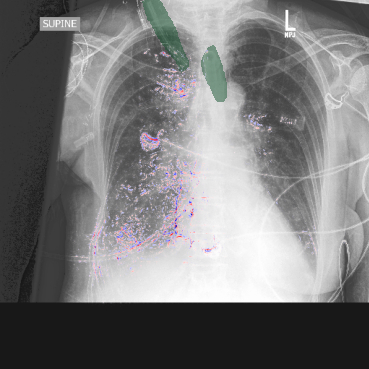}}
 \\
    &
     0.94589 &
     0.94416 &
     0.94416 &
     0.94479
\end{tabular}

\caption{Saliency maps from a class-level CheXpert with four concept present, trained using the sequential CBM method. The number beneath the saliency map is the concept prediction which is in the range of 0 and 1.}
\label{fig:CBM chexpert concept saliency appendix 8}
\end{figure*}

\begin{figure*}[p]

%sample: 24

\centering
\begin{tabular}{m{0.03\textwidth} m{0.15\textwidth} m{0.15\textwidth} m{0.15\textwidth} m{0.15\textwidth} m{0.15\textwidth} }
    &
    Atelectasis &
    Consolidation &
    Lung Opacity & 
    Pleural Effusion &
    Support Devices
 \\
    \rotatebox[origin=l]{90}{\parbox{0.16\textwidth}{IG with Smoothgrad squared}} &
    \frame{\includegraphics[width=0.16\textwidth]{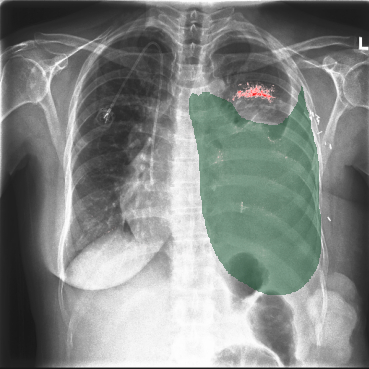}} &  
    \frame{\includegraphics[width=0.16\textwidth]{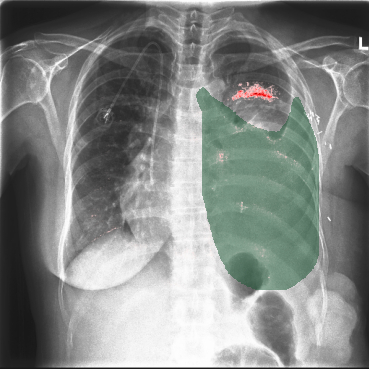}} &  
    \frame{\includegraphics[width=0.16\textwidth]{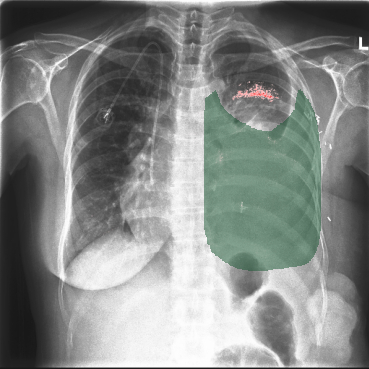}} &  
    \frame{\includegraphics[width=0.16\textwidth]{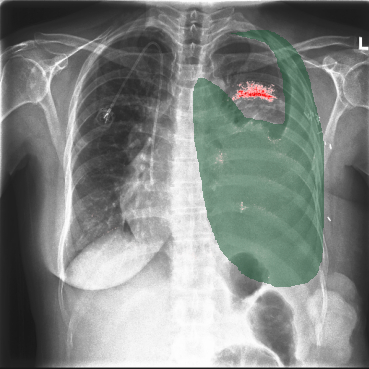}} &  
    \frame{\includegraphics[width=0.16\textwidth]{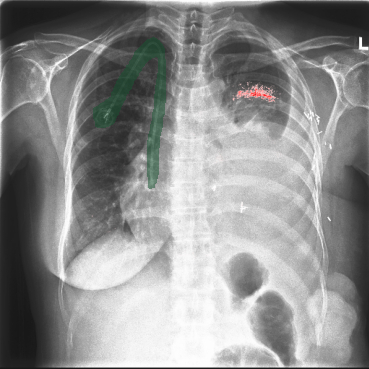}}
 \\
    \rotatebox[origin=l]{90}{\parbox{0.16\textwidth}{IG with Smoothgrad}} &
    \frame{\includegraphics[width=0.16\textwidth]{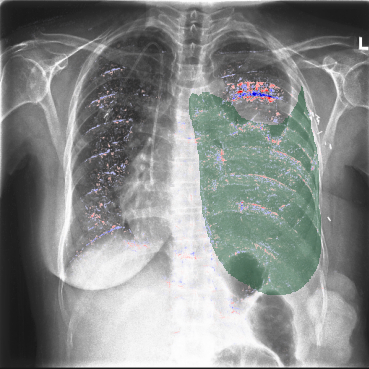}} &  
    \frame{\includegraphics[width=0.16\textwidth]{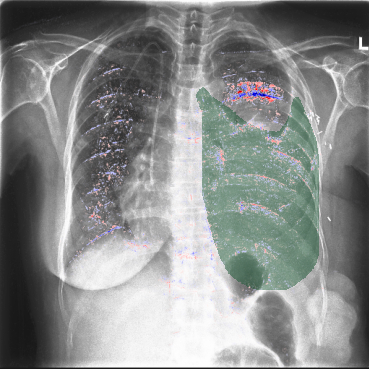}} &  
    \frame{\includegraphics[width=0.16\textwidth]{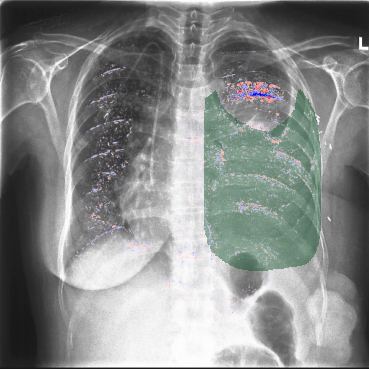}} &  
    \frame{\includegraphics[width=0.16\textwidth]{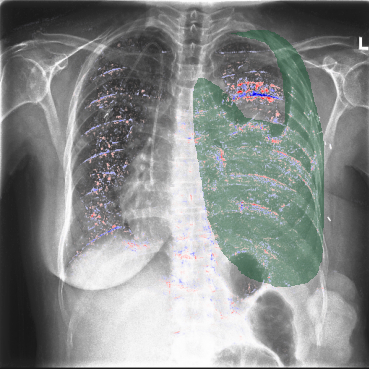}} &  
    \frame{\includegraphics[width=0.16\textwidth]{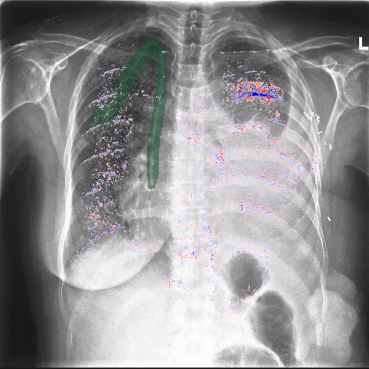}}
 \\
\end{tabular}

\caption{Saliency maps from a class-level CheXpert with five concept present, trained using the sequential CBM method. The number beneath the saliency map is the concept prediction which is in the range of 0 and 1.}
\label{fig:CBM chexpert concept saliency appendix 9}
\end{figure*}

For the most part, relevance is localised but fairly noisy, especially for IG. For a number of concepts, saliency is also not within ground truth segmentations (the areas highlighted with green). The main take away however is the models are predicting concepts using distinctly different input features. Class-level CheXpert tells a different story as all concepts share very similar saliency maps and thus, the model is using the same input features to predict all concepts. This can be seen in Figure~\ref{fig:CBM chexpert concept saliency appendix 5}, \ref{fig:CBM chexpert concept saliency appendix 6}, \ref{fig:CBM chexpert concept saliency appendix 7}, \ref{fig:CBM chexpert concept saliency appendix 8}, \ref{fig:CBM chexpert concept saliency appendix 9} and \ref{fig:CBM chexpert concept saliency appendix 10}.

\clearpage

\begin{figure*}[p]

%sample: 34

\centering
%

\caption{Saliency maps from a class-level CheXpert with five concept present, trained using the sequential CBM method. The number beneath the saliency map is the concept prediction which is in the range of 0 and 1.}
\label{fig:CBM chexpert concept saliency appendix 10}
\end{figure*}

\end{document}